\documentclass[manuscript,screen,nonacm]{acmart}
\AtBeginDocument{%
  }

\begin{document}

\title{NeuroRule: Making Black-Box Neural Networks Explainable\\through Rule-set Evolution}

\author{Tapaswini Kodavanti}
\email{tk24428@utexas.edu}
\affiliation{%
  \institution{The University of Texas at Austin}
  \city{Austin}
  \state{Texas}
  \country{USA}
}

\author{Hormoz Shahrzad}
\email{hshahrzad@cs.utexas.edu}
\orcid{0000-0002-5983-4531}
\affiliation{%
  \institution{The University of Texas at Austin}
  \city{Austin}
  \state{Texas}
  \country{USA}
}
\affiliation{%
  \institution{Cognizant AI Lab}
  \city{San Francisco}
  \state{California}
  \country{USA}
}

\author{Risto Miikkulainen}
\email{risto@cs.utexas.edu}
\orcid{0000-0002-0062-0037}
\affiliation{%
  \institution{The University of Texas at Austin}
  \city{Austin}
  \state{Texas}
  \country{USA}
}
\affiliation{%
  \institution{Cognizant AI Lab}
  \city{San Francisco}
  \state{California}
  \country{USA}
}

\renewcommand{\shortauthors}{Kodavanti, Shahrzad, and Miikkulainen}

\begin{abstract}
  High-capacity neural network models have achieved state-of-the-art performance across diverse classification tasks, yet they frequently operate as black-box models, lacking the transparency necessary for critical decision-making. Such opacity creates a persistent trade-off between performance and explainability. This paper proposes a solution to address this gap: the NeuroRule knowledge distillation framework that results in explainable rule-sets from neural network models. NeuroRule adapts the EVOTER rule-set evolution infrastructure to treat neural networks as targets for the evolution process, distilling their performance into concise sets of propositional logic expressions. There are three primary contributions: (1) an evolutionary method for distilling black-box neural network models into explicit rule-set models; (2) a method for making rule sets more explainable by including a conciseness objective to evolution; and (3) a demonstration that the distillation is viable even without access to the original neural network training data. The paper thus establishes that black-box neural network models can be made explainable and therefore useful in real-world applications where trustworthiness is paramount.
\end{abstract}




\maketitle

\section{Introduction}
High-capacity neural network models are frequently used in classification tasks, achieving state-of-the-art performance \cite{lecun2015deep}. However, they often operate as black-box models and lack the transparency required for accountability \cite{rudin2019stop}. In some domains, this issue can lead to costly mistakes (e.g.\ failures in self-driving cars \cite{batishchev2025black}) or raise ethical concerns (e.g.\ gender and racial biases in health care \cite{challen2019artificial, cross2024bias}). This opacity results in a performance vs.\ explainability trade-off: simpler models that are easier to understand lack performance, while high-performing models often remain the least explainable. While post-hoc explanations such as LIME \cite{ribeiro2016whyitrustyou} or SHAP \cite{lundberg2017unifiedapproachinterpretingmodel} provide per-instance explanations, few offer an understanding of the entire model's behavior. 

In contrast, frameworks such as Evolution of Transparent Explainable Rule-sets, or EVOTER, \cite{shahrzad2024evoterevolutiontransparentexplainable} focus on developing models that are explainable by design. Propositional logic rule-sets are evolved for the task using the same training data as for neural networks. However, it is difficult to achieve the level of performance of neural networks, and the best-performing rule-sets are often overly complex, making them less explainable. Furthermore, the original data used to train the neural network model may not be available, so such rule-sets cannot always be evolved directly.

This paper addresses these gaps by developing a knowledge distillation framework, NeuroRule, which extracts explainable rule-set models from black-box neural networks. Building on EVOTER, three further mechanisms are proposed: (1) utilizing the neural network as a target for the rule-set evolution; (2) adding rule-set conciseness as a secondary objective; and (3) evolving rule-sets by sampling the neural network model when the original network training data is not available.

The results are surprising: First, NeuroRule often performs better than EVOTER, and sometimes even better than the target neural network itself, particularly with out-of-distribution samples. The likely reason is that the neural network provides more regularized samples, and the extracted rules capture general principles that generalize well. Second, the conciseness objective significantly reduces the complexity of the evolved rule-sets, forcing them to represent principles rather than noise, resulting in better explainability. Third, evolving with data sampled from the neural network results in only a slight reduction in performance compared to evolving with samples from the original dataset, demonstrating that the NeuroRule approach is viable even when the original data is not available.

Thus, knowledge distillation through NeuroRule is a viable pathway for translating black-box neural networks into human-readable logic, making it possible to take advantage of machine learning in applications where explainability is critical.

\section{Related Work}
This section will review previous work on explainability methods for neural networks  (Section~\ref{sec:post_hoc_explanations}) and frameworks for developing models that are explainable by design  (Section~\ref{sec:related_work_evoter}). In the literature, \emph{transparency} is often used to refer to elements, \emph{interpretability} to mechanisms, and \emph{explainability} to understanding, but these terms overlap significantly in meaning and usage.

\subsection{Post-hoc Explanation methods for Neural Networks}
\label{sec:post_hoc_explanations}
As neural network models have become the standard for classification tasks, their inherent black-box nature has prompted a surge in research into post-hoc explainability \cite{simonyan2013deep, selvaraju2017grad}.These methods aim to explain the decisions of a pre-trained model without altering its underlying architecture. Currently, the most prominent techniques in this domain are local feature-attribution methods such as LIME (Local Interpretable Model-agnostic Explanations \cite{ribeiro2016whyitrustyou}) and SHAP (SHapley Additive exPlanations \cite{lundberg2017unifiedapproachinterpretingmodel}). 

LIME operates by perturbing a specific input instance and observing the resulting changes in the model's output to fit a simplified, linear surrogate model into the local neighborhood of that point. While effective in providing a point-of-interest explanation, LIME often suffers from instability: small changes in the sampling process can lead to different explanations for the same input. SHAP improves on this technique by utilizing a game theory approach to assign each feature an importance value (Shapley value) based on its contribution to the final prediction. 

However, both LIME and SHAP are fundamentally local explainability tools. They explain why a specific data point was assigned a class label, but they do not provide a comprehensive logical interpretation of the model's decision behavior. In a clinical setting, a medical professional requires more than a per-instance explanation; they require a set of persistent, auditable rules that remain consistent across the entire patient population. Furthermore, these methods do not produce a standalone model; explanations remain external to the main classification network. An alternative approach is based on models that are explainable by design, such as EVOTER.

\subsection{EVOTER Framework}
\label{sec:related_work_evoter}
EVOTER (Evolution of Transparent Explainable Rule-sets \cite{shahrzad2024evoterevolutiontransparentexplainable}) evolves rule-sets based on extended propositional logic expressions. Unlike traditional Genetic Programming (GP) \cite{poli2008field}, which often utilizes tree-based structures prone to bloat (i.e.\ sparse and overly complex structures) and high computational overhead, EVOTER employs a linear list of rules that maps input features to discrete actions or classifications.

The primary advantage of the EVOTER grammar is its ability to discover meaningful, human-readable logic that behaves similarly to black-box models. Because the rules are expressed in standard propositional logic, they provide direct insight into the decision manifold and make hidden biases explicit. Furthermore, the symbolic nature of EVOTER allows for human-in-the-loop interventions; unlike neural weights, these rules can be directly edited by domain experts to remove identified biases or add logical constraints. By evolving these rule-sets, EVOTER establishes a foundation for AI systems where the final model is inherently auditable.

Previous applications of EVOTER have shown success in both prediction and policy search. In this paper, EVOTER is placed into a new role: it will emulate the behavior of black-box neural networks, hence making them explainable. However, while EVOTER provides the grammar for interpretability, the search space for optimal rule-sets is vast and characterized by a sharp trade-off between accuracy and complexity. A straightforward application of EVOTER in this role results in rule-sets that are overly complex. Therefore, a method is developed in this paper to utilize a conciseness objective to improve explainability.

\section{Methodology}
This section will present the rule-set model framework (Section~\ref{rule_set_framework}), fitness function design (Section~\ref{fitness_function_design}), and knowledge distillation pipeline (Section~\ref{knowledge_distillation_pipeline}) used in the experiments.

\subsection{Rule-Set Model Framework}
\label{rule_set_framework}
NeuroRule adopts the propositional logic framework established in the EVOTER system \cite{shahrzad2024evoterevolutiontransparentexplainable} to represent model behavior and to perform single and multi-objective optimization. 

\subsubsection{Rule-Set Representation}
A rule-set is defined as a collection of $n$ discrete rules structured as a set of disjunct conditional statements. The approach consists of three designs, i.e.\ rule structure, condition structure, and the execution process.

\paragraph{Rule Structure}
Each individual rule follows a standard antecedent-consequent structure:
\begin{equation}
\mathbf{IF} \quad \left( Condition_1 \land Condition_2 \land \dots \land Condition_k \right) \quad \mathbf{THEN} \quad Action( \mathcal{A}).
\end{equation}
The antecedents are conjunctions (AND) of feature comparisons, while the relationships between separate rules are logical disjunctions (OR). This form results in a linear list structure that mitigates bloat that can occur during the evolution of tree structures while still maintaining logical completeness.

\begin{figure}[!t]
\begin{minipage}{0.7\textwidth}
\centering
\includegraphics[width=\linewidth]{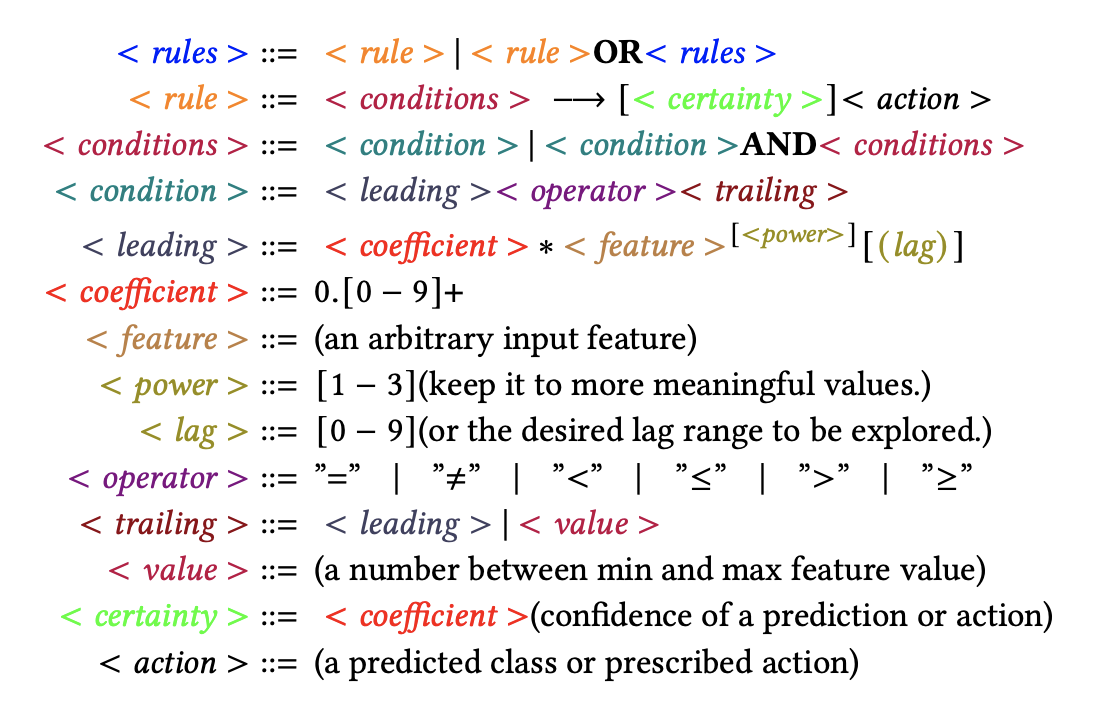}
\end{minipage}
\vspace*{-2ex}
\caption{The grammar for the rules. Rules are written in propositional logic, and include feature comparison, feature exponentiation, and time lags. For simplicity, this paper does not use time lags; they are shown for completeness. This representation allows for easy understanding. Figure from \cite{shahrzad2024evoterevolutiontransparentexplainable}.}
\label{fig:evoter_grammar}
\end{figure}

\begin{figure}[!t]
\begin{minipage}{0.7\textwidth}
\centering
\includegraphics[width=\linewidth]{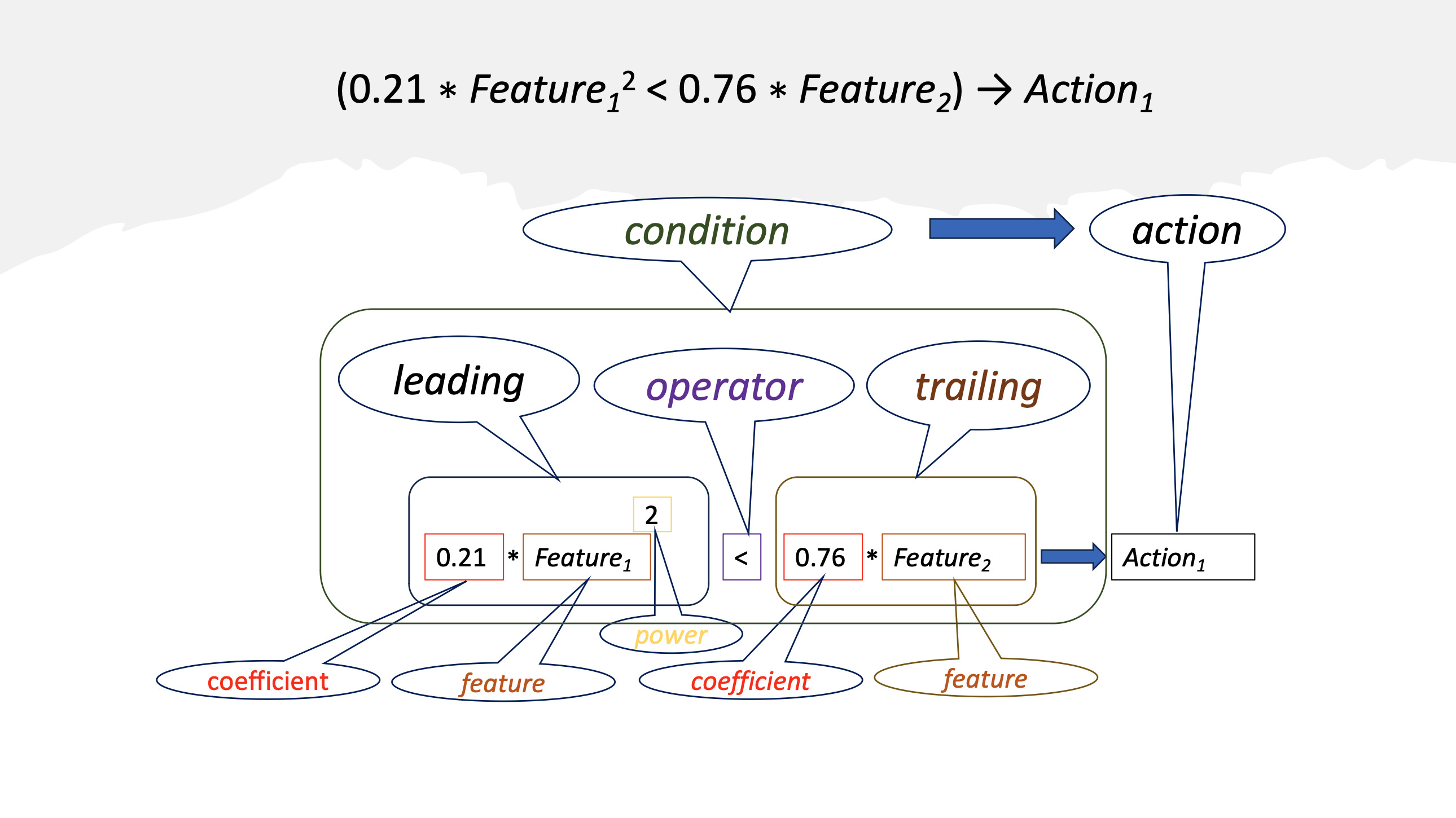}
\end{minipage}
\vspace*{-6ex}
\caption{An example rule with elements color-coded from the grammar. Rules can contain multiple conditions, and the rule-set includes a default rule that is used if none of the non-default rules are applied. This diagram demonstrates the standard structure followed by all rules in the rule-sets. Figure from \cite{shahrzad2024evoterevolutiontransparentexplainable}.}
\label{fig:evoter_sample_rule}
\end{figure}

\paragraph{Conditional Structure}
A condition within a rule compares a feature $x_i$ against a constant threshold $\tau$ or another feature $x_j$. To capture nonlinear relationships, the framework supports
\begin{itemize}
    \item Linear coefficients: $c \cdot x_i$,
    \item Power expressions: $(x_i)^p$,
    \item Comparison operators: $>, <, \ge, \le$.
\end{itemize}
In this paper, the time-lag features available in the original EVOTER framework are omitted to simplify the search space for classification tasks.  The complete grammar from EVOTER is shown in Figure~\ref{fig:evoter_grammar}, and a sample rule in Figure~\ref{fig:evoter_sample_rule}. 

\paragraph{Execution Process}
When an input vector is passed through the rule-set, the conditions are evaluated in sequence. For classification tasks in this paper, the Hard-Max Filter method is used as described in EVOTER \cite{shahrzad2024evoterevolutiontransparentexplainable}:

\begin{enumerate}
    \item All rules whose antecedents evaluate to \emph{True} are triggered. 
    \item Each triggered rule proposes an action (class label) with an associated coefficient. 
    \item The final prediction is the class label with the highest cumulative coefficient value.
    \item Default: Each set concludes with a default rule, which acts as a logical {\bf ELSE} statement. This rule captures the most common class label, allowing the rule-sets to evolve to find exceptions to the default rule (e.g. markers for a disease). 
\end{enumerate}

\subsubsection{Evolutionary Optimization}
The evolution of rule-sets is conducted through a specialized genetic algorithm designed to navigate a variable-length, discrete search space. This algorithm consists of crossover and mutation components:

\paragraph{Crossover}
To maintain a diverse population and explore logical combinations efficiently, the system uses three crossover mechanisms, chosen randomly at each generation:
\begin{itemize}
    \item Single-Point Crossover: Combines parent rule-sets at randomly selected crossover points (rules above one crossover point are concatenated with rules below the other crossover point). This strategy allows the number of rules in the offspring to grow or shrink in comparison to the parents.
    \item Uniform Crossover: Randomly samples rules from parent rule-sets to form a new set. This strategy preserves high-performing logic groups.
    \item Combine-Rules Crossover: Merges conditions and rules that share the same action, resulting in hybrid rules. This strategy promotes the evolution of higher-precision rules by consolidating overlapping logical predicates. 
\end{itemize}


%
%
%
%
%

\paragraph{Mutation}
Mutation occurs at three granularities to ensure both fine-tuning and global solution exploration: 
\begin{itemize}
    \item Condition Level: Modifies elements such as the condition's feature threshold ($\tau$), relational operators, and linear coefficients.
    \item Rule Level: Alters a rule's complexity by adding, removing, and modifying its specific conjunction of antecedents.
    \item Rule-Set Level: Operates on the global architecture by adding/removing entire rules, reordering rules, and changing the default rule.
\end{itemize}

To improve search efficiency, the framework implements a policy where any mutation that results in a contradiction (e.g.\ $\emph{Age} < 20 \land  \emph{Age} > 30$) or redundancy (e.g. $\emph{Age} < 20  \land \emph{Age} < 15$) is discarded. This reduction ensures that the evolutionary search remains grounded in valid logic. 


\subsection{Fitness Function Design}
\label{fitness_function_design}
Evolutionary search in NeuroRule is guided by a fitness vector ${f}(\theta)$ that evaluates the quality of a candidate rule-set $\theta$. In Experiments~1 and~3, the search is driven exclusively by an accuracy objective, $f_\mathrm{acc}(\theta)$. In Experiment~2, a conciseness objective  $f_\mathrm{con}(\theta)$ is added to optimize for conciseness alongside accuracy. Thus, the fitness vector is defined as
\begin{equation}
{f}(\theta) =
\begin{cases}
[f_\mathrm{acc}(\theta)] & \text{Single-Objective}, \\
[f_\mathrm{acc}(\theta), f_\mathrm{con}(\theta)] & \text{Multi-Objective}.
\end{cases}
\end{equation}

\subsubsection{Accuracy Objective}
\label{subsec:accuracy_objective}
The primary objective of the evolutionary search is to maximize the accuracy of the rule-set in predicting the outputs of the target neural network. Given a set of $N$ data points and a black-box neural network model NN, the accuracy fitness ${f}_{\rm acc}(\theta)$ for a rule-set $\theta$ is defined as the proportion of correct classifications:
\begin{equation}
f_\mathrm{acc}(\theta)=\frac{1}{N}\sum^N_{i=1}\mathrm{RuleSet}_{\theta}(x_i)=\mathrm{NN}(x_i).
\end{equation}

\subsubsection{Conciseness Objective}
\label{subsec:conciseness_objective}
The second objective, conciseness, is defined as the minimization of complexity to promote human readability. Fewer rules, smaller rule structures, and fewer literals all reduce the cognitive overhead required for a domain expert to understand the model's behavior \cite{miller1956magical, lage2019evaluation}. In safety-critical environments (such as clinical diagnostics), an interpretable model is only as useful as it is accessible; by minimizing the total number of conditions, the evolutionary search is pressured to discover logic that maintains accuracy without forcing a human observer to parse through redundant or overly granular logical branches.

Conciseness is quantified through the Condition Count metric $\mathcal{L}$, which sums the total number of atomic conditions across all rules in a set. This metric serves as a proxy for explainability: a lower $\mathcal{L}$ corresponds to a more concise rule-set representation, facilitating an easier understanding of the model's logic.

 For a rule-set $\theta$ containing rules $R=\{r_1, r_2, ... r_k\}$, where each rule $r_j$ contains a conjunction of conditions $c$, the conciseness fitness $f_{con}$ is defined as
\begin{equation}
f_\mathrm{con}(\theta)=-1 \times \sum_{r \in R} | \{c : c \in r \} |.
\end{equation}

\subsubsection{Multiobjective optimization}
To optimize both accuracy and conciseness at once, the NeuroRule framework utilizes the Non-dominated Sorting Genetic Algorithm II (NSGA-II \cite{nsgaii}). Unlike single-objective evolution, which collapses all metrics into a single scalar value, NSGA-II maintains a Pareto front of non-dominated solutions. Selection is facilitated by two primary mechanisms:
\begin{itemize}
    \item Non-dominated Sorting: Individuals are assigned to hierarchical fronts based on the principle of Pareto Dominance. A rule-set $\theta_1$ \textit{dominates} $\theta_2$ $(\theta_1 \prec \theta_2$) if it is no worse in any objective and strictly better in at least one. 
    \item Crowding Distance: To ensure a diverse spread of solutions across the complexity spectrum, NSGA-II prioritizes individuals in less populated regions of the objective space. This selection heuristic prevents the population from prematurely converging on a single complexity level. 
\end{itemize}

\noindent These mechanisms allow the NSGA-II algorithm to facilitate the search across a Pareto front of multiple objectives, finding the best tradeoffs between them. 

\subsection{Knowledge Distillation Pipeline}
\label{knowledge_distillation_pipeline}
The core of the methodology is a distillation process where a complex black-box model, i.e.\ the target neural network NN, acts as a teacher to the transparent model (the rule-set) \cite{hinton2015distilling}. This pipeline ensures that the evolutionary search is guided by the behavior of the NN rather than the original training data \cite{frosst2017distilling}. 

\begin{figure}[!t]
\begin{minipage}{0.82\textwidth}
\centering
\includegraphics[width=\linewidth]{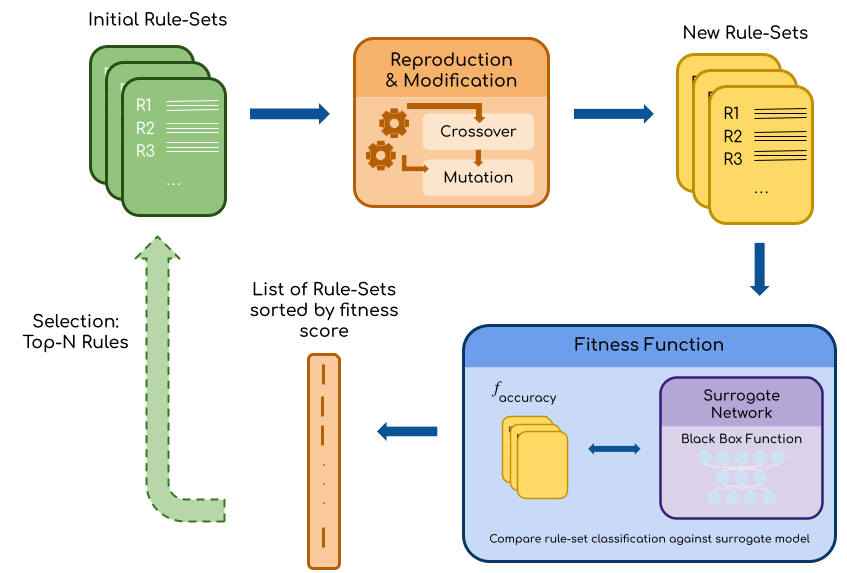}
\end{minipage}
\vspace*{-1ex}
\caption{The NeuroRule method with an Accuracy objective. A target neural network (NN) is first trained to optimize accuracy in predicting the labels in the original dataset. Rule-sets are then evolved through a genetic algorithm to match the behavior of the NN. Whereas the NN is a black box, the rule-sets are explainable. Knowledge in the network is thus distilled into an explainable model.} 
\label{fig:single_objective_pipeline}
\end{figure}

The distillation process begins with the training of the NN on the available dataset $\mathcal{D}$. The objective is to produce a model that achieves maximum accuracy in predicting the labels in the data. Once trained, the NN is integrated into the evolutionary fitness function. During each generation, the rule-sets are evaluated against the predictions generated by the NN, not the original labels in $\mathcal{D}$. This comparison allows the evolutionary algorithm to optimize the degree to which the rule-set reconstructs the NN's behavior rather than the fit to the original data. This process is illustrated in Figure~\ref{fig:single_objective_pipeline}.

\begin{figure}[!t]
\begin{minipage}{1.0\textwidth}
\centering
\includegraphics[width=\linewidth]{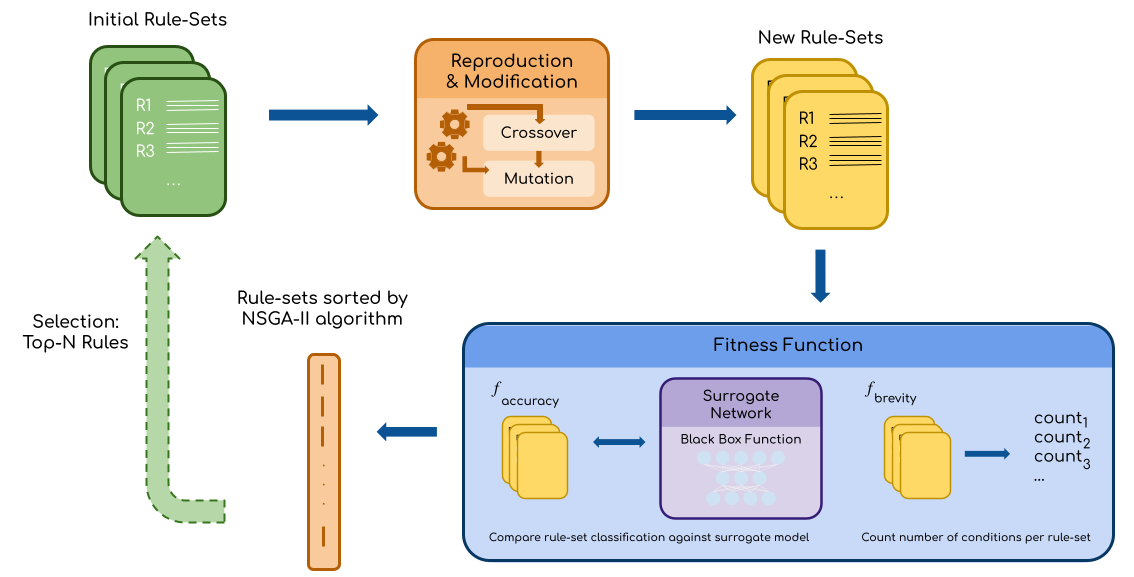}\\[-1ex]
\end{minipage}
\vspace*{-1ex}
\caption{The NeuroRule method with Accuracy and Conciseness objectives. Conciseness of the solutions is measured through Condition Counts. Optimizing both objectives together results in solutions that are less complex and therefore more explainable than when optimizing for accuracy alone.}
\label{fig:multi_objective_pipeline}
\end{figure}

In the case of optimizing for both performance and explainability, the fitness function is revised to use a multi-objective algorithm, as described in Section~\ref{subsec:conciseness_objective}. The multi-objective process is illustrated in Figure~\ref{fig:multi_objective_pipeline}. 

\section{Experimental Setup}
The experimental design is characterized by the choice of datasets (Section~\ref{exp_sec:data}), experimental parameters (Section~\ref{exp_sec:param}), and metrics (Section~\ref{exp_sec:metrics}).

\subsection{Datasets}
\label{exp_sec:data}
This section details the three clinical datasets used for evaluation. It also describes the data partitioning strategy into in-distribution and out-of-distribution partitions, and the generation of a synthetic dataset based on the behavior of the NN only, without access to the original data with which it was trained.

\subsubsection{Clinical Datasets}
\label{clinical_datasets}
In the medical domain, the black-box nature of standard neural networks poses a significant barrier to adoption. Critical diagnostic decisions require post-hoc auditability to allow clinicians to verify the logical steps leading to a prediction. Distillation of neural networks into rule-sets provides a transparency required for such auditability. NeuroRule was thus evaluated with three medical classification datasets from the UC Irvine collection \cite{UCI_Repository}:

\begin{itemize}
    \item Breast Cancer Wisconsin: A diagnostic dataset with features derived from digitized images of fine needle aspirates (FNA) of breast masses \cite{bc_wisconsin}.
    \item Heart Failure Prediction: A comprehensive dataset that combines multiple heart disease datasets (Cleveland, Hungary, Switzerland, Long Beach VA, and Stalog) \cite{heart_disease}.
    \item Diabetes Health Indicators: A large-scale dataset derived from the CDC's Behavioral Risk Factor Surveillance System \cite{diabetes}.
\end{itemize}

\subsubsection{ID/OOD Partitioning of Data}
\label{data_partitioning}
To evaluate the generalization capabilities of the models, a \textit{Data Partitioning} strategy was implemented. The data was divided based on its statistical distribution, categorizing samples as either middle data (i.e.\ samples where all continuvariable values were in the middle of the distribution) or tail data (i.e\ samples with extreme variable values at the tail of the distribution). These partitions are referred to as in-distribution (ID) and out-of-distribution (OOD) in this paper. The NN was trained on three distinct partitions where ID comprised the middle 80\%, 90\%, and 95\% of the data, and OOD the rest. It was then possible to evaluate the models in two respects: (1) how they scale wrt.\ the amount of available training data, and (2) how their performance decreases as they extrapolate farther away from the training data.

\subsubsection{Creating Synthetic Data}
\label{sec:sampling_strat}
In the most challenging real-world scenarios, only the NN is available; the original data is no longer available. To evaluate NeuroRule under such conditions, the pipeline was extended to such data-agnostic scenarios. Input data samples were generated randomly in the space of input features; they were then given to the NN (instead of the original training samples), and the corresponding outputs generated by the NN were recorded. The dataset $\mathcal{D}_\mathrm{syn}$ obtained in this manner, called the synthetic dataset, was then used to evolve the rule-sets.

Formally, for each feature $x_i$, the sampled input values are constrained to the original domain ($\mathcal{I}_i = [x_{i, \min}, x_{i, \max}]$), ensuring the structure
\begin{equation}
\mathcal{D}_\mathrm{syn} = \left\{ \left( \mathbf{x}^{(j)}, \mathrm{NN}(\mathbf{x}^{(j)}) \right) \right\}_{j=1}^{N} \quad \text{s.t.} \quad x_i^{(j)} \in [x_{i, \min}, x_{i, \max}], \,\, N = |\mathcal{D}_\mathrm{orig}|,
\end{equation}

\noindent where $N = |\mathcal{D}_\mathrm{orig}|$ is the size of the original dataset.

Note that such distillation with synthetic data leads to a slightly different result than distillation with the original data. The NN still defines the target behavior that the rule-set must emulate, but the distribution of behavior is different. That is, the generated data $\mathcal{D}_\mathrm{syn}$ provides a random sampling of the decision manifold, thus forcing evolution to cover situations that may be underrepresented in the original training data.

Evolving with $\mathcal{D}_\mathrm{syn}$ tests the hypothesis that a rule-set can recover the behavior of the NN even without the original training data. Thus, Experiment~3 demonstrates the utility of NeuroRule for general post-hoc auditability.

\subsection{Experimental Parameters}
\label{exp_sec:param}
This section describes the NN and the rule-set evolution parameters, the EVOTER baseline comparison, and the single-objective vs.\ multi-objective optimization settings.

\begin{table}[!t]
\centering
\caption{Architectural specifications of the black-box NN. This network is a multi-layer perceptron (MLP) where $d$ represents the input dimensionality of the specific dataset. The design utilizes decreasing hidden layer sizes ($32$ to $16$) and a final Softmax output that is discretized into binary ($0$ or $1$) labels. The labels in the original dataset are replaced by the labels generated by the NN to create a fitness function dataset for distillation.}
\label{tab:mlp_architecture}
\small
\vspace*{-3.5ex}
\begin{tabular}{@{}llll@{}}
\toprule
\textbf{Layer} & \textbf{Type} & \textbf{Dimensions} & \textbf{Activation} \\ \midrule
Input          & Fully Connected & $d \times 32$      & ReLU                \\
Hidden 1       & Fully Connected & $32 \times 16$     & ReLU                \\
Output         & Fully Connected & $16 \times 2$      & Softmax             \\ \bottomrule
\end{tabular}
\vspace*{-2ex}
\end{table}

\begin{table}[!t]
\centering
\caption{Configuration used to train the black-box NN. To reflect standard machine learning practices, the network is trained using empirical defaults established in modern deep learning literature.}
\label{tab:surrogate_hyperparams}
\small
\vspace*{-1.5ex}
\begin{tabular}{@{}ll@{}}
\toprule
\textbf{Parameter}       & \textbf{Setting} \\ \midrule
Optimizer                & Adam ($\beta_1=0.9, \beta_2=0.999$) \\
Learning Rate            & $1 \times 10^{-3}$ \\
Loss Function            & Cross-Entropy Loss \\
Batch Size               & 32 \\
Training Epochs          & 100 \\
Weight Initialization    & He Normal \\ \bottomrule
\end{tabular}
\vspace*{-2ex}
\end{table}

\subsubsection{NN Parameters}
\label{surrogate_model}
The target NN is a black-box model from which the rule-set evolution attempts to distill knowledge. This NN is implemented as a Multi-Layer Perceptron (MLP) with a funnel-shaped architecture (Table~\ref{tab:mlp_architecture}) \cite{lecun2015deep} utilizing a standard set of training parameters (Table~\ref{tab:surrogate_hyperparams}). A simple MLP architecture promotes feature abstraction by providing a smoother, more continuous fitness landscape (optimization function to approximate) \cite{rahaman2019spectral}, resulting in a consistent fitness signal that makes the distillation process robust.


\subsubsection{Rule-set Evolution Parameters}

Rule-set evolution parameters are shown in Table~\ref{tab:ea-config-pretty}. The high Mutation Probability (0.7) and Novelty Selection Multiplier (1.5) encourage broad exploration of the high-dimensional logical space, preventing the algorithm from converging prematurely on local optima. This aggressive exploration is made viable by the model-guided fitness function; the NN provides a stable enough environment so that the algorithm can maintain high mutation rates without losing the evolutionary progress achieved in previous generations.

\subsubsection{EVOTER Baseline Comparison}
\label{baseline_config}
Rule-sets evolved with the original dataset directly (instead of predictions generated by the NN) were used as the baseline comparison for performance and explainability. This setup serves to isolate the impact of the NN, allowing for a direct comparison between rule-sets induced from the original noisy data by EVOTER and those distilled from the smoothed decision manifold of the NN. By establishing this baseline, the relative advantages of network guidance and regularization can be measured quantitatively.

\subsubsection{Single-Objective vs.\ Multi-Objective Optimization}
\label{so_mo_config}
The two optimization settings both utilize the rule-set infrastructure defined in Section~\ref{rule_set_framework}, but differ in their fitness evaluations:

\begin{itemize}
    \item Single-Objective: Selection is driven exclusively by the Accuracy objective. This function represents an unconstrained search for rule-sets with the best performance, typically resulting in complex rule-sets.
    \item Multi-Objective: Selection is driven by the NSGA-II algorithm, optimizing for both the Accuracy objective and the Conciseness objective. This function forces the evolution to discover the Pareto-optimal frontier of high-performing, high-explainability solutions.
\end{itemize}

Thus, the goal of the multi-objective approach is to find solutions that perform as well as those found by the single-objective approach, but are more concise and therefore more explainable.

\begin{table}[!t]
\centering
\caption{Parameters for rule-set evolution. The same set of parameters was used for the EVOTER baseline and NeuroRule. They are designed to favor aggressive exploration, which is likely to result in diverse, creative solutions.}
\label{tab:ea-config-pretty}
\small
\vspace*{-1.5ex}
\begin{tabular}{ll}
\toprule
\textbf{Hyperparameter}      & \textbf{Setting} \\ \midrule
Population Size              & 100              \\
Generations                  & 200              \\
Mutation Probability         & 0.7              \\
Selection Method             & Tournament       \\ 
Novelty Selection Multiplier & 1.5              \\ \bottomrule
\end{tabular}
\end{table}

\subsection{Metrics}
\label{exp_sec:metrics}
Performance and explainability are evaluated using the following three metrics:

\begin{itemize}
    \item In-Distribution (ID) Accuracy: The accuracy achieved on data within a known data distribution (the 80\%, 90\%, and 95\% partitions), calculated using test data that was withheld during training. This value measures the accuracy within specific distribution limits.
    \item Out-of-Distribution (OOD) Accuracy: The accuracy achieved on the data held out in training the NN (the remaining 20\%, 10\%, and 5\% partitions). This value is the primary measure of generalization capability.
    \item Conciseness: Quantified as the total Condition Count (i.e.\ the number of atomic conditions) within the rule-set. A lower condition count indicates a more explainable model.
\end{itemize}

These metrics allow for well-rounded comparisons of the different approaches evaluated in this paper, as will be described in the next section.

\section{Results}

This section evaluates the performance of evolved rule-set models across the three main contributions: distilling a black-box NN into an explainable rule-set representation (Section~\ref{sec:surrogate}), enhancing explainability through conciseness (Section~\ref{sec:multi}), and distilling a black-box network into a rule-set representation with synthetic data (Section~\ref{sec:synthetic}). 

\subsection{Experiment~1: Distilling Black-Box NNs into Rule-Sets}
\label{sec:surrogate}
\begin{figure}[!t]
\centering

\begin{minipage}{1.0\textwidth}
\includegraphics[width=0.49\linewidth]{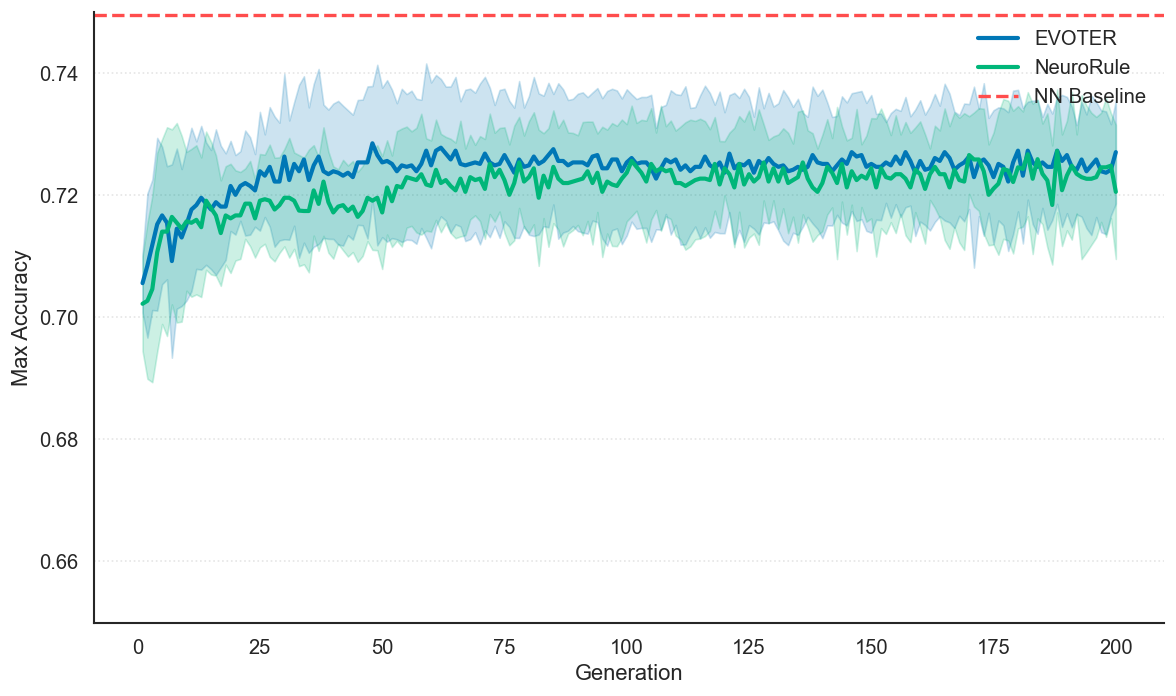}
\includegraphics[width=0.49\linewidth]{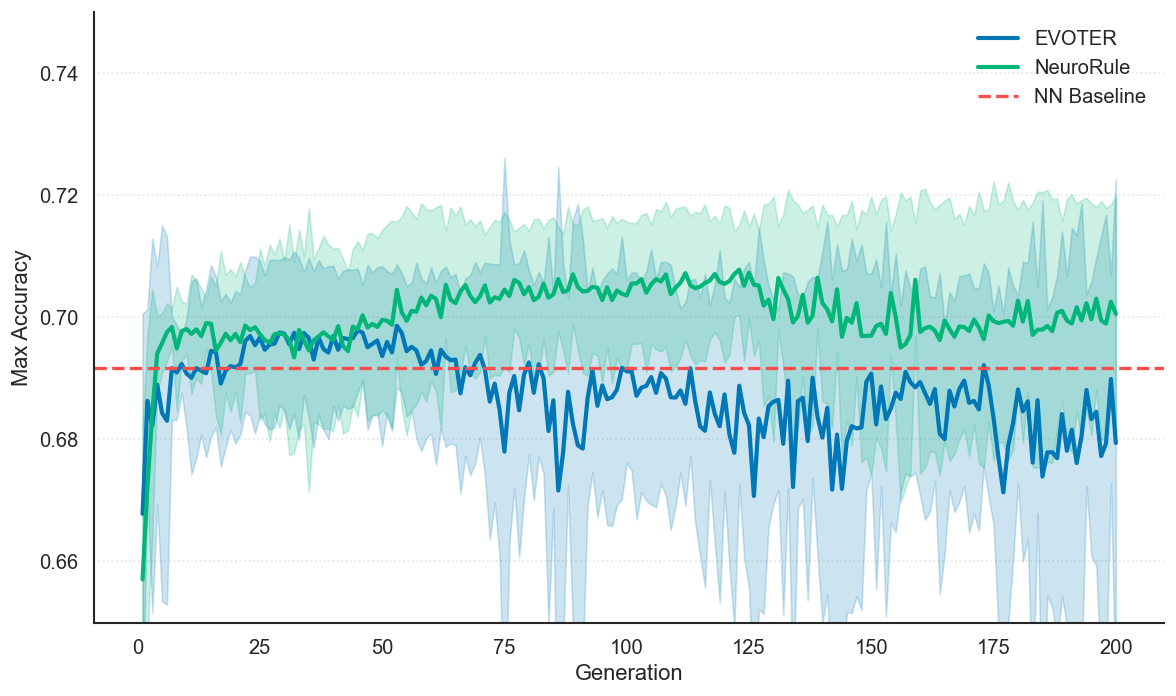}\\[-1ex]
\centerline{\small 80\% ID/OOD split: In-Distribution (Left) vs. Out-of-Distribution (Right)}
\end{minipage}

\vspace{2ex}

\begin{minipage}{1.0\textwidth}
\includegraphics[width=0.49\linewidth]{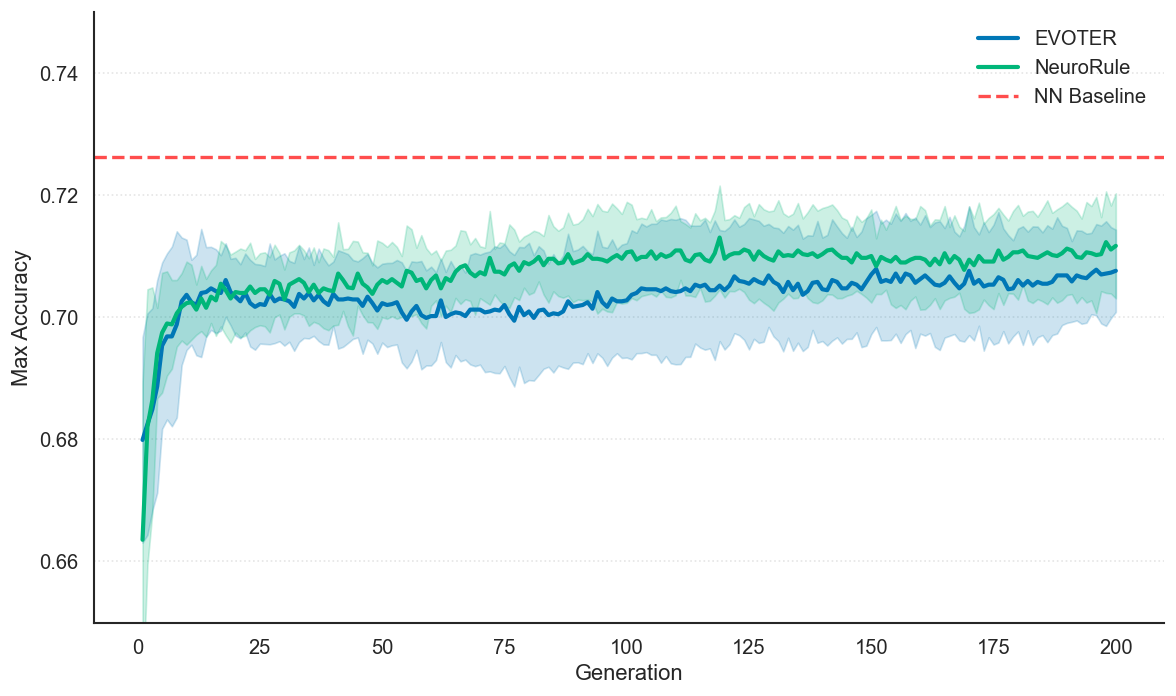}
\includegraphics[width=0.49\linewidth]{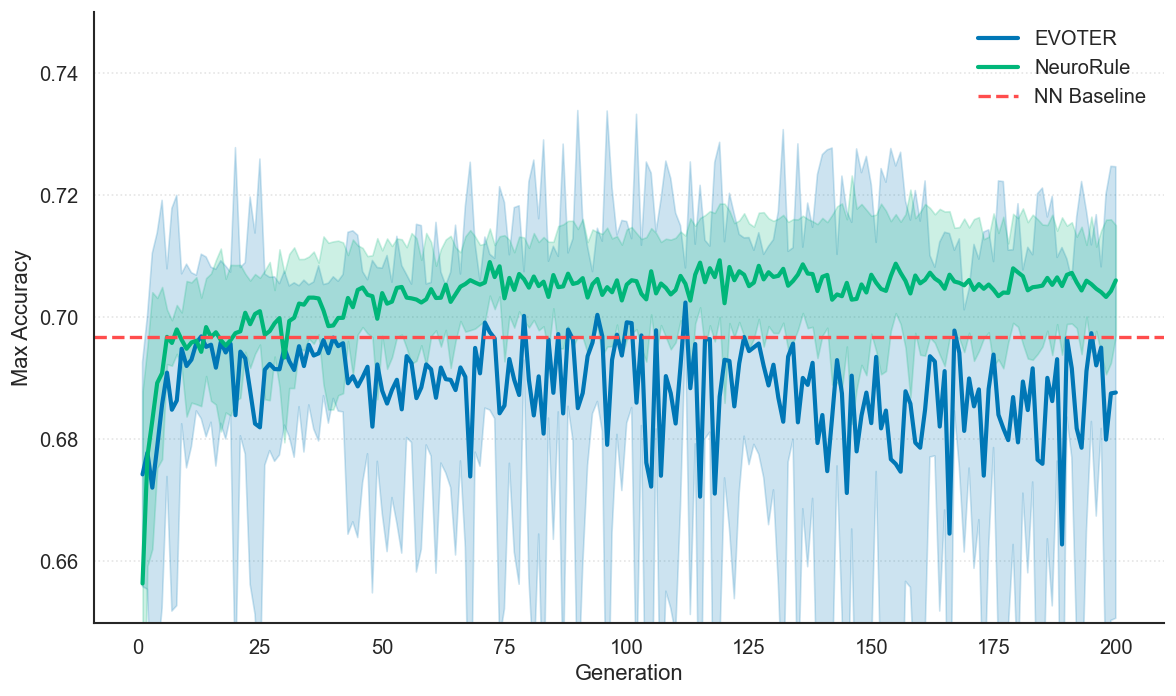}\\[-1ex]
\centerline{\small 90\% ID/OOD split: In-Distribution (Left) vs. Out-of-Distribution (Right)}
\end{minipage}

\vspace{2ex}

\begin{minipage}{1.0\textwidth}
\includegraphics[width=0.49\linewidth]{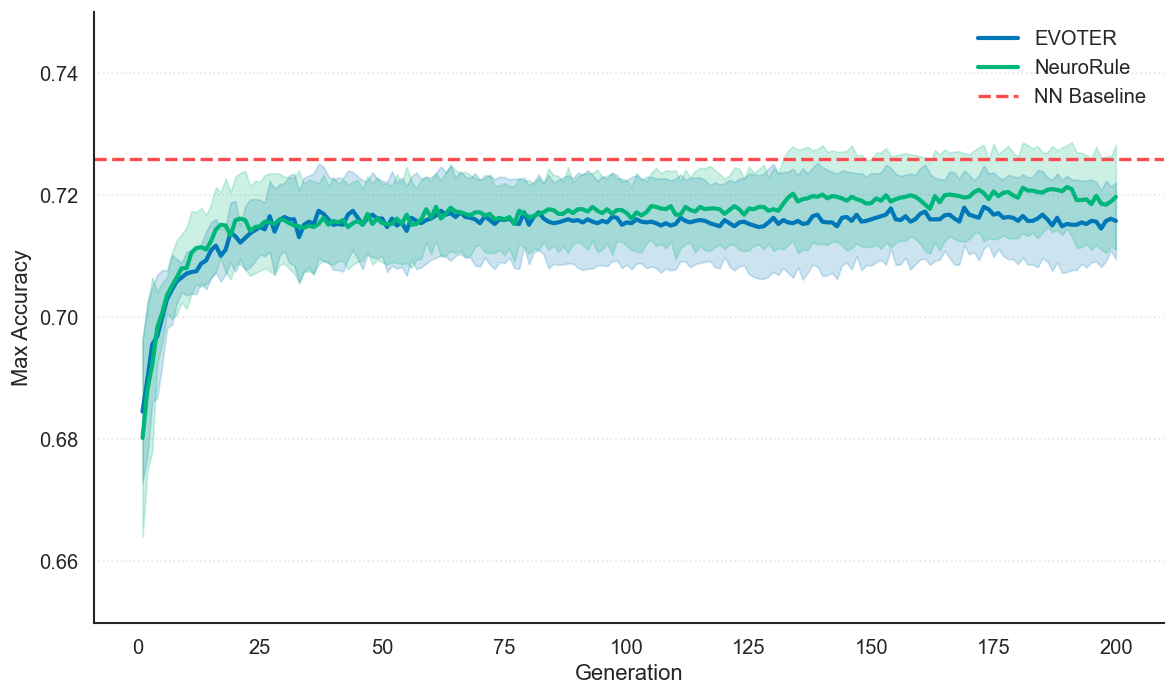}
\includegraphics[width=0.49\linewidth]{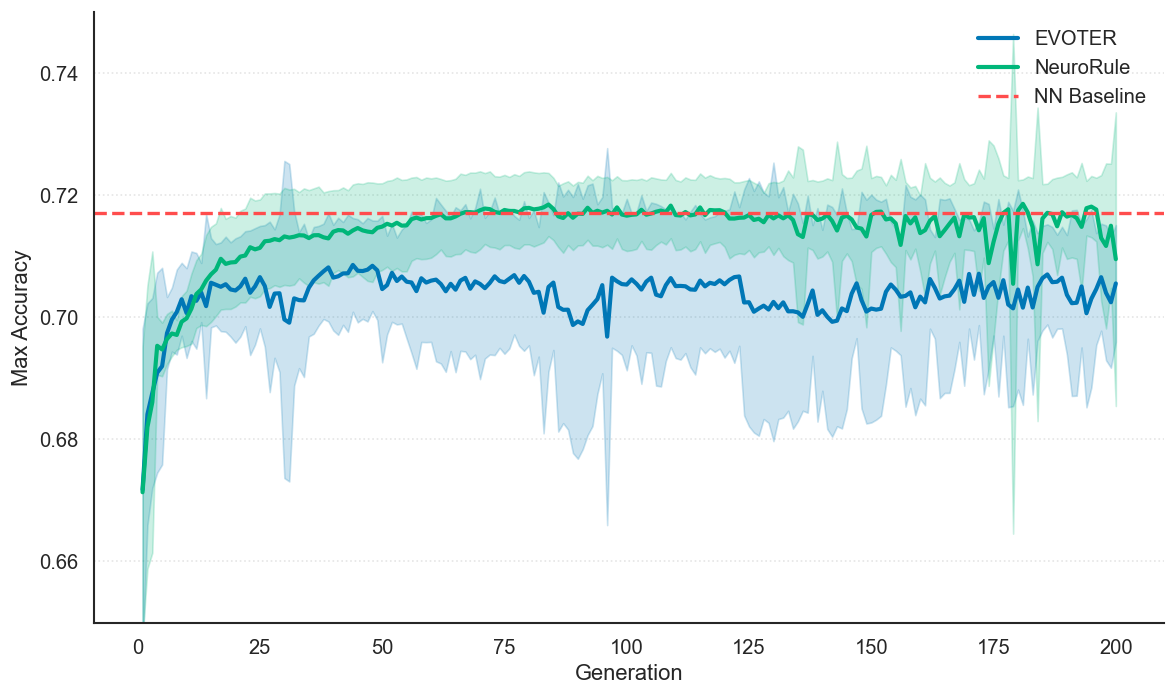}\\[-1ex]
\centerline{\small 95\% ID/OOD split: In-Distribution (Left) vs. Out-of-Distribution (Right)}
\end{minipage}

\caption{Learning curves for EVOTER and NeuroRule compared to the NN accuracy at different ID/OOD splits on the Diabetes dataset. The curves are averages over 10 runs, and the bands indicate standard deviation. The ID graphs (left) demonstrate that NeuroRule rule-sets (green) generally achieved higher accuracy than the EVOTER baseline (blue). This result suggests that the NN helps in avoiding overfitting to noisy samples. Both rule-set approaches initially exhibit a period of rapid improvement and then plateau at a level slightly below the NN (red), suggesting that the rule-sets have a slight performance cost. With OOD (right), NeuroRule (and EVOTER in other datasets) actually performs better than the NN. This result suggests that the rule-sets discover underlying principles of data rather than overfit to noise. Results for the Breast Cancer and Heart Disease datasets are available in Figures~\ref{fig:experiment1_bc} and~\ref{fig:experiment1_heart}.} 
\label{fig:experiment1_diabetes}
\end{figure}

Distillation with NeuroRule was evaluated both with seen in-distribution data and unseen out-of-distribution data. The results are summarized in Table~\ref{tab:single_accuracies}, and learning curves shown in  Figures~\ref{fig:experiment1_diabetes},~\ref{fig:experiment1_bc}, and \ref{fig:experiment1_heart}.

\subsubsection{In-Distribution Performance}
\label{subsec:id}
The evolved rule-sets were evaluated on their ability to serve as accurate approximations of NNs. In most cases, the NeuroRule rule-sets plateaued at an accuracy level slightly below that of the NN. This accuracy ceiling suggests a fundamental difference in model capacity. The NNs, characterized by a large number of parameters, have a high variance and therefore are likely to achieve high accuracy by effectively capturing the nuances of the training data. In contrast, the rule-sets are more concise and therefore have a low variance, which may prevent them from emulating the NN on these nuances. This result suggests that while the black-box NNs are optimized for fitting known data distributions, the rule-sets can at most approximate their behavior. On the other hand, the rule-sets have a high bias; this property may make them better at capturing principles in the data, and therefore allow them to generalize better. This hypothesis is further supported by the OOD performance (Section~\ref{subsec:ood}).

Compared to the EVOTER baseline, NeuroRule significantly improved accuracy in most cases. This result suggests a unique distillation effect: whereas EVOTER is likely to converge on brittle, overly specific rules to represent the sometimes noisy original labels, the NN may smooth out the label noise. It can thus provide NeuroRule with a more continuous fitness landscape and guide it toward the principles underlying the data. Again, the OOD experiments support this hypothesis.

\begin{table}[!t]
\centering
\caption{Experiment~1: Accuracy of NN, EVOTER, and NeuroRule models on ID and OOD data at different ID/OOD splits on the three clinical benchmarks.
The numbers are averages of 10 runs; the bolded values identify the best performance in each category. In ID, the NN is the most accurate, suggesting that NeuroRule models indeed evolve to approximate the NN. However, NeuroRule is usually more accurate than EVOTER, suggesting that the NN provides more regularized targets for rule evolution than the original data. In OOD, rule-set networks in general and NeuroRule in particular are more accurate than the NN, suggesting that they have learned the underlying principles of the data, allowing them to extrapolate better. The learning curves with standard deviations are shown in Figures~\ref{fig:experiment1_diabetes},~\ref{fig:experiment1_bc}, and~\ref{fig:experiment1_heart}.}
\label{tab:single_accuracies}
\vspace*{-1ex}
\centering
{\small
Breast Cancer Dataset Performance\\[1ex]
\begin{tabular}{lrrr|rrr}
\hline
             & \multicolumn{3}{c}{In-Distribution (ID)} & \multicolumn{3}{c}{Out-of-Distribution (OOD)} \\ \cline{2-7} 
ID/OOD split & NN    & EVOTER & NeuroRule      & NN    & EVOTER & NeuroRule      \\ \hline
80\%         & \textbf{1.000} & 0.944  & 0.947 & 0.759 & 0.892  & \textbf{0.909} \\
90\%         & \textbf{0.979} & 0.919  & 0.938 & 0.887 & 0.908  & \textbf{0.938} \\
95\%         & \textbf{0.925} & 0.875  & 0.899 & \textbf{0.987} & 0.927  & 0.933 \\ \hline \\
\end{tabular}}

\centering
{\small
Heart Disease Dataset Performance\\[1ex]
\begin{tabular}{lrrr|rrr}
\hline
             & \multicolumn{3}{c}{In-Distribution (ID)} & \multicolumn{3}{c}{Out-of-Distribution (OOD)} \\ \cline{2-7}
ID/OOD split & NN    & EVOTER         & NeuroRule      & NN    & EVOTER         & NeuroRule      \\ \hline
80\%         & 0.888 & 0.898          & \textbf{0.900} & 0.777 & \textbf{0.842} & 0.829          \\
90\%         & \textbf{0.880} & 0.868          & 0.876 & 0.744 & 0.815          & \textbf{0.844} \\
95\%         & \textbf{0.902} & 0.891 & 0.886          & 0.719 & 0.832          & \textbf{0.857} \\ \hline \\
\end{tabular}}

\centering
{\small
Diabetes Dataset Performance\\[1ex]

\begin{tabular}{lrrr|rrr}
\hline
             & \multicolumn{3}{c}{In-Distribution (ID)} & \multicolumn{3}{c}{Out-of-Distribution (OOD)} \\ \cline{2-7}
ID/OOD split & NN    & EVOTER         & NeuroRule      & NN    & EVOTER & NeuroRule      \\ \hline
80\%         & \textbf{0.749} & 0.727 & 0.720          & 0.692 & 0.679  & \textbf{0.700} \\
90\%         & \textbf{0.726} & 0.708          & 0.712 & 0.697 & 0.688  & \textbf{0.706} \\
95\%         & \textbf{0.726} & 0.716          & 0.721 & \textbf{0.717} & 0.705  & 0.708 \\ \hline  \\
\end{tabular}}
\end{table}

\subsubsection{Out-of-Distribution Performance}
\label{subsec:ood}
Next, the generalization capabilities of evolved rule-sets were evaluated. Both the NNs and the evolved rule-sets were tested on the OOD data that was not only previously unseen but also outside the range of the ID variable values. This evaluation tested the hypothesis that the rule-sets may extrapolate better than the black-box NNs.

The results indeed supported this hypothesis: NeuroRule performed the best, and EVOTER was still better than the NNs.
he rule-sets' high bias and low variance make them less prone to modeling noise in the data. Instead, the evolutionary process likely abstracts the underlying principles, i.e.\ the logical relationships that persist beyond the specific samples.

%

\subsection{Experiment~2: Enhancing Explainability Through Conciseness}
\label{sec:multi}
The second experiment focused on a central theme in explainable machine learning: the trade-off between performance and explainability. This focus required a multi-objective evolutionary approach, jointly optimizing for both accuracy and conciseness, i.e., the size of the rule-set model. As expected, the resulting solutions are more concise, but surprisingly, often also more accurate, and more consistently so.

\subsubsection{Effect on Explainability}
\label{subsec:functional_similarity}

Table~\ref{tab:conciseness} shows the average condition count of the final highest performing (by in-distribution accuracy) rule-set models, averaged across 10 independent experimental runs. The results demonstrate that adding the conciseness objective clearly results in smaller models in both EVOTER and NeuroRule. Additionally, NeuroRule's average condition count is consistently lower than EVOTER's. 
Such concise rule-sets are easier to verify with domain experts and can provide intuitive insight into the underlying domain mechanisms. This capability is critical for real-world deployment scenarios where explainability is just as important as performance. 

\begin{table}[!t]
\centering
\caption{Average Condition Count for EVOTER and NeuroRule Rule-Sets with Single-objective and Multi-objective Optimization. In each run, a representative candidate as selected by maximum in-distribution accuracy. The numbers are averages across 10 independent experimental runs; the bold values identify the best performance in each category. EVOTER and NeuroRule consistently produce smaller rule sets, with NeuroRule in particular finding very concise models. Such models are easier to verify with domain experts and can provide intuitive insight into the underlying distilled model. This feature is critical for real-world scenarios where explainability and performance are both very important.}
\label{tab:conciseness}
\vspace*{-1ex}
\centering
{\small
Breast Cancer Dataset\\[1ex]
\begin{tabular}{lrr|rr}
\hline 
ID/OOD split & EVOTER Single & EVOTER Multi & NeuroRule Single & NeuroRule Multi \\ \hline
80\% & 47.500 & \textbf{3.0000} & 38.625 & \textbf{1.5000} \\
90\% & 52.800 & \textbf{3.5000} & 73.333 & \textbf{1.9000} \\
95\% & 150.40 & \textbf{3.3333} & 64.400 & \textbf{2.5000} \\ \hline \\
\end{tabular}}

\centering
{\small
Heart Disease Dataset\\[1ex]
\begin{tabular}{lrr|rr}
\hline 
ID/OOD split & EVOTER Single & EVOTER Multi & NeuroRule Single & NeuroRule Multi \\ \hline
80\% & 61.333 & \textbf{11.00} &  132.38 & \textbf{7.0000} \\
90\% & 72.666 & \textbf{7.8333} & 75.000 & \textbf{6.5000} \\
95\% & 128.50 & \textbf{4.8571} & 117.00 & \textbf{5.1000} \\ \hline \\
\end{tabular}}

\centering
{\small
Diabetes Dataset\\[1ex]
\begin{tabular}{lrr|rr}
\hline 
ID/OOD split & EVOTER Single & EVOTER Multi & NeuroRule Single & NeuroRule Multi \\ \hline
80\% & 150.60 & \textbf{10.000} & 102.89 & \textbf{5.8000} \\
90\% & 181.22 & \textbf{7.5000} & 101.22 & \textbf{13.900} \\
95\% & 177.00 & \textbf{8.000}  & 196.83 & \textbf{10.100} \\ \hline \\
\end{tabular}}
\end{table}



\subsubsection{Effect on Performance}
\label{subsec:accuracy_conciseness}
One concern with optimizing for conciseness is that the accuracy of the resulting rule-sets may suffer. Surprisingly, the opposite effect was observed: the multi-objective runs actually performed better in several cases (Table~\ref{tab:appendix_multi}). By limiting the number of conditions in the model, the evolutionary search is forced to prioritize generalizable logic over covering specific examples. Whereas single-objective evolution is free to overfit by creating complex rules for outliers, multi-objective evolution is unlikely to do so. The resulting models focus on more general features, leading to better performance.

\begin{table}[!t]
\centering
\caption{Experiment~2: Accuracy of EVOTER and NeuroRule with multi-objective evolution of accuracy and conciseness. The NN baseline is the same as in Table~\ref{tab:single_accuracies}. In each run, a representative candidate as selected by maximum in-distribution accuracy. The numbers are averages across 10 independent experimental runs; the bold values identify the best performance in each category. Surprisingly, accuracy did not suffer; the resulting rule-sets generally performed better than those evolved for accuracy only (Table~\ref{tab:single_accuracies}), very close to the NN accuracy. The learning curves with standard deviations are shown in Figures~\ref{fig:diabetes_multi_curves},~\ref{fig:bc_multi_curves}, and-\ref{fig:heart_multi_curves}.}
\label{tab:appendix_multi}
\vspace*{-1ex}
\centering
{\small
Breast Cancer Dataset\\[1ex]
\begin{tabular}{lrrr|rrr}
\hline
             & \multicolumn{3}{c}{In-Distribution (ID)} & \multicolumn{3}{c}{Out-of-Distribution (OOD)} \\ \cline{2-7} 
ID/OOD split & NN    & EVOTER & NeuroRule  & NN    & EVOTER & NeuroRule      \\ \hline
80\%         & \textbf{1.000} & 0.953  & 0.947 & 0.759 & 0.922      & \textbf{0.925} \\
90\%         & \textbf{0.979} & 0.957  & 0.961 & 0.887 & 0.932      & \textbf{0.937} \\
95\%         & \textbf{0.925} & 0.922  & 0.918 & \textbf{0.987} & 0.937      & 0.930 \\ \hline  \\
\end{tabular}}

\centering
{\small
Heart Disease Dataset\\[1ex]
\begin{tabular}{lrrr|rrr}
\hline
             & \multicolumn{3}{c}{In-Distribution (ID)} & \multicolumn{3}{c}{Out-of-Distribution (OOD)} \\ \cline{2-7}
ID/OOD split & NN    & EVOTER         & NeuroRule      & NN    & EVOTER & NeuroRule      \\ \hline
80\%         & 0.888 & 0.887          & \textbf{0.893} & 0.777 & \textbf{0.849} & 0.848 \\
90\%         & \textbf{0.880} & 0.869          & 0.868          & 0.744 & \textbf{0.846} & 0.831 \\
95\%         & \textbf{0.902} & 0.881          & 0.885          & 0.719 & 0.839          & \textbf{0.844} \\ \hline  \\
\end{tabular}}

\centering
{\small
Diabetes Dataset\\[1ex]
\begin{tabular}{lrrr|rrr}
\hline
             & \multicolumn{3}{c}{In-Distribution (ID)} & \multicolumn{3}{c}{Out-of-Distribution (OOD)} \\ \cline{2-7}
ID/OOD split & NN    & EVOTER         & NeuroRule      & NN    & EVOTER & NeuroRule      \\ \hline
80\%         & \textbf{0.749} & 0.727          & 0.724          & 0.692 & \textbf{0.709} & 0.698 \\
90\%         & \textbf{0.726} & 0.708          & 0.716          & 0.697 & \textbf{0.709} & 0.707 \\
95\%         & \textbf{0.726} & 0.716          & 0.713          & \textbf{0.717} & 0.709          & 0.712 \\ \hline \\
\end{tabular}}
\end{table}

\subsubsection{Accuracy vs.\ Conciseness tradeoff}
\begin{figure}[!t]
\centering
\begin{minipage}{1.0\textwidth}
\includegraphics[width=0.49\linewidth]{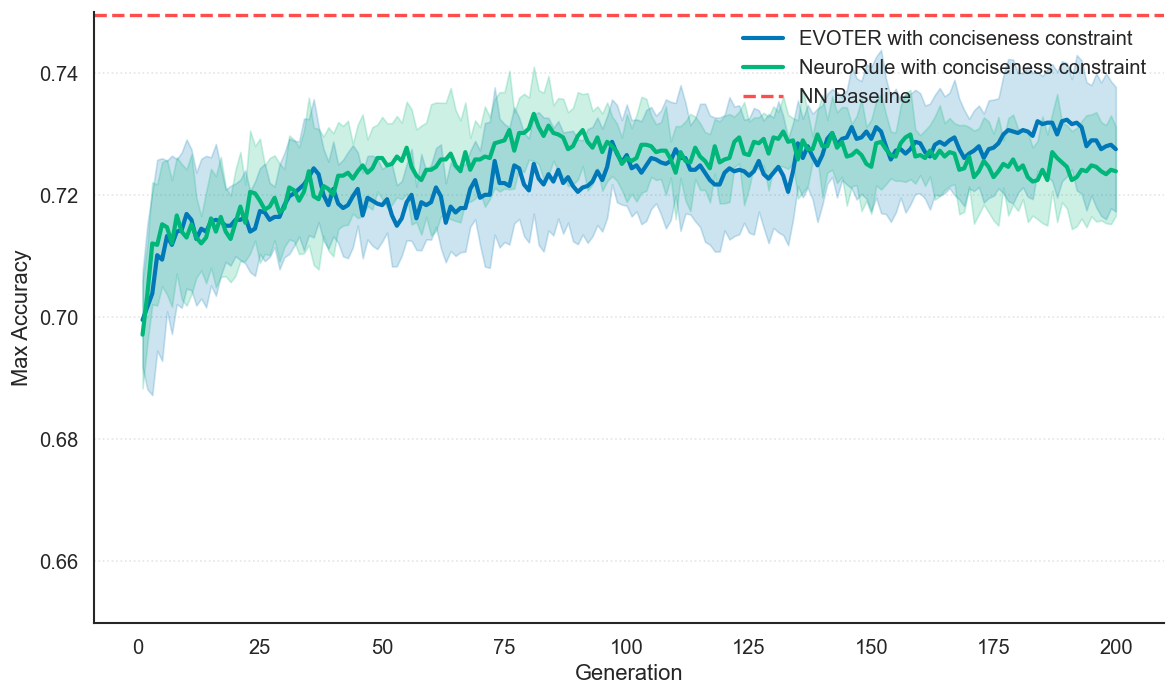}
\hfill
\includegraphics[width=0.49\linewidth]{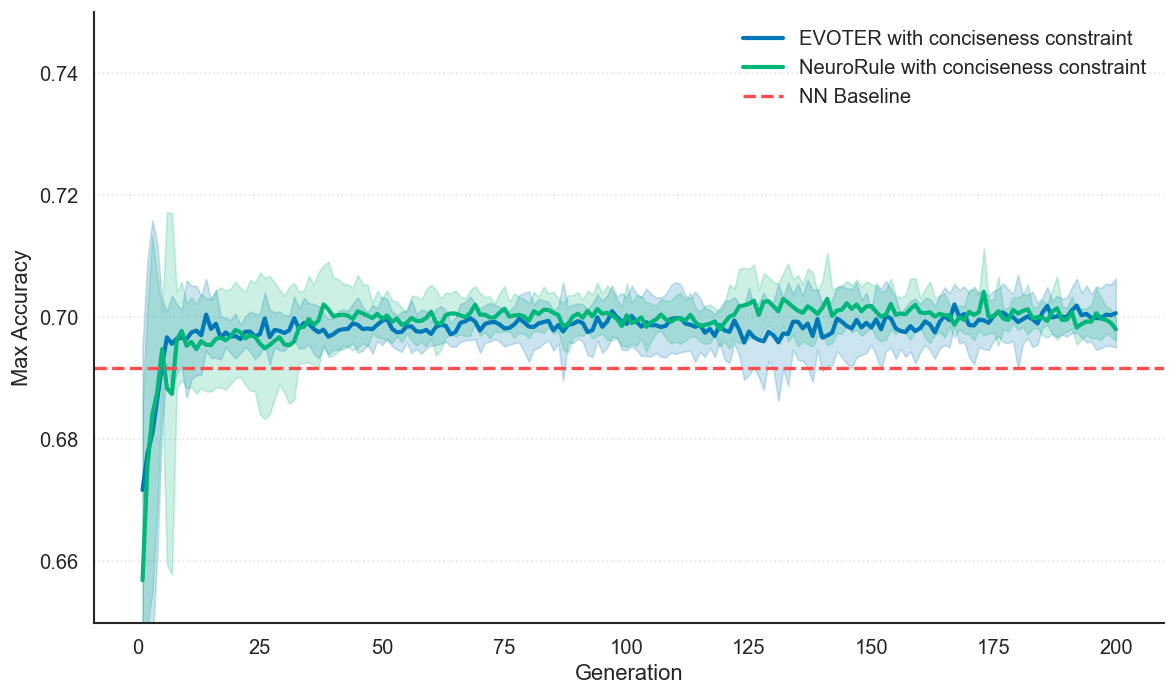}\\[-1ex]
\centerline{\small 80\% ID/OOD split: In-Distribution (Left) vs. Out-of-Distribution (Right)}
\end{minipage}\\[3ex]
\begin{minipage}{1.0\textwidth}
\includegraphics[width=0.49\linewidth]{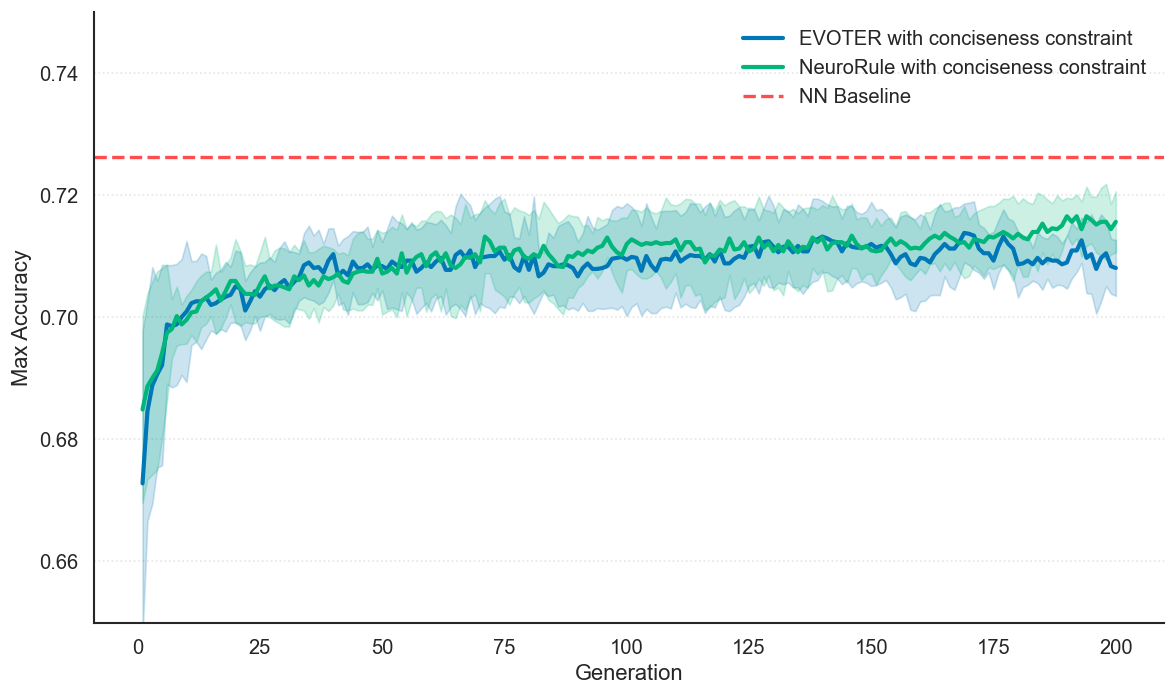}
\hfill
\includegraphics[width=0.49\linewidth]{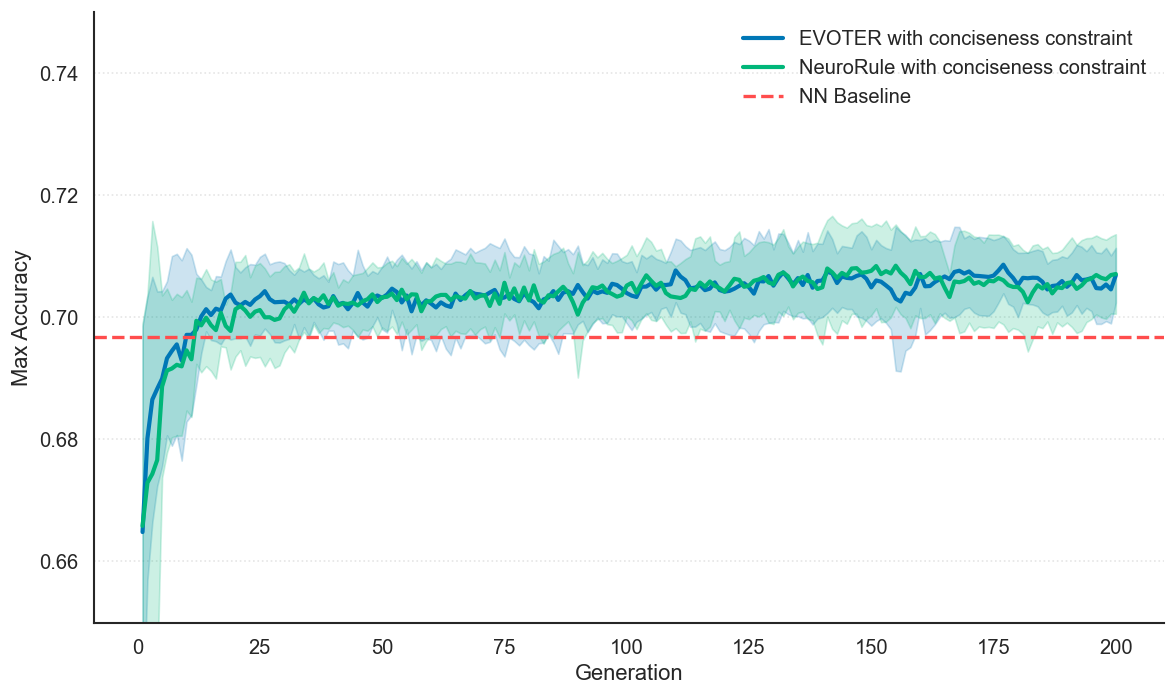}\\[-1ex]
\centerline{\small 90\% ID/OOD split: In-Distribution (Left) vs. Out-of-Distribution (Right)}
\end{minipage}\\[3ex]
\vspace{2ex}
\begin{minipage}{1.0\textwidth}
\includegraphics[width=0.49\linewidth]{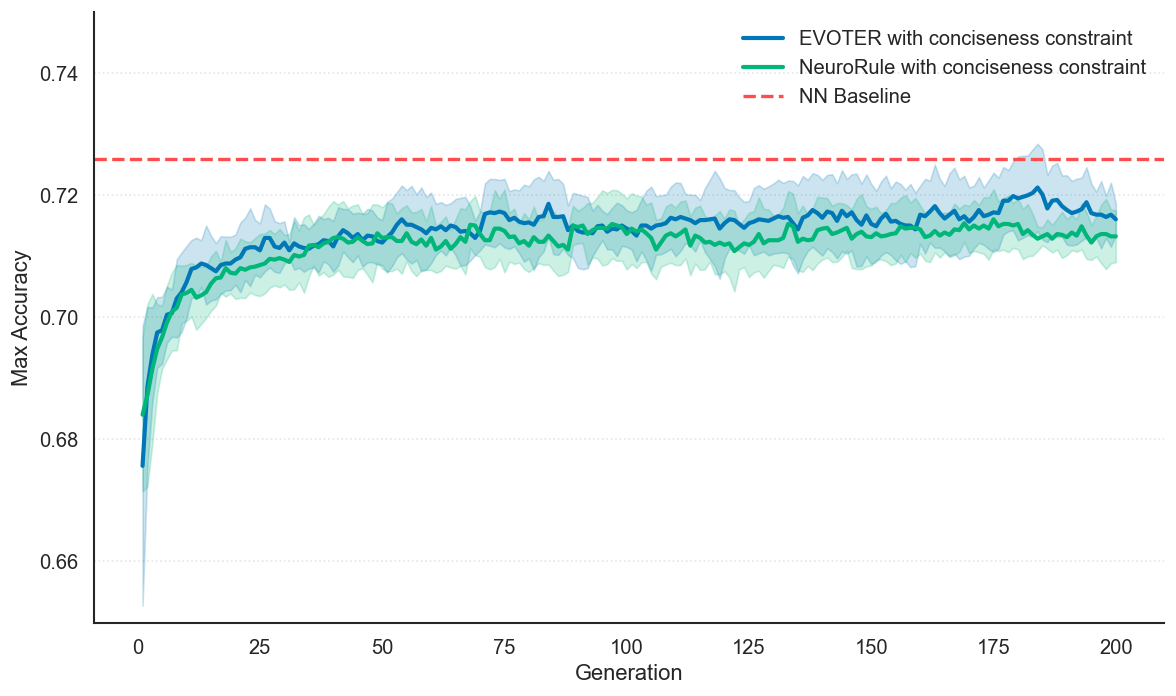}
\hfill
\includegraphics[width=0.49\linewidth]{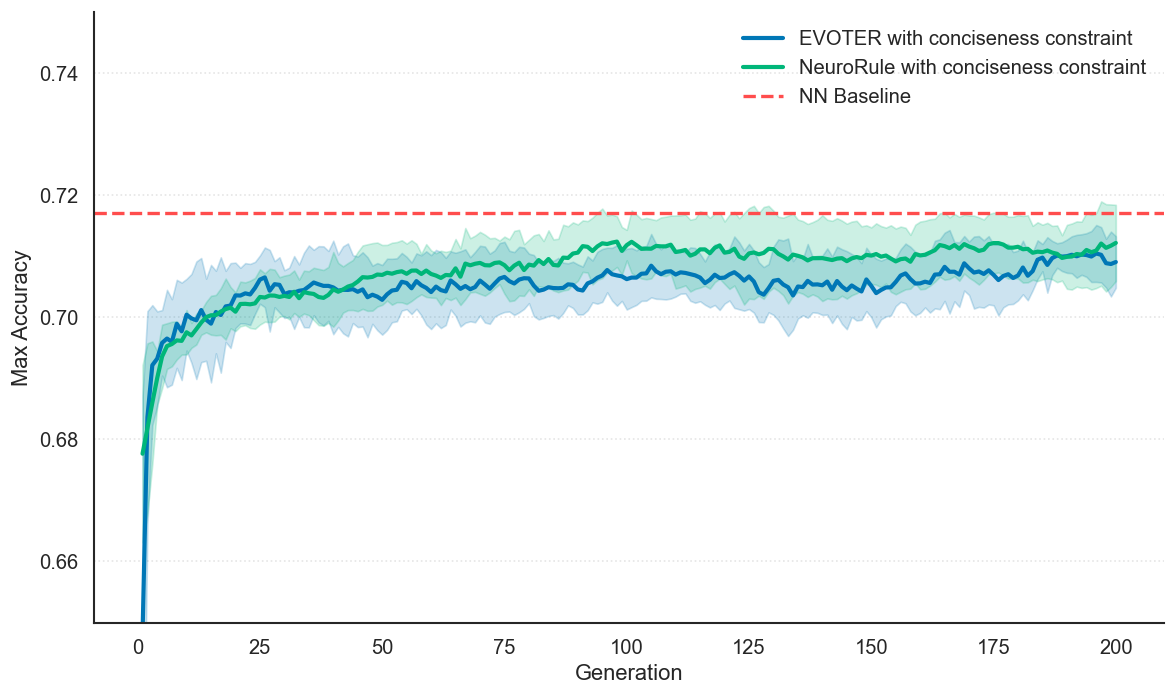}\\[-1ex]
\centerline{\small 95\% ID/OOD split: In-Distribution (Left) vs. Out-of-Distribution (Right)}
\end{minipage}
\vspace*{-1ex}
\caption{Learning curves for EVOTER and NeuroRule compared to NN accuracy at different ID/OOD splits on the Diabetes dataset under multi-objective optimization of accuracy and conciseness. The curves are averages over 10 runs, and the bands indicate standard deviation. Compared to single-objective optimization, accuracy has improved significantly, suggesting that conciseness encourages rule-set evolution to find representations that are more principled and generalize better. Also, the standard deviation is smaller, presumably because search focuses on a more limited set of solutions. The results for Breast Cancer and Heart Disease datasets are in Figures~\ref{fig:bc_multi_curves} and~\ref{fig:heart_multi_curves}.}
\label{fig:diabetes_multi_curves}
\end{figure}

Interestingly, the solutions found by multi-objective evolution also performed more consistently, as can be seen in the learning curves of Figures~\ref{fig:diabetes_multi_curves},~\ref{fig:bc_multi_curves}, and~\ref{fig:heart_multi_curves}. The conciseness constraint essentially acts as a anchor by limiting the search space. Restricting the evolutionary process to smaller and more efficient rule-sets results in a more predictable optimization path and consequently, narrower standard deviation.

\begin{figure}[!t]
\centering
\begin{minipage}{1.0\textwidth}
\includegraphics[width=0.49\linewidth]{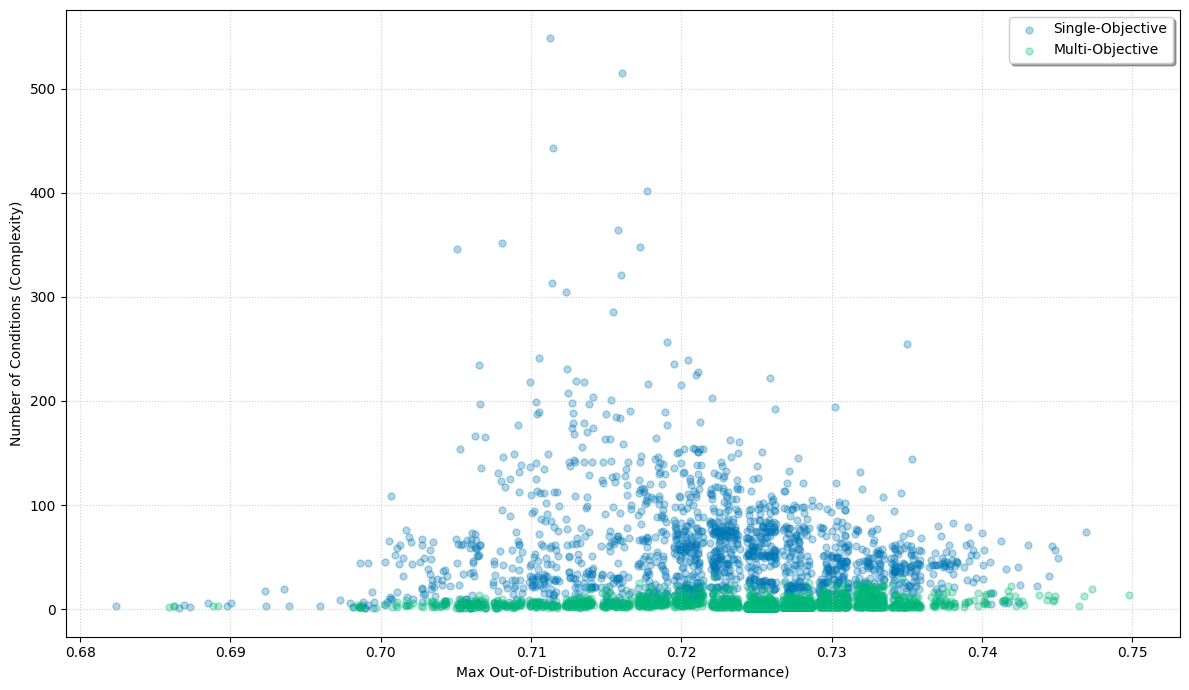}
\hfill
\includegraphics[width=0.49\linewidth]{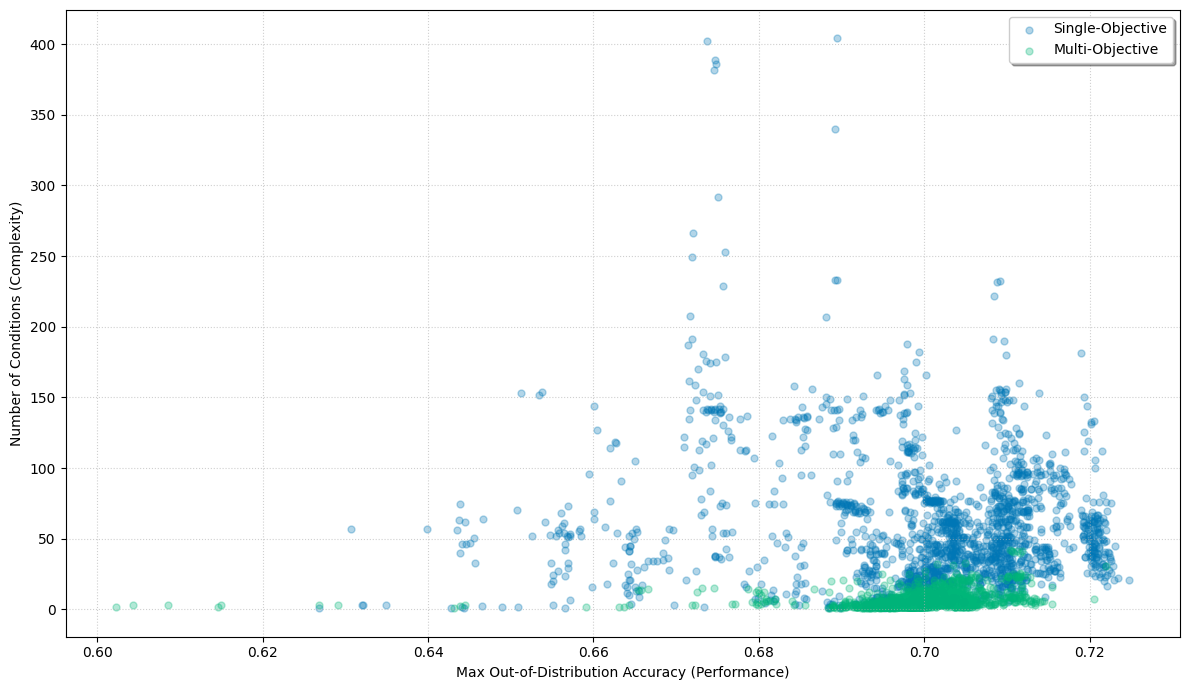}\\[-1ex]
\centerline{\small 80\% ID/OOD split: In-Distribution (Left) vs. Out-of-Distribution (Right)}
\end{minipage}\\[2ex]
\begin{minipage}{1.0\textwidth}
\includegraphics[width=0.49\linewidth]{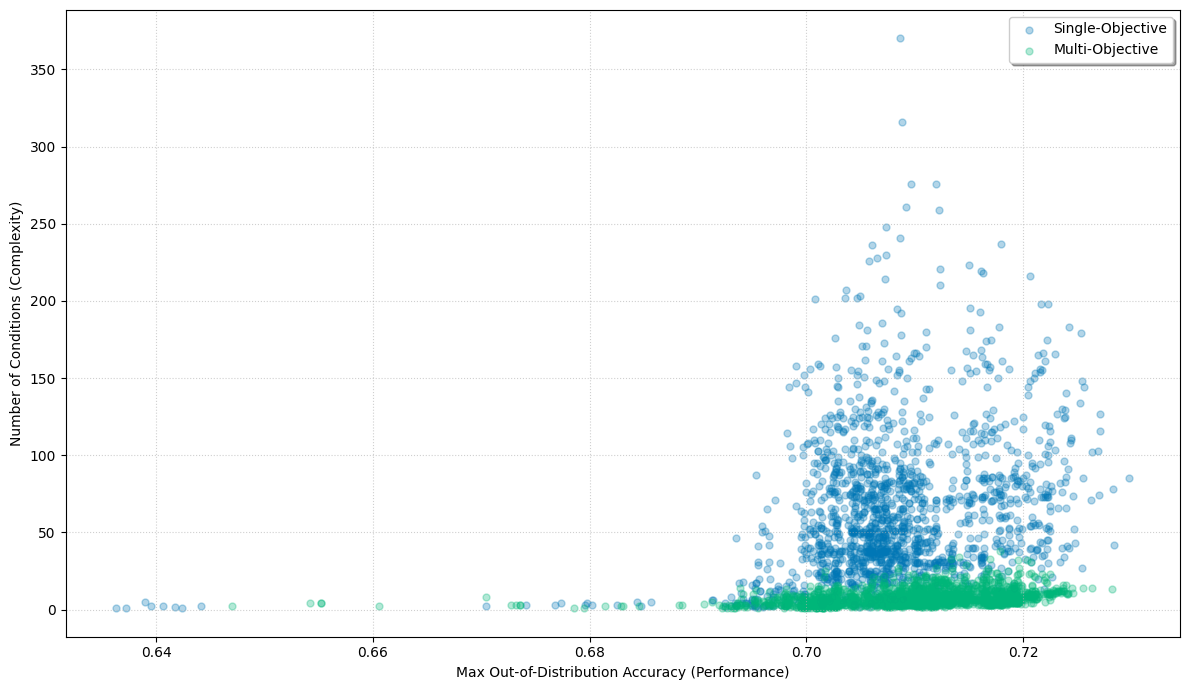}
\hfill
\includegraphics[width=0.49\linewidth]{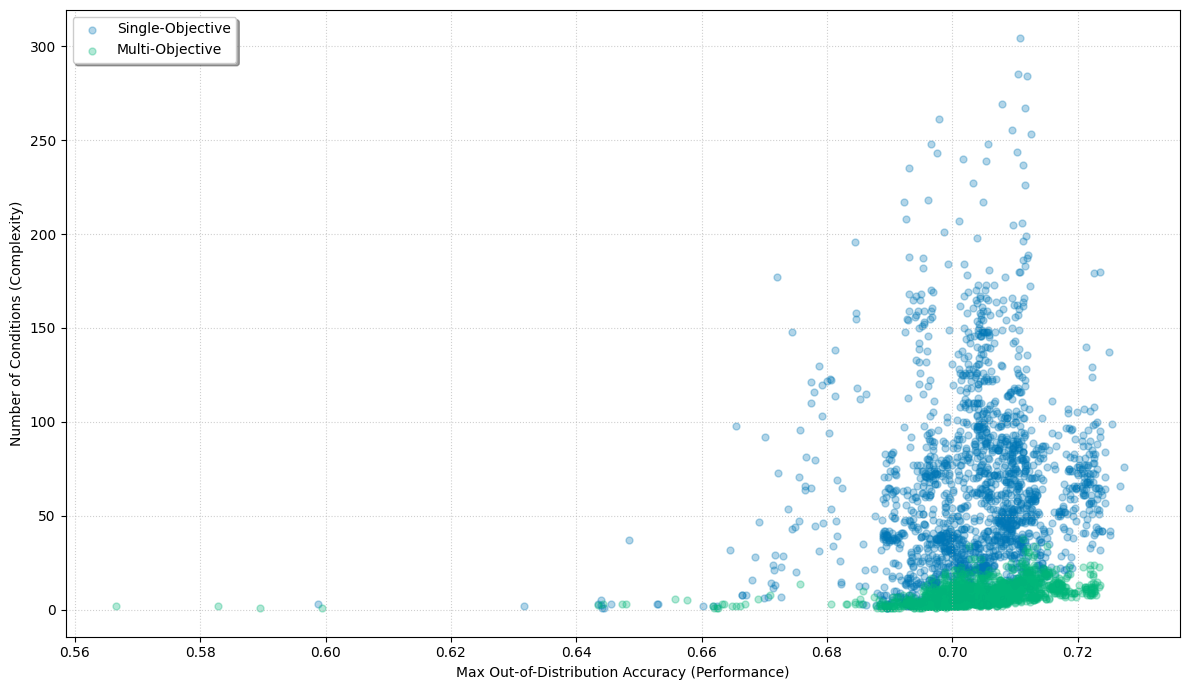}\\[-1ex]
\centerline{\small 90\% ID/OOD split: In-Distribution (Left) vs. Out-of-Distribution (Right)}
\end{minipage}\\[2ex]
\begin{minipage}{1.0\textwidth}
\includegraphics[width=0.49\linewidth]{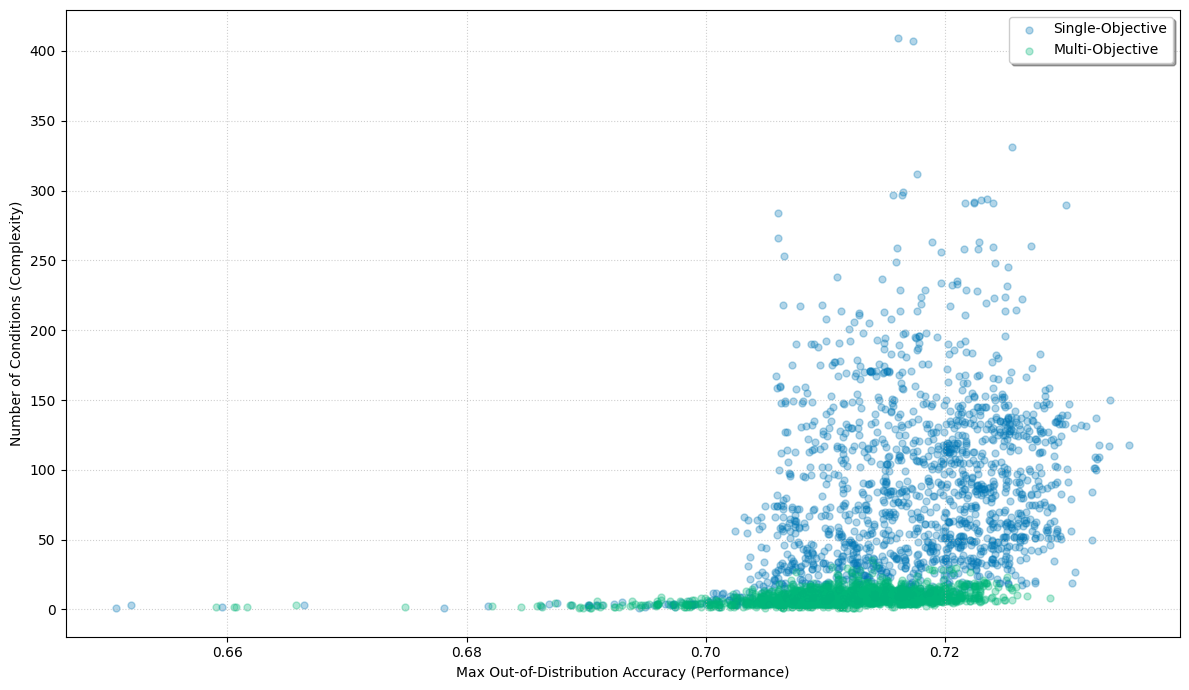}
\hfill
\includegraphics[width=0.49\linewidth]{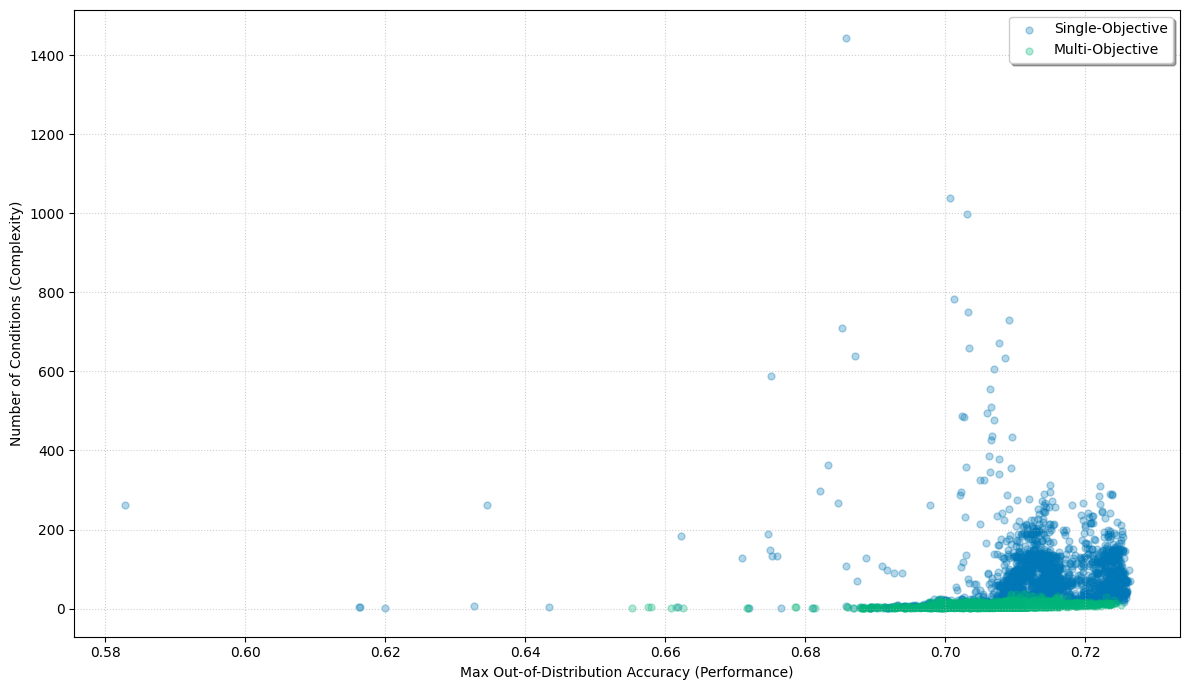}\\[-1ex]
\centerline{\small 95\% ID/OOD split: In-Distribution (Left) vs. Out-of-Distribution (Right)}
\end{minipage}
\vspace*{-1ex}
\caption{Accuracy vs.\ conciseness in single and multi-objective NeuroRule evolutionary runs at different ID/OOD splits on the Diabetes dataset. The model accuracy is on the $x$-axis and conciseness on the $y$-axis; the scatter plot shows all individuals in the final population of all 10 single-objective (blue) and multi-objective (green) runs. The multi-objective solutions are concentrated at the bottom, indicating that they have much lower complexity than single-objective solutions. Notably, the most accurate solutions (furthest to the right) are also often among the most concise (low). Results for the Breast Cancer and Heart Disease datasets are in Figures~\ref{fig:bc_pareto_comparisons} and~\ref{fig:heart_pareto_comparisons}.}
\label{fig:diabetes_pareto_comparisons}
\end{figure}

The effects on explainability and performance can be made concrete by mapping the final populations of the 10 evolutionary runs into the accuracy vs.\ conciseness space (Figures~\ref{fig:diabetes_pareto_comparisons},~\ref{fig:bc_pareto_comparisons}, and~\ref{fig:heart_pareto_comparisons}). The multi-objective solutions are more concise, and in many cases, the ones with the highest performance also had very close to lowest complexity.

%

\subsection{Experiment~3: Distilling with Synthetic Data}
\label{sec:synthetic}
The third experiment focused on evaluating the NeuroRule approach when the original NN training data is not available. As described in Section~\ref{sec:sampling_strat}, random samples within the input space were generated, given to the NN, and its outputs recorded. NeuroRule was then used to evolve a rule-set with this dataset. Its performance was then evaluated with the same ID and OOD datasets as before.

\begin{figure}[!t]
\centering
\begin{minipage}{1.0\textwidth}
\includegraphics[width=0.49\linewidth]{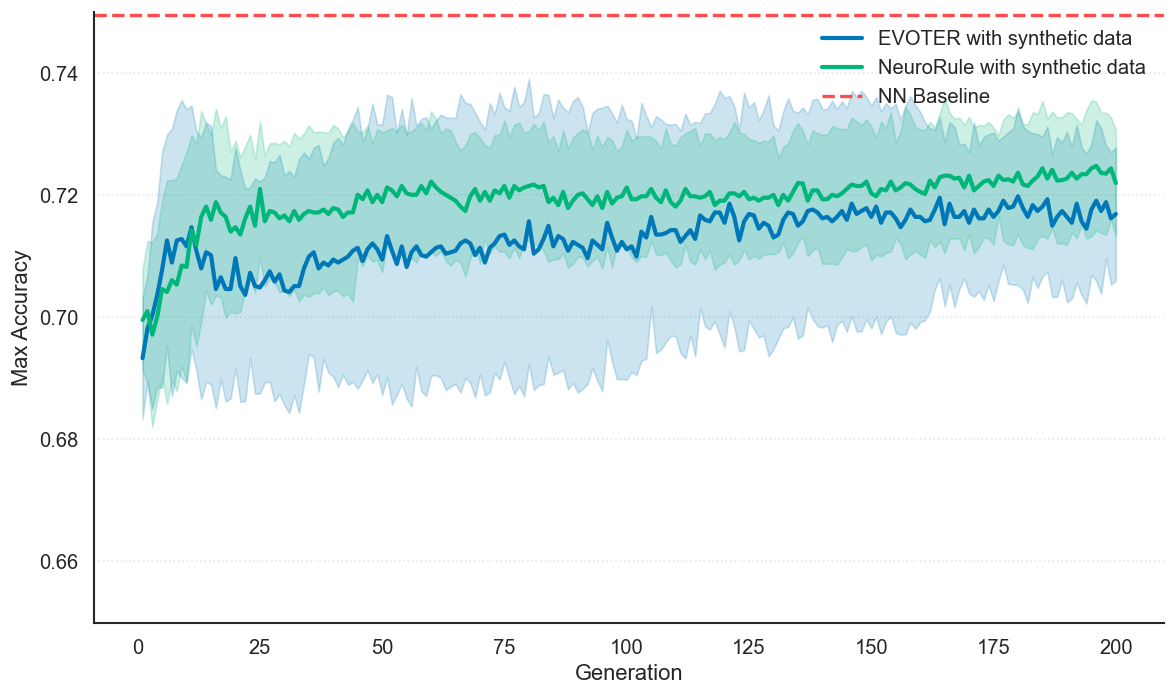}
\hfill
\includegraphics[width=0.49\linewidth]{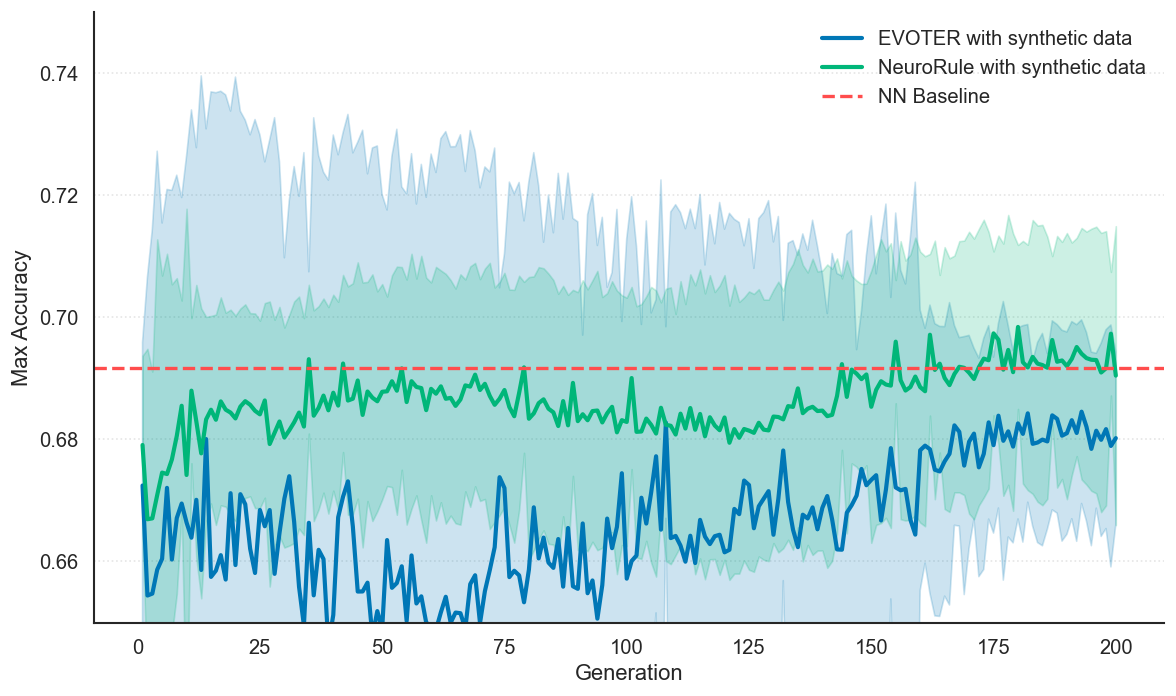}\\[-1ex]
\centerline{\small 80\% ID/OOD split: In-Distribution (Left) vs. Out-of-Distribution (Right)}
\end{minipage}\\[3ex]
\begin{minipage}{1.0\textwidth}
\includegraphics[width=0.49\linewidth]{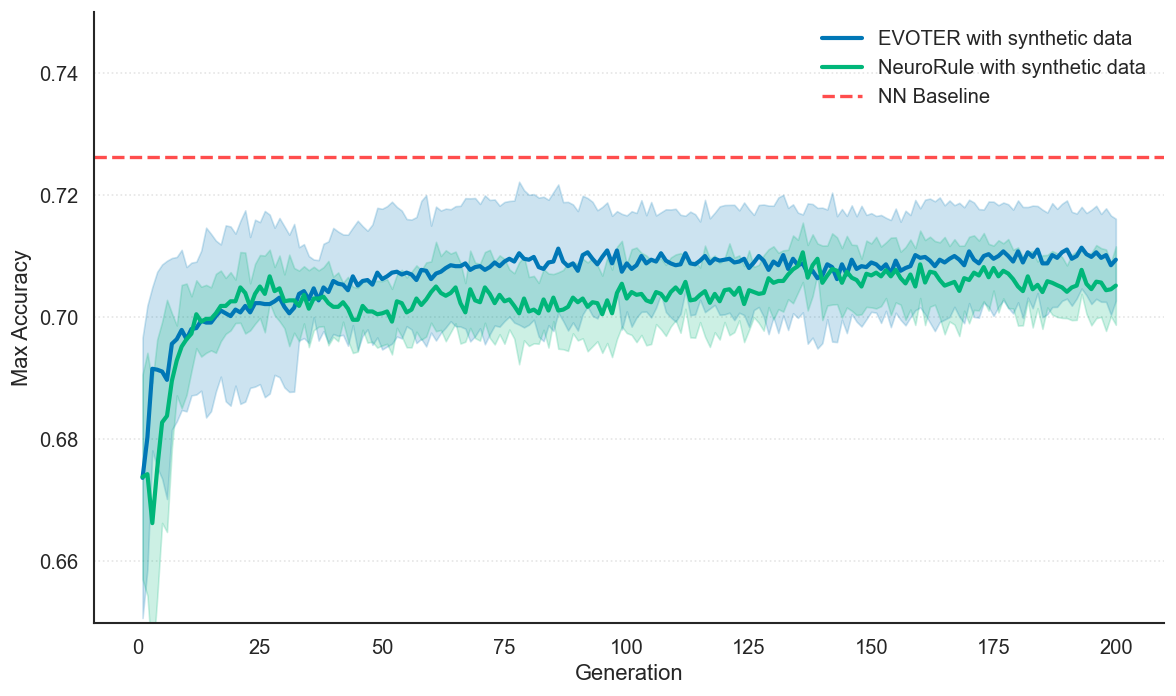}
\hfill
\includegraphics[width=0.49\linewidth]{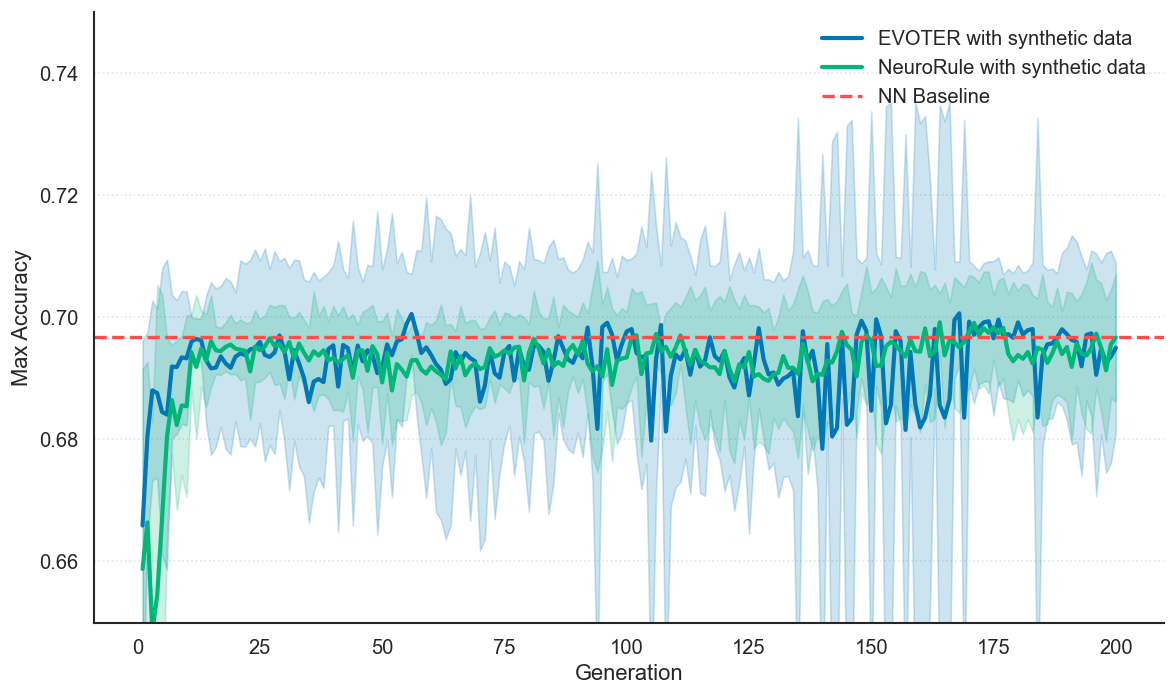}\\[-1ex]
\centerline{\small 90\% ID/OOD split: In-Distribution (Left) vs. Out-of-Distribution (Right)}
\end{minipage}\\[3ex]
\begin{minipage}{1.0\textwidth}
\includegraphics[width=0.49\linewidth]{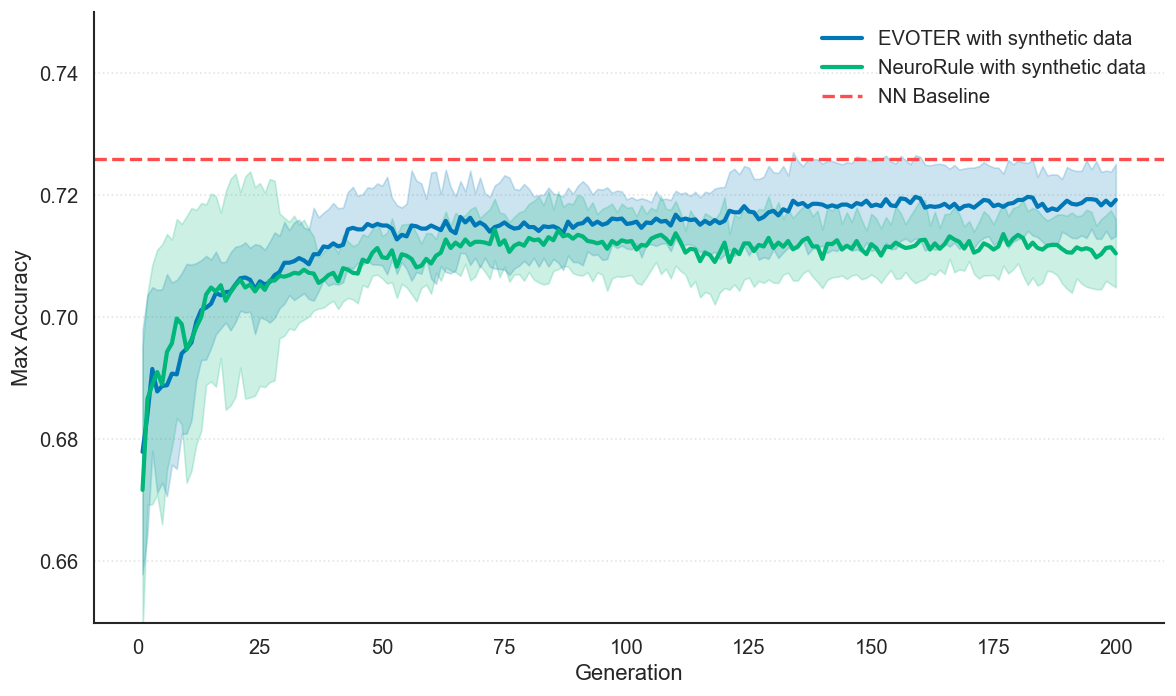}
\hfill
\includegraphics[width=0.49\linewidth]{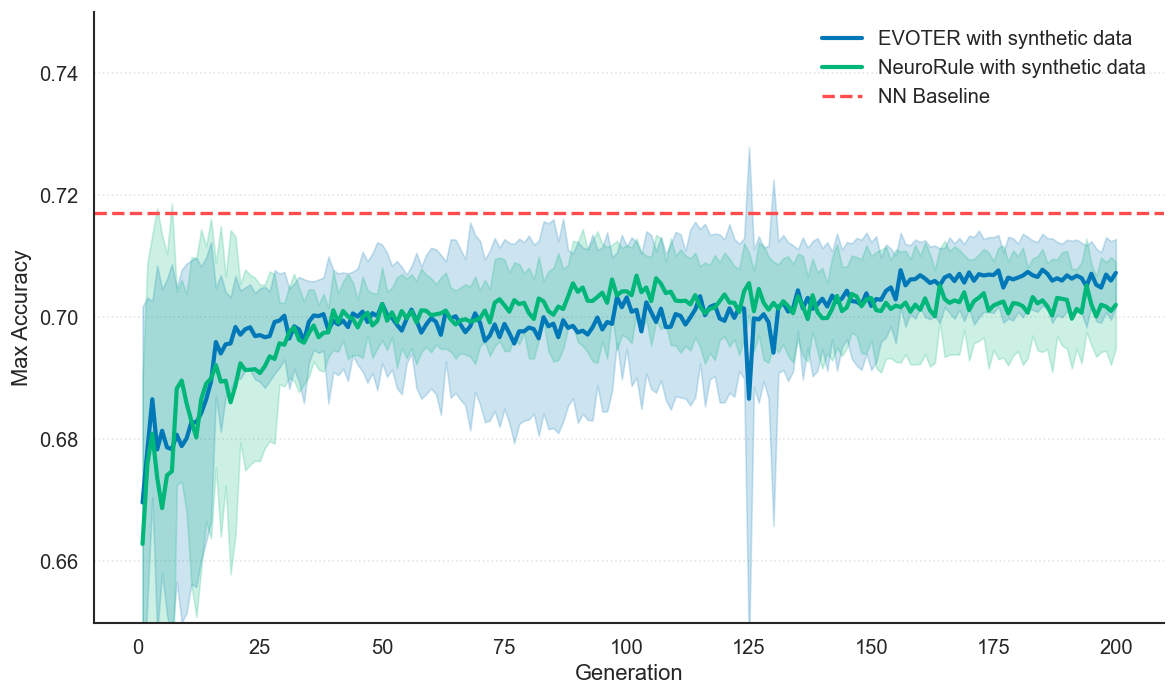}\\[-1ex]
\centerline{\small 95\% ID/OOD split: In-Distribution (Left) vs. Out-of-Distribution (Right)}
\end{minipage}
\vspace*{-1ex}
\caption{Learning curves for EVOTER and NeuroRule compared to NN accuracy at different ID/OOD splits on the Diabetes dataset when NeuroRule was evolved with synthetic data only. The curves are averages over 10 runs and the bands indicate standard deviation. The accuracy of NeuroRule is lower than when evolved with samples from the original dataset, but still viable, demonstrating that the network represents a portable source of knowledge about the domain even when the original dataset is not available. The results for Breast Cancer and Heart Disease datasets are in Figures~\ref{fig:bc_synthetic} and~\ref{fig:heart_synthetic}.}
\label{fig:diabetes_synthetic}
\end{figure}

Compared to evolution with original data, the resulting rule-sets indeed had lower accuracy on the original samples (Table~\ref{tab:synthetic}; Figures~\ref{fig:diabetes_synthetic},~\ref{fig:bc_synthetic}, and~\ref{fig:heart_synthetic}). This decrease suggests that the NN contains inaccurate decision boundaries in the input space regions where training data were sparse. The rule-sets evolved to match the NN's behavior in these regions more than before, which in turn decreased its accuracy on the original samples. This result highlights a critical sensitivity in the distillation process. Because the rule-set is inherently sparse, it is sensitive to noise in the NN's behavior. While the NN acts as a regularizer when anchored to real data, its ungrounded predictions on synthetic samples introduce logic-distorting noise. 

\begin{table}[!t]
\centering
\caption{Experiment~3: Accuracy of NeuroRule evolved with synthetic data only. The NN and EVOTER baselines are the same as in Table~\ref{tab:single_accuracies}. The numbers are averages of 10 runs; the bolded values identify the best performance in each category. NeuroRule had a lower accuracy than when evolved with actual data samples (Table~\ref{tab:single_accuracies}), and also lower than EVOTER with actual data samples, presumably because it invests more effort into areas of input space where data is actually and the NN is inaccurate. However, even with this performance gap, the models are accurate enough for the explanations to make sense. The learning curves with standard deviations are shown in Figures~\ref{fig:diabetes_synthetic},~\ref{fig:bc_synthetic}, and-\ref{fig:heart_synthetic}.}
\label{tab:synthetic}
\vspace*{-1ex}
\centering
{\small
Breast Cancer Dataset\\[1ex]
\begin{tabular}{lrrr|rrr}
\hline
              & \multicolumn{3}{c}{In-Distribution (ID)} & \multicolumn{3}{c}{Out-of-Distribution (OOD)} \\ \cline{2-7} 
ID/OOD split & NN    & EVOTER          & NeuroRule & NN    & EVOTER          & NeuroRule \\ \hline
80\%         & \textbf{1.000} & 0.944 & 0.927     & 0.759 & \textbf{0.892} & 0.867     \\
90\%         & \textbf{0.979} & 0.919 & 0.883     & 0.887 & \textbf{0.908}          & \textbf{0.908}     \\
95\%         & \textbf{0.925} & 0.875 & 0.824     & \textbf{0.987} & 0.927 & 0.887     \\ \hline  \\
\end{tabular}}

\centering
{\small
Heart Disease Dataset\\[1ex]
\begin{tabular}{lrrr|rrr}
\hline
              & \multicolumn{3}{c}{In-Distribution (ID)} & \multicolumn{3}{c}{Out-of-Distribution (OOD)} \\ \cline{2-7}
ID/OOD split & NN    & EVOTER          & NeuroRule & NN    & EVOTER          & NeuroRule      \\ \hline
80\%         & 0.888 & \textbf{0.898} & 0.835     & 0.777 & \textbf{0.842} & 0.800          \\
90\%         & \textbf{0.880} & 0.868 & 0.833     & 0.744 & 0.815          & \textbf{0.828} \\
95\%         & \textbf{0.902} & 0.891 & 0.849     & 0.719 & 0.832          & \textbf{0.834} \\ \hline  \\
\end{tabular}}

\centering
{\small
Diabetes Dataset\\[1ex]
\begin{tabular}{lrrr|rrr}
\hline
              & \multicolumn{3}{c}{In-Distribution (ID)} & \multicolumn{3}{c}{Out-of-Distribution (OOD)} \\ \cline{2-7}
ID/OOD split & NN    & EVOTER          & NeuroRule & NN    & EVOTER          & NeuroRule      \\ \hline
80\%         & \textbf{0.749} & 0.722 & 0.691     & 0.692 & 0.679          & \textbf{0.700} \\
90\%         & \textbf{0.726} & 0.708 & 0.705     & 0.697 & 0.688          & \textbf{0.696} \\
95\%         & \textbf{0.726} & 0.716 & 0.711     & \textbf{0.717} & 0.705 & 0.701          \\ \hline \\
\end{tabular}}
\end{table}

Despite the observed performance gap, it is important to note that Experiment~3 provides a significant feasibility proof. Although synthetic-data rule-sets underperform their original-data counterparts, they still achieve a competent level of accuracy. In real-world applications where data privacy or proprietary constraints prevent the sharing of datasets, this distillation with synthetic data demonstrates that a rule-set can still be extracted from a black-box model. NNs can thus serve as portable knowledge repositories, allowing for the construction of interpretable models even when the original data is not available.

%

\section{Discussion}
This section describes the trends observed across the experiments and discusses the limitations of the current approach.

\paragraph{The Performance vs.\ Explainability Trade-off}
An inverse relationship between performance and explainability is a common observation in machine learning literature \cite{arrieta2019explainableartificialintelligencexai, adadi2018peekingInsideTheBlackBox}. The results in this paper challenge this trade-off: the conciseness-constrained multi-objective models were often more accurate than the unconstrained single-objective models, i.e.\ explainability does not necessarily result in decreased performance. In other words, the behavior captured by the NN can be effectively distilled into a minimal set of logical expressions without significant information loss. 

\paragraph{Symbolic Regularization through Conciseness Constraint}
Cases where multi-objective models outperform single-objective models provide strong evidence for symbolic regularization. Without a conciseness constraint, many of the rule-set solutions became complex logical structures that captured the noise in the training set. By penalizing the number of conditions in a rule-set, the multi-objective fitness function prevents the evolution from overfitting to outliers. Instead, the algorithm is forced to find the most dominant conditions. This result shows that conciseness is not only an advantage in explainability but could also result in more robust generalization in noisy environments.

\paragraph{Distributional Invariance and Extrapolation}
The disparity between the NN's and NeuroRule's OOD accuracy highlights a fundamental difference in how these two model architectures generalize, reflecting the bias-variance dilemma in machine learning \cite{geman1992neural}. Neural networks are high-variance interpolators: they perform well with ID data but less so with OOD data. In contrast, rule-sets function as high-bias extrapolators. Concise logical structures are less likely to incorporate spurious correlations that are unique to the training distribution. Because the conciseness constraint anchors the model to a minimal set of logical expressions, the rule-set maintains its predictive power even when the input distribution deviates from the training set. 

\paragraph{The NN as a Regularizer}
Rather than serving as a simple predictive model, the NN in NeuroRule functions as a high-dimensional feature extractor that maps noisy inputs to a smoothed decision manifold. By evolving rule-sets against this manifold, the evolutionary process is partially shielded from the noise in the data, and performs better than EVOTER evolved directly with the data. This result confirms that the distillation process is not merely a translation of weights into rules, but a regularization mechanism that results in better performance

\paragraph{Limitations}
While the NeuroRule framework is robust, certain experimental configurations were harder for it than others. Specifically, in cases where the rule-set failed to significantly outperform the baseline or the NN, two factors may be at play:

\begin{itemize}
    \item Feature Dimensionality and Data Distribution: The complexity of the underlying feature space—including the ratio of categorical to continuous variables—likely dictates how well the decision manifold can be distilled into a rule-set. In datasets with high-frequency noise or heavily skewed distributions, the NN may not provide a sufficiently smooth fitness signal, leading to sub-optimal convergence of the rule-set. Regularization techniques designed for neural networks, such as stochastic noise injection or landscape-smoothing methods \cite{ding2023randomsmoothingregularizationkernel, NEURIPS2021_29301521, baldi2025losslandscapeanalysisreliable} could help the performance of NeuroRule in such cases as well.
    \item Stochasticity in Evolutionary Search: As a population-based heuristic, evolution is subject to stochastic variance. Future work should investigate the impact of increasing population size and implementing more sophisticated mutation operators to ensure more consistent convergence.
\end{itemize}

\section{Future Work}

Future research will explore how the NeuroRule distillation process can be extended to a wider variety of conditions:
\paragraph{NN Architecture sensitivity:} The NN's accuracy directly influences the rule-set's accuracy, but it is possible that the results are sensitive to the type of NN as well. A more granular sensitivity analysis of e.g.\ varying network depth and activation functions is required.

\paragraph{Rule-Set Probability Outputs:} The current NeuroRule framework supports rules that produce deterministic outputs. The framework can be extended with probabilistic or distributional outputs, producing probabilities or maximum likelihoods of a range of actions. This extension would be particularly useful for applications such as robotics and control-based environments, where actions are not deterministic.
    
\paragraph{Modeling Beyond Feature Comparisons:} Other extensions to the NeuroRule framework include rules for modeling 2D/3D data inputs and language-based comparisons. Such rules can explore the distillation of Convolutional Neural Networks (CNNs) \cite{cnnsExplainability} and sequence-based networks, such as Recurrent Neural Networks (RNNs) \cite{rnnsBasedRules}. 

\paragraph{Explainability for Vector Embeddings:} A possible extension is to apply NeuroRule to explain vector embeddings. Understanding why a model converts some input values into a given embedding and how it correlates with human-readable values could eventually lead to insights into understanding large language models.

\paragraph{Interactive Framework:} The rule-sets resulting from NeuroRule are explicit and accessible to domain experts for monitoring, validation, and even editing. A compelling practical extension is a framework that makes such human-in-the-loop interventions easy to implement, allowing for broader applications of NeuroRule beyond a research setting.

In this manner, NeuroRule can serve as a foundation for explainable AI in the future.

\section{Conclusion}
This research demonstrates how the complex decision-making processes of high-capacity neural networks can be expressed as human-readable logic. Through the NeuroRule framework, black-box NN models can be transformed into transparent rule-sets, ensuring that performance does not come at the cost of explainability. Several key insights were uncovered:

First, optimizing rule-set models for accuracy in emulating the NN (rather than accuracy with the original data) can improve performance. The resulting rule-sets are less likely to overfit to noise in the original data, resulting in better generalization. Second, conciseness as a secondary evolutionary objective significantly reduces complexity. Such rule-sets are not only more explainable but may also generalize better—even better than the NN, especially in OOD cases. Third, the behavior of the NN can be reconstructed even when the original training data is no longer accessible. Thus, neural networks can act as a portable knowledge store.

Ultimately, this paper demonstrates that black-box AI systems can, in fact, be made explainable. As safety-critical industries increasingly demand explainable AI, NeuroRule makes it possible to take advantage of high-performance machine learning in those industries as well.

\bibliographystyle{ACM-Reference-Format}
\bibliography{acmart}

\newpage
\appendix
\section{Experiment~1 Learning Curves for the Breast Cancer and Heart Disease Datasets}
\label{appendix:surrogate_guided}

\begin{figure}[!b]
\centering

\begin{minipage}{1.0\textwidth}
\includegraphics[width=0.49\linewidth]{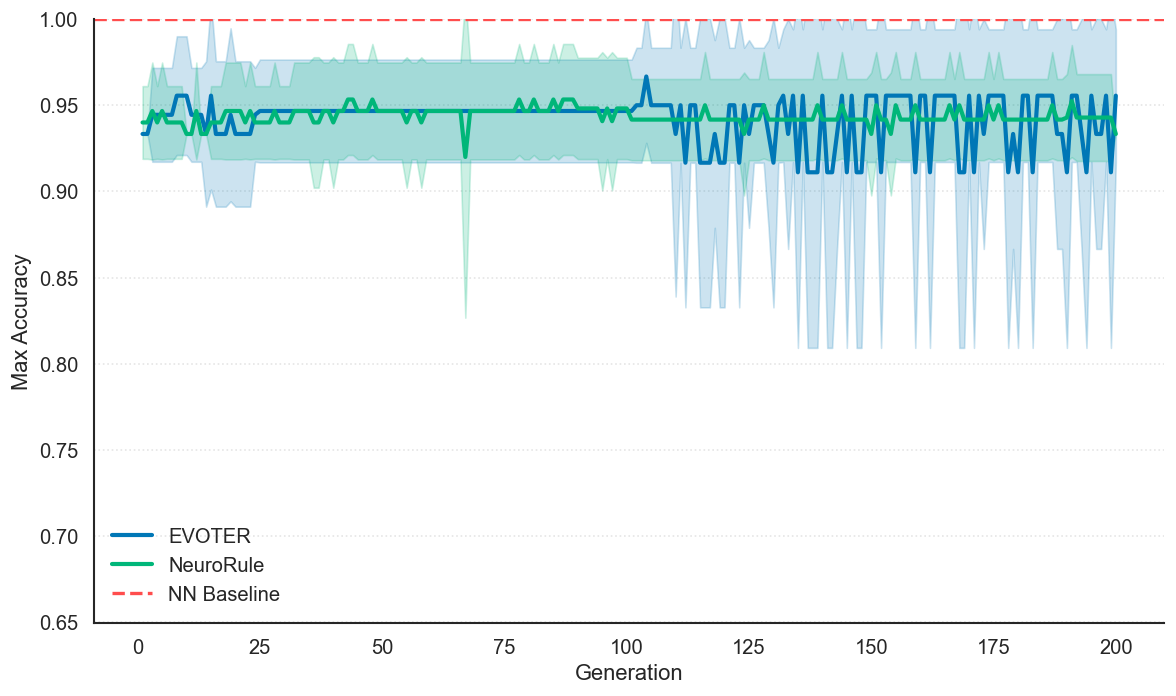}
\hfill
\includegraphics[width=0.49\linewidth]{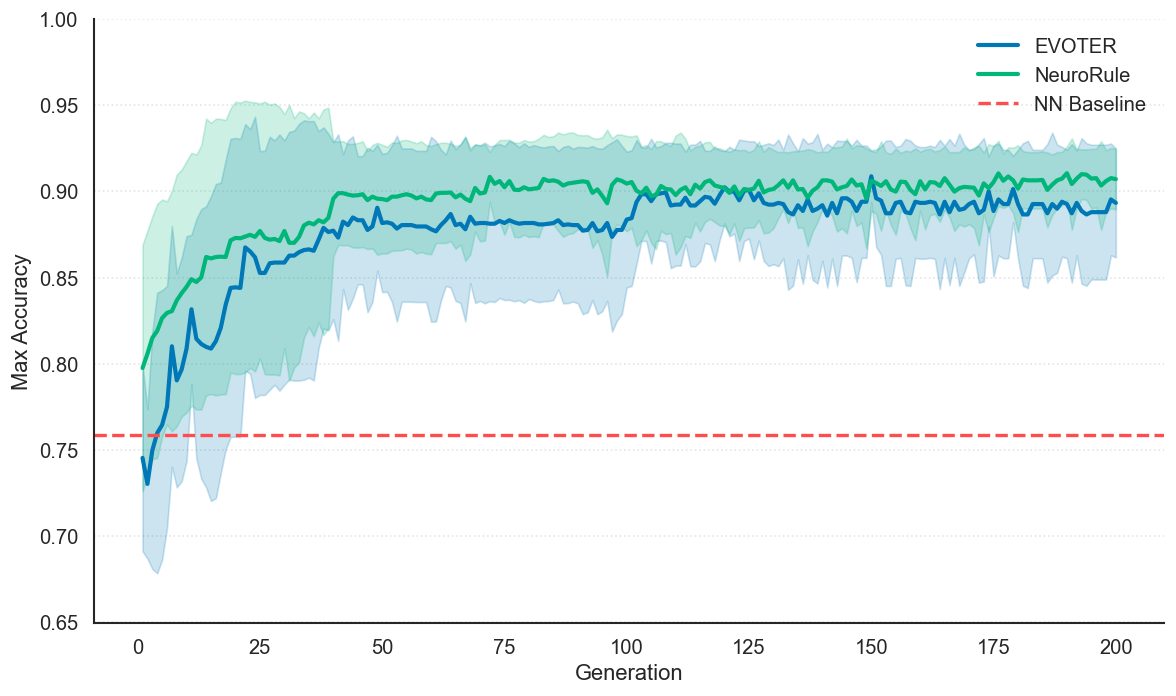}\\[-1ex]
\centerline{\small 80\% ID data: In-Distribution (Left) vs. Out-of-Distribution (Right)}
\end{minipage}\\[3ex]
\begin{minipage}{1.0\textwidth}
\includegraphics[width=0.49\linewidth]{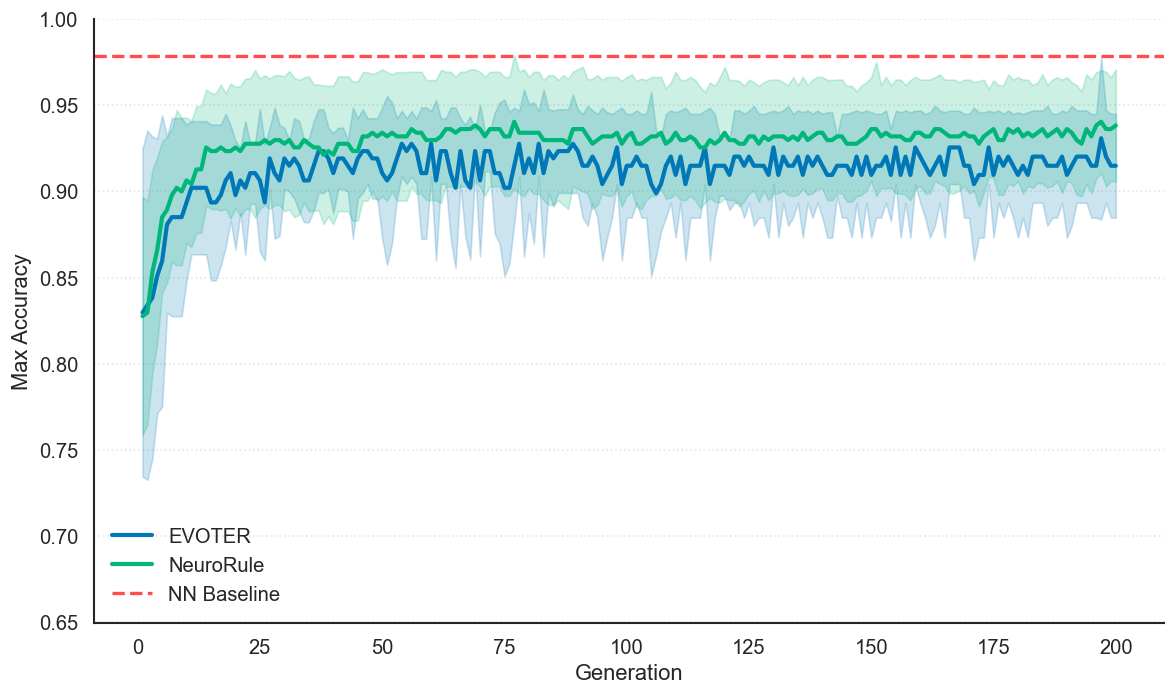}
\hfill
\includegraphics[width=0.49\linewidth]{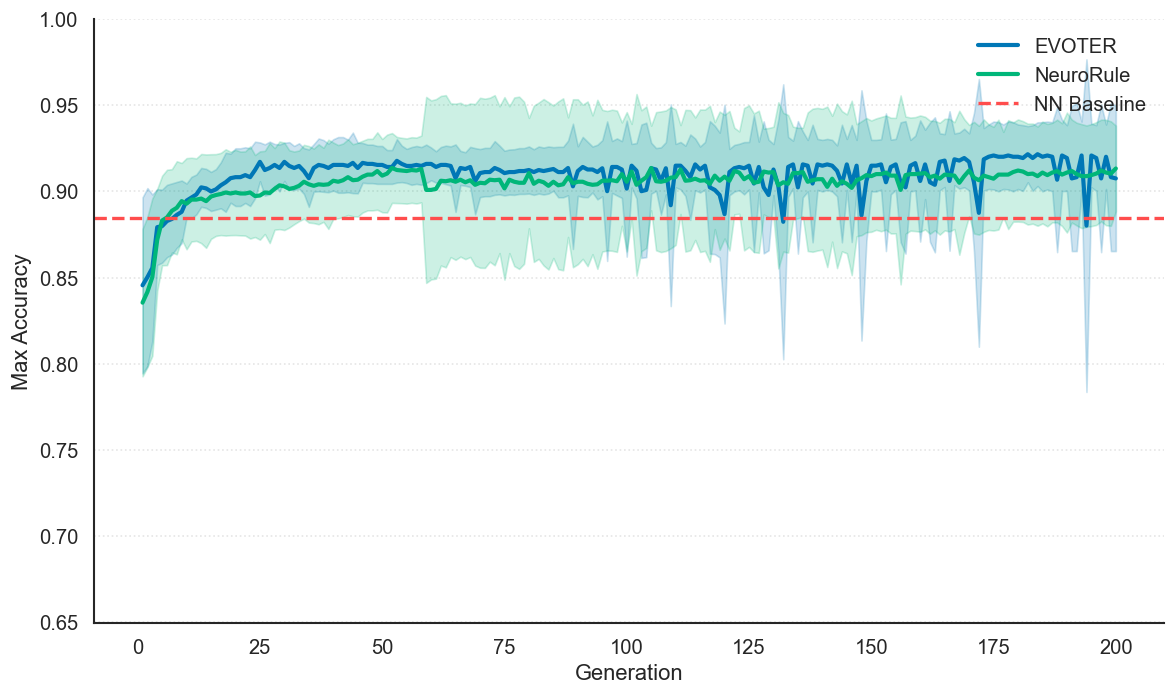}\\[-1ex]
\centerline{\small 90\% ID data: In-Distribution (Left) vs. Out-of-Distribution (Right)}
\end{minipage}\\[3ex]
\begin{minipage}{1.0\textwidth}
\includegraphics[width=0.49\linewidth]{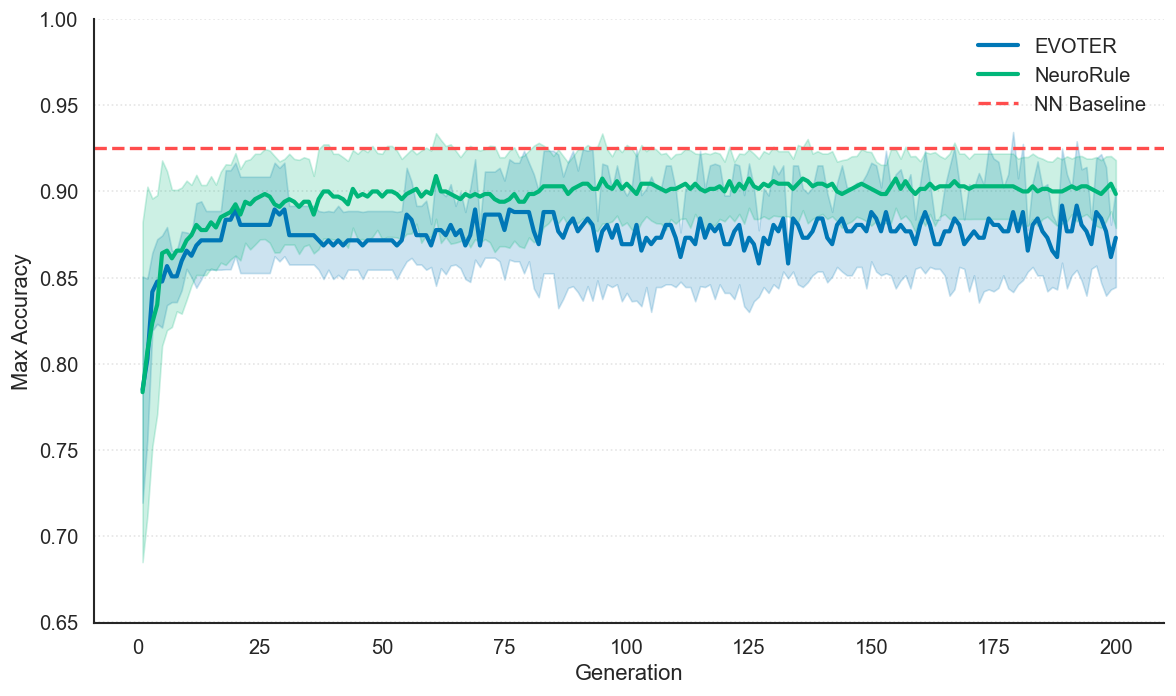}
\hfill
\includegraphics[width=0.49\linewidth]{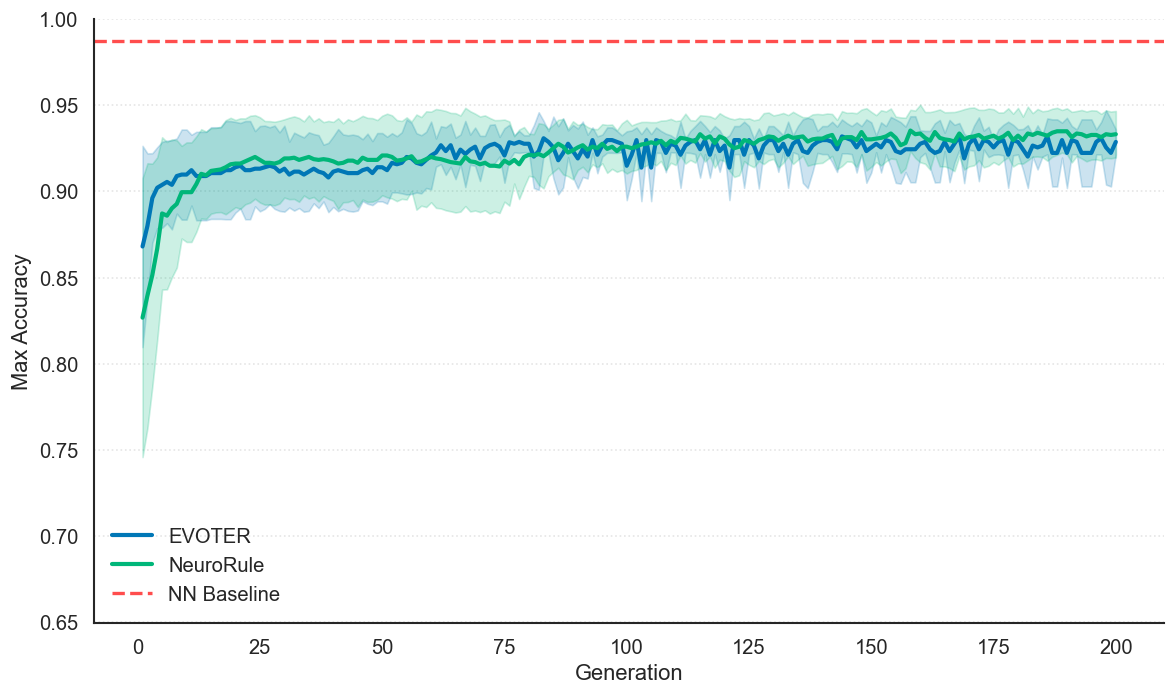}\\[-1ex]
\centerline{\small 95\% ID data: In-Distribution (Left) vs. Out-of-Distribution (Right)}
\end{minipage}
\vspace*{-1ex}
\caption{Learning curves for EVOTER and NeuroRule compared to NN accuracy at different ID/OOD splits on the Breast Cancer Dataset. This dataset is smaller than Diabetes, but the results are similar. However, likely due to the small amount of data, the EVOTER accuracy has more variance.}
\label{fig:experiment1_bc}
\end{figure}

\begin{figure}[!t]
\centering

\begin{minipage}{1.0\textwidth}
\includegraphics[width=0.49\linewidth]{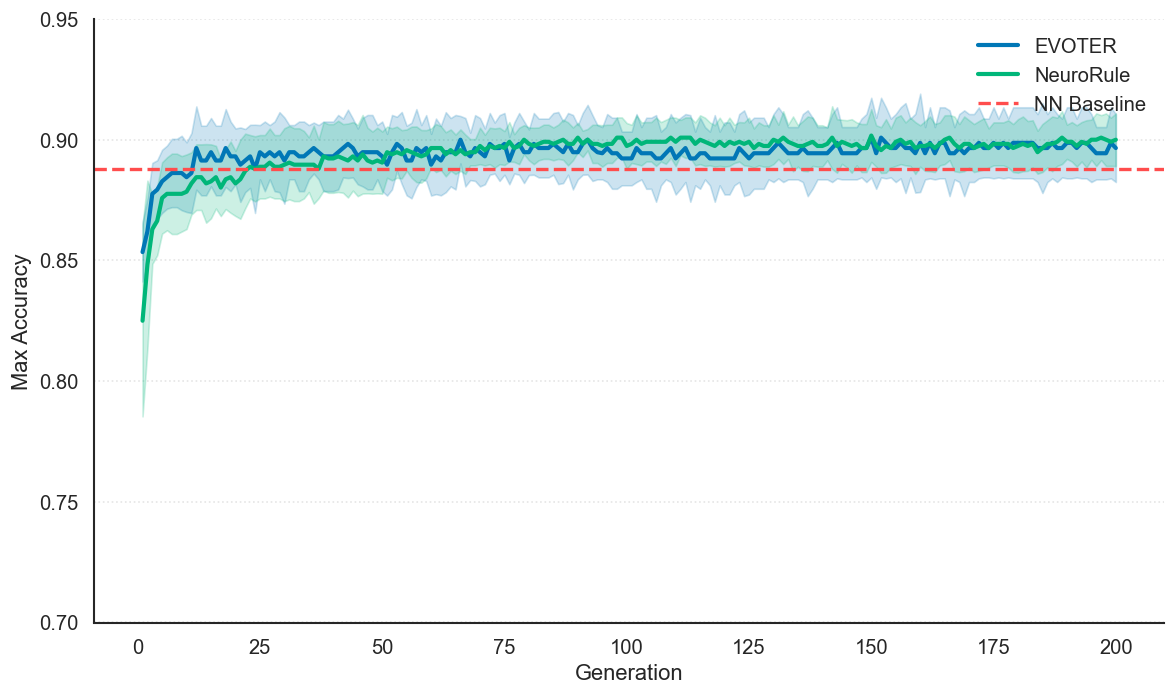}
\hfill
\includegraphics[width=0.49\linewidth]{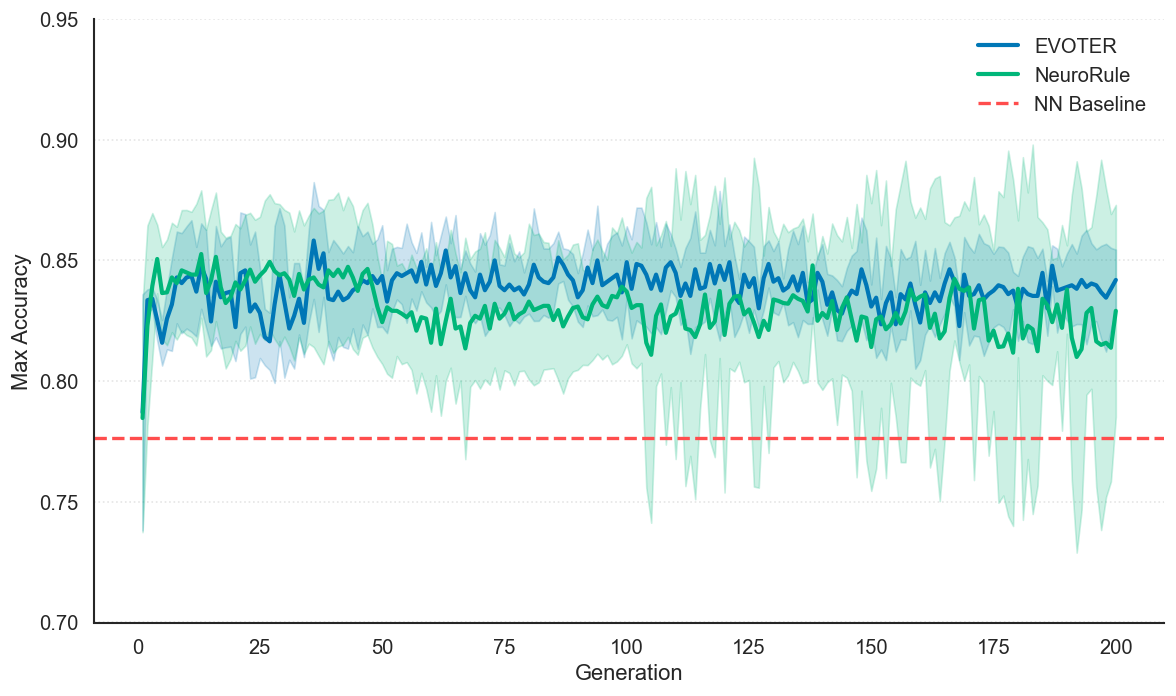}\\[-1ex]
\centerline{\small 80\% ID data: In-Distribution (Left) vs. Out-of-Distribution (Right)}
\end{minipage}\\[3ex]
\begin{minipage}{1.0\textwidth}
\includegraphics[width=0.49\linewidth]{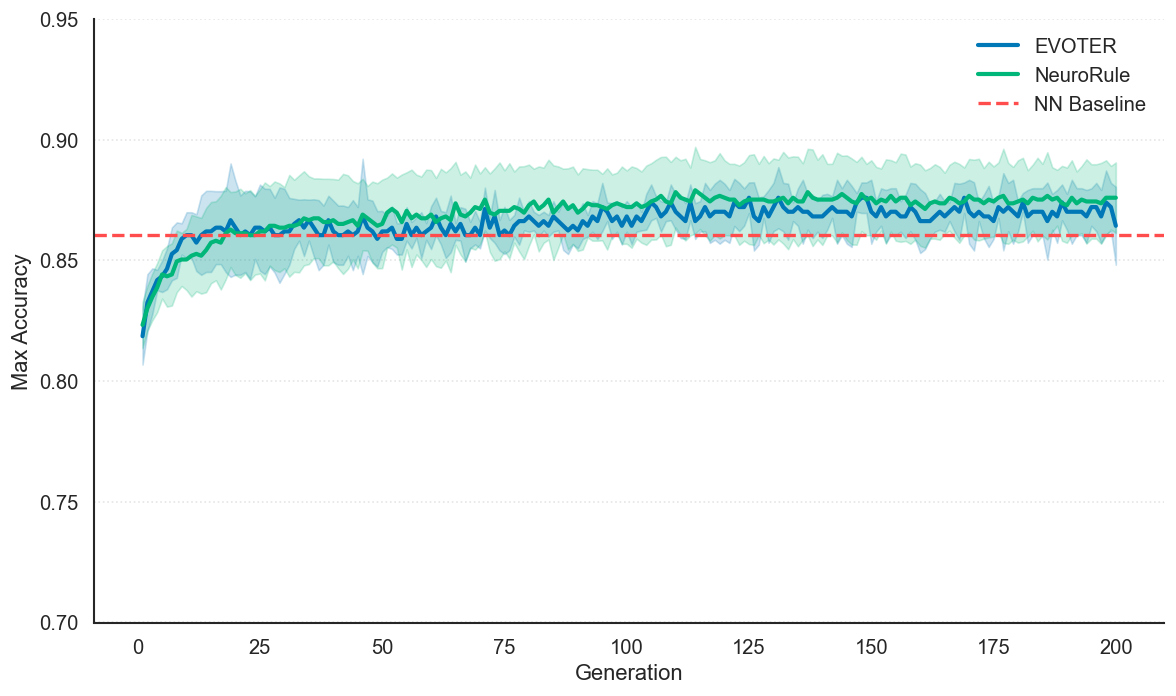}
\hfill
\includegraphics[width=0.49\linewidth]{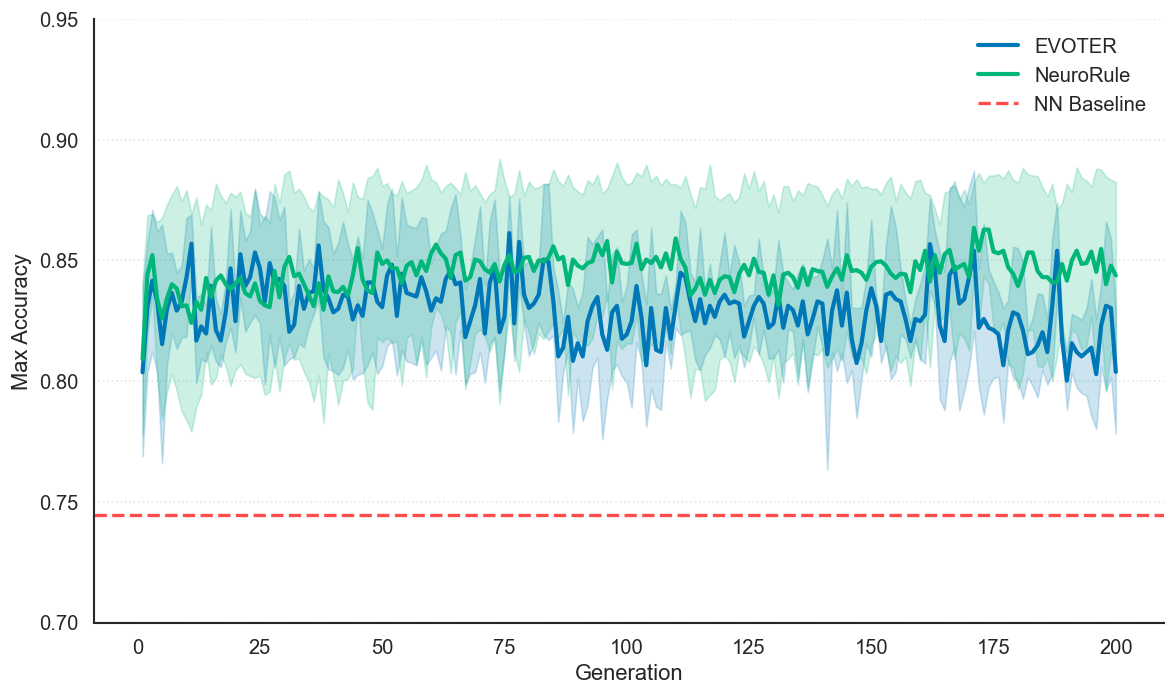}\\[-1ex]
\centerline{\small 90\% ID data: In-Distribution (Left) vs. Out-of-Distribution (Right)}
\end{minipage}\\[3ex]
\begin{minipage}{1.0\textwidth}
\includegraphics[width=0.49\linewidth]{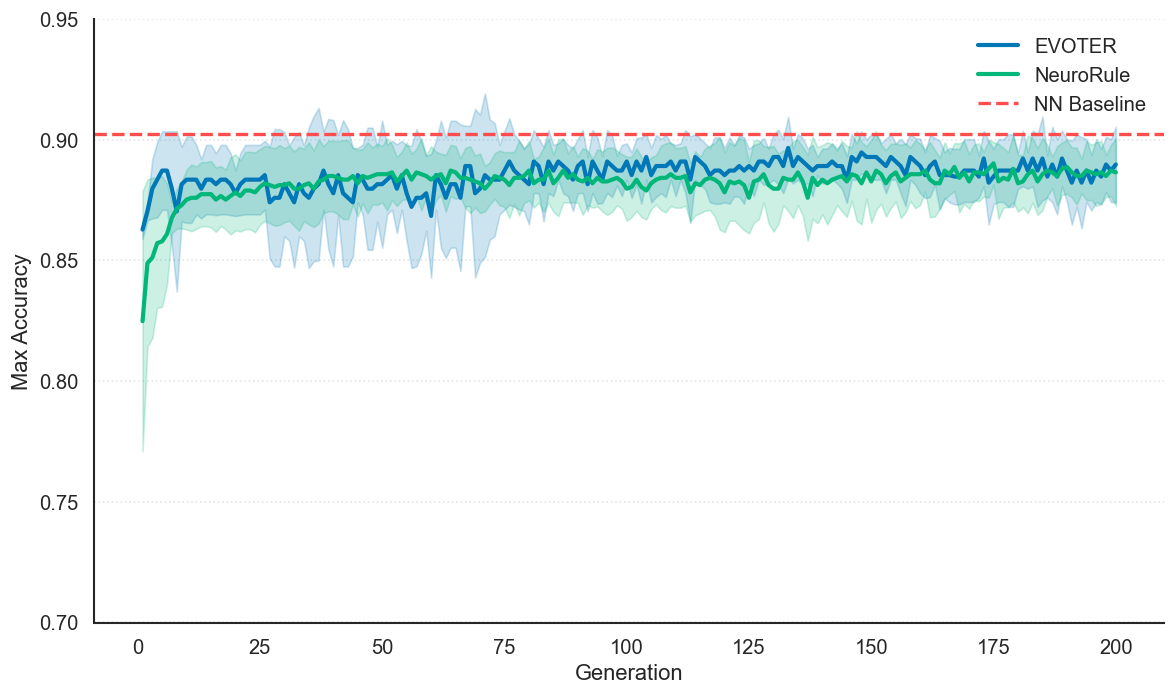}
\hfill
\includegraphics[width=0.49\linewidth]{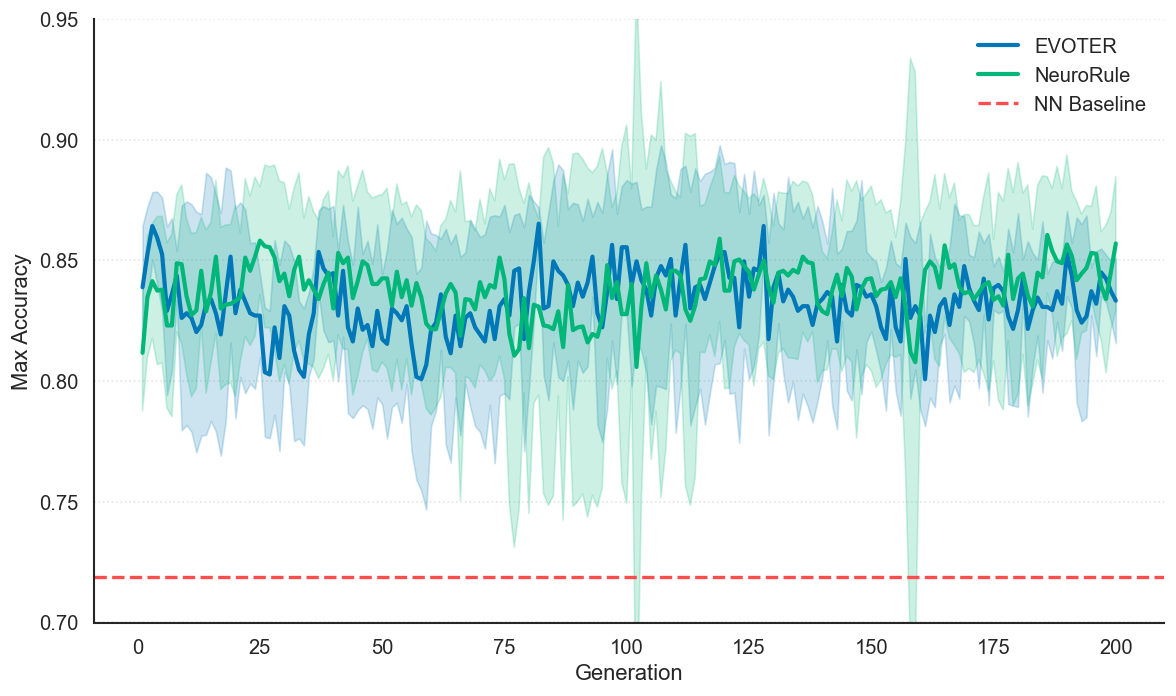}\\[-1ex]
\centerline{\small 95\% ID data: In-Distribution (Left) vs. Out-of-Distribution (Right)}
\end{minipage}
\vspace*{-1ex}
\caption{Learning curves for EVOTER and NeuroRule compared to NN accuracy at different ID/OOD splits on the Heart Disease Dataset. This dataset is also smaller, but it turned out to be a particularly good fit for the rule-set approaches: They are more accurate in five of the six cases, including the 80\% and 90\% ID cases.}
\label{fig:experiment1_heart}
\end{figure}

The learning curves for Breast Cancer are in Figure~\ref{fig:experiment1_bc} and for Heart Disease in Figure~\ref{fig:experiment1_heart}.

\clearpage
\section{Experiment~2 Learning Curves and Accuracy vs.\ Conciseness Plots for the Breast Cancer and Heart Disease Datasets}
\label{appendix:multi_objective}

\begin{figure}[!b]
\centering
\begin{minipage}{1.0\textwidth}
\includegraphics[width=0.49\linewidth]{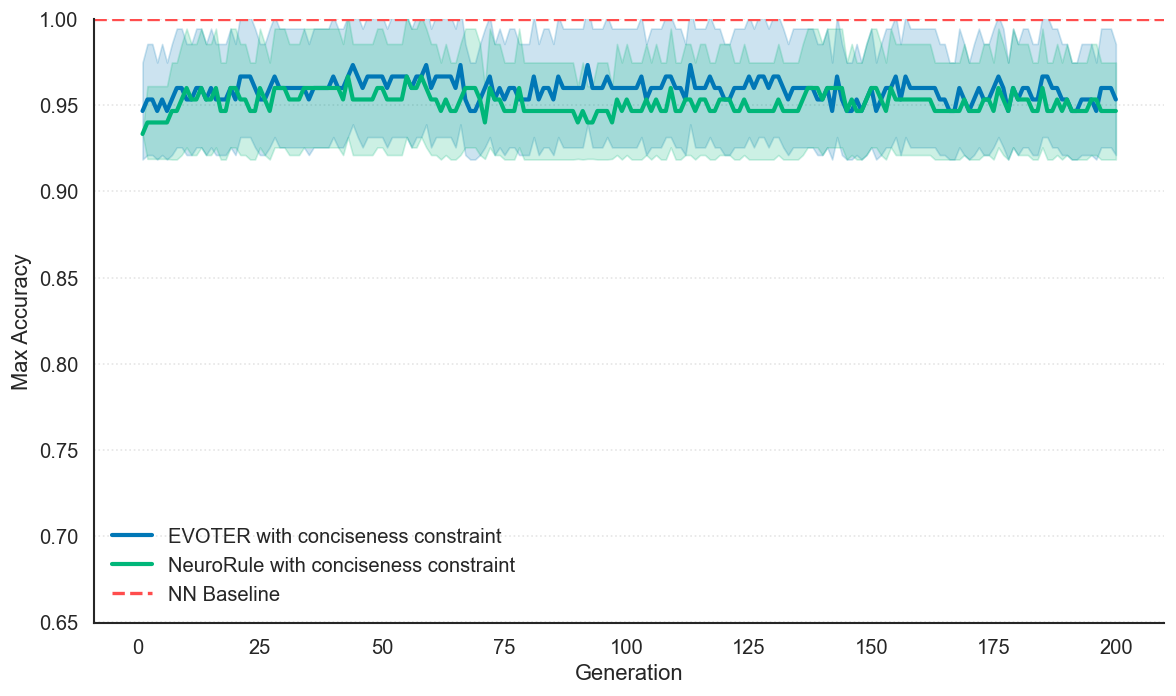}
\hfill
\includegraphics[width=0.49\linewidth]{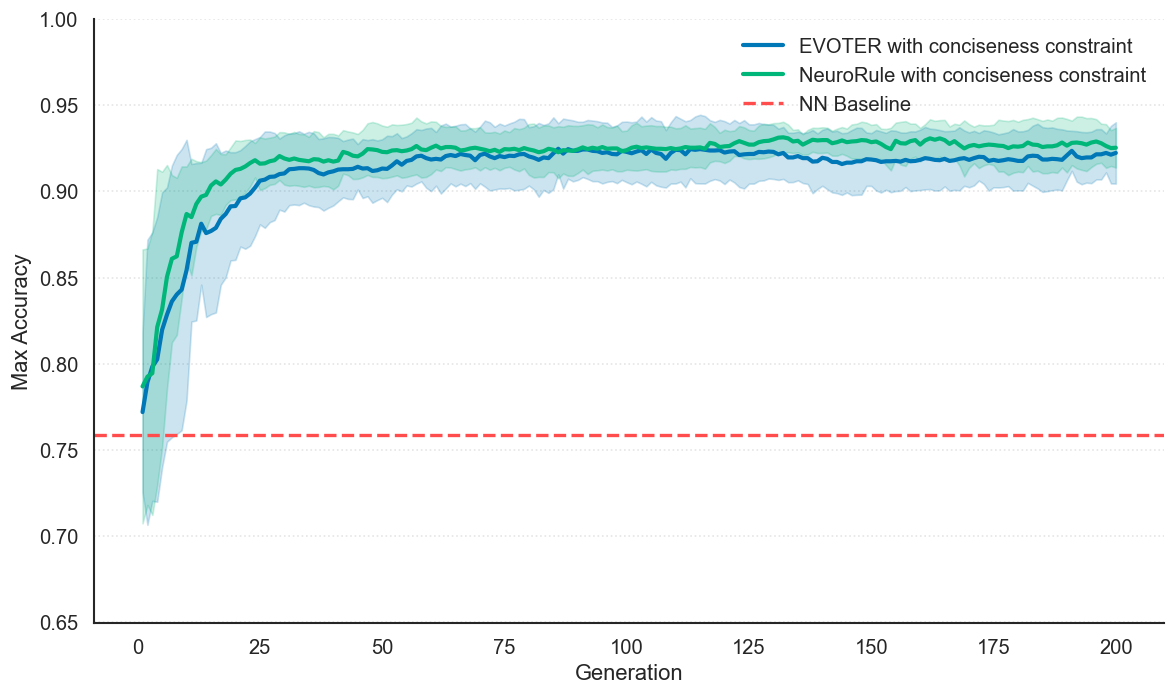}\\[-1ex]
\centerline{\small 80\% ID/OOD split: In-Distribution (Left) vs. Out-of-Distribution (Right)}
\end{minipage}\\[3ex]
\begin{minipage}{1.0\textwidth}
\includegraphics[width=0.49\linewidth]{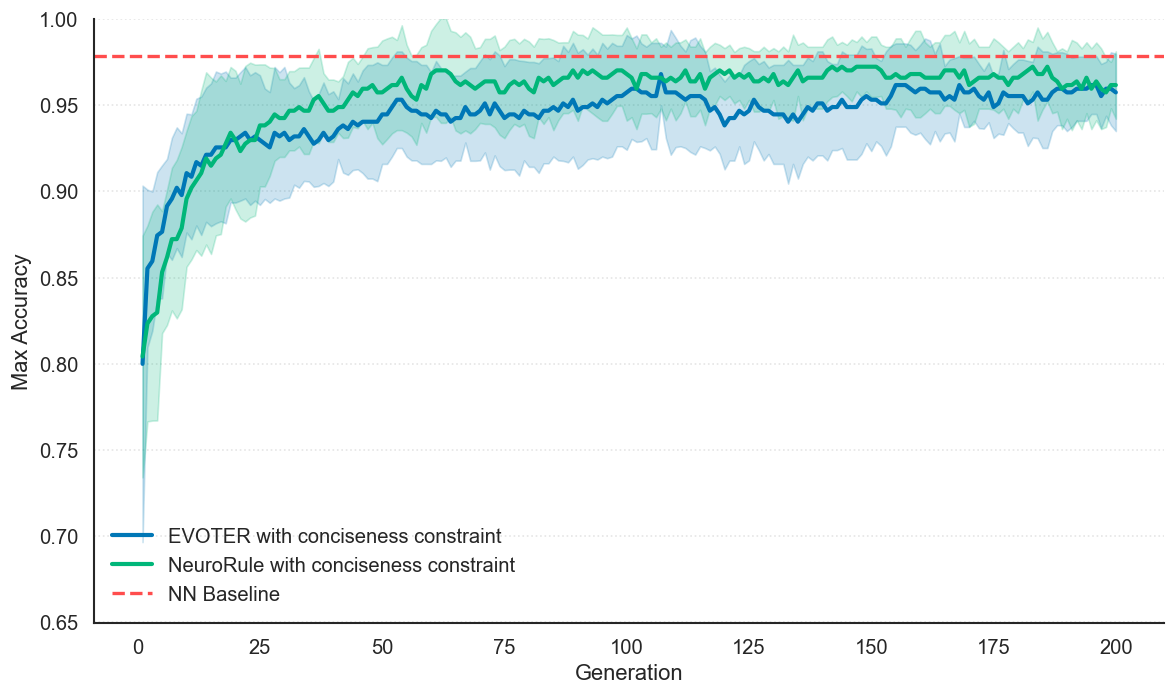}
\hfill
\includegraphics[width=0.49\linewidth]{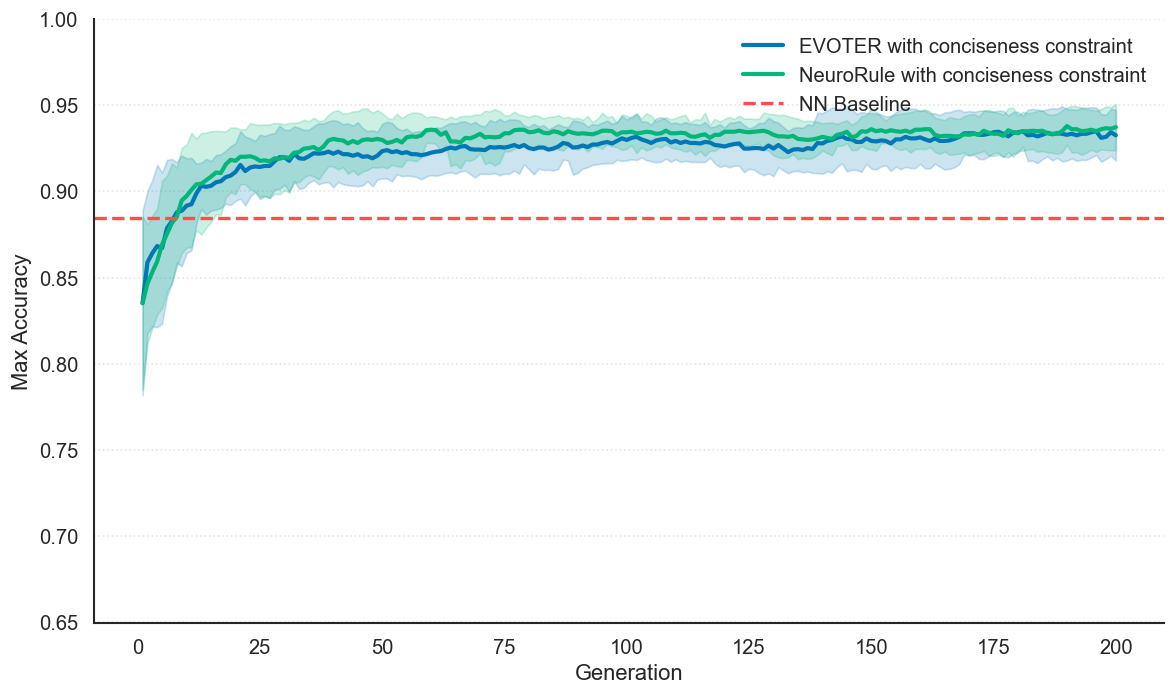}\\[-1ex]
\centerline{\small 90\% ID/OOD split: In-Distribution (Left) vs. Out-of-Distribution (Right)}
\end{minipage}\\[3ex]
\begin{minipage}{1.0\textwidth}
\includegraphics[width=0.49\linewidth]{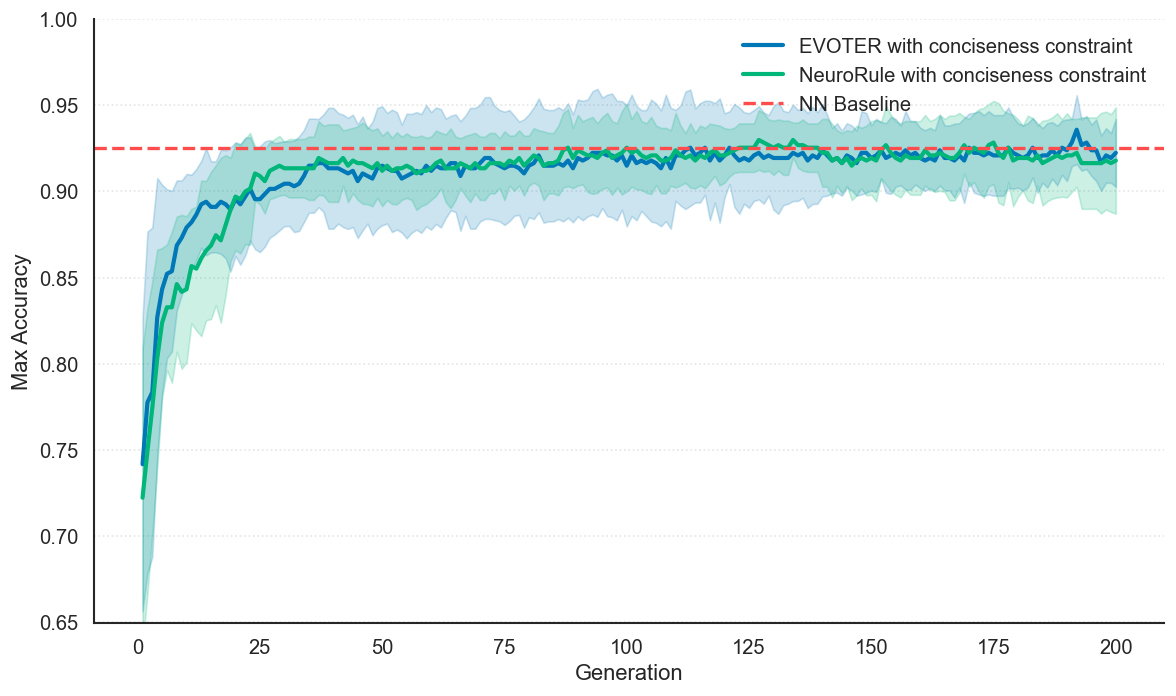}
\hfill
\includegraphics[width=0.49\linewidth]{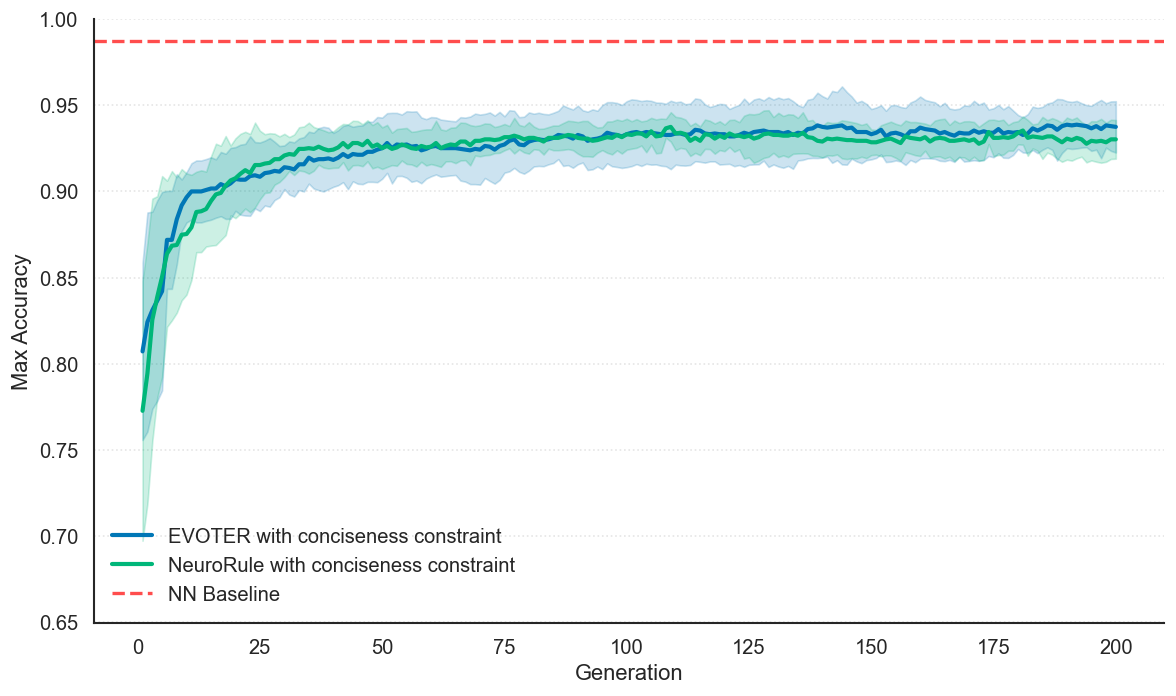}\\[-1ex]
\centerline{\small 95\% ID/OOD split: In-Distribution (Left) vs. Out-of-Distribution (Right)}
\end{minipage}
\vspace*{-1ex}
\caption{Learning curves for EVOTER and NeuroRule compared to NN accuracy at different ID/OOD splits on the Breast Cancer dataset under multi-objective optimization of accuracy and conciseness. The ID results initial form vertical lines due to limited data but gradually become more scattered like the OOD results as more data is introduced in the training interval. The results are similar to those with the Diabetes dataset, demonstrating significant improvement over single-objective evolution, and less variance.}
\label{fig:bc_multi_curves}
\end{figure}

\begin{figure}[!t]
\centering
\begin{minipage}{1.0\textwidth}
\includegraphics[width=0.49\linewidth]{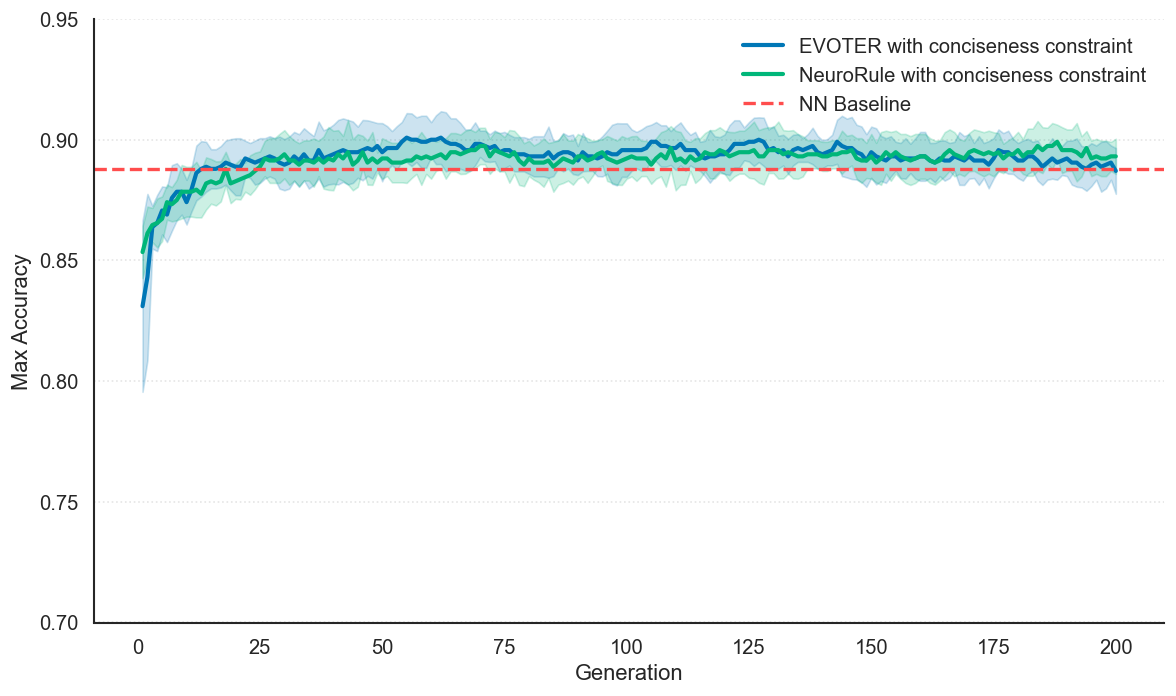}
\hfill
\includegraphics[width=0.49\linewidth]{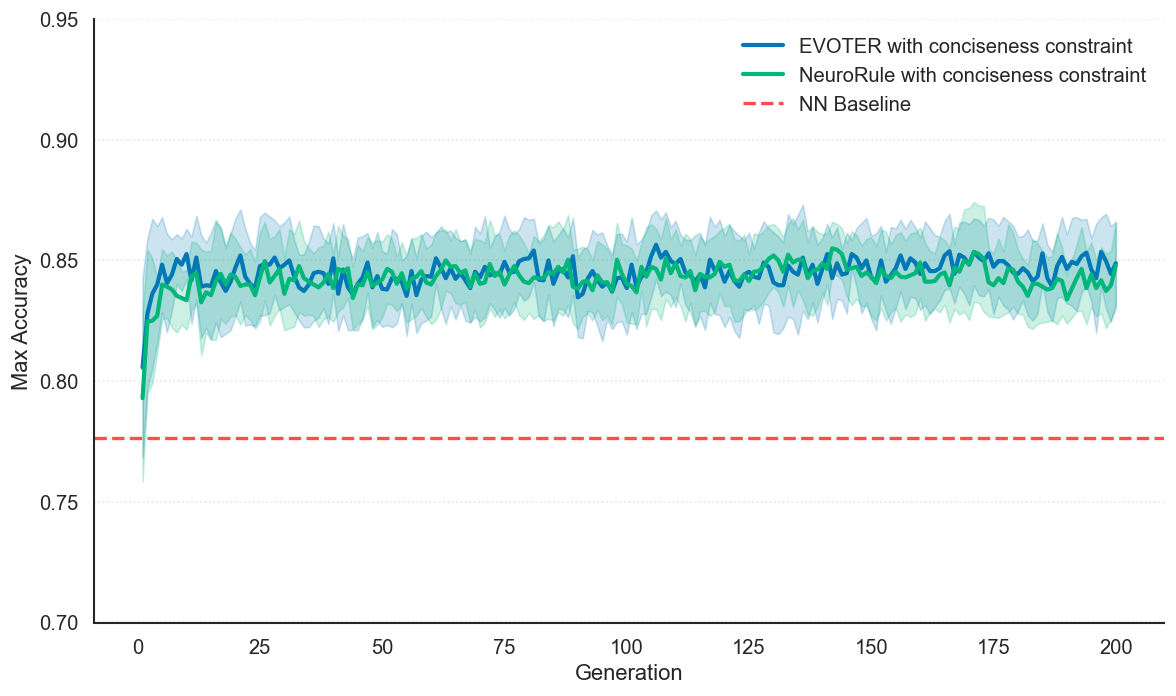}\\[-1ex]
\centerline{\small 80\% ID/OOD split: In-Distribution (Left) vs. Out-of-Distribution (Right)}
\end{minipage}\\[3ex]
\begin{minipage}{1.0\textwidth}
\includegraphics[width=0.49\linewidth]{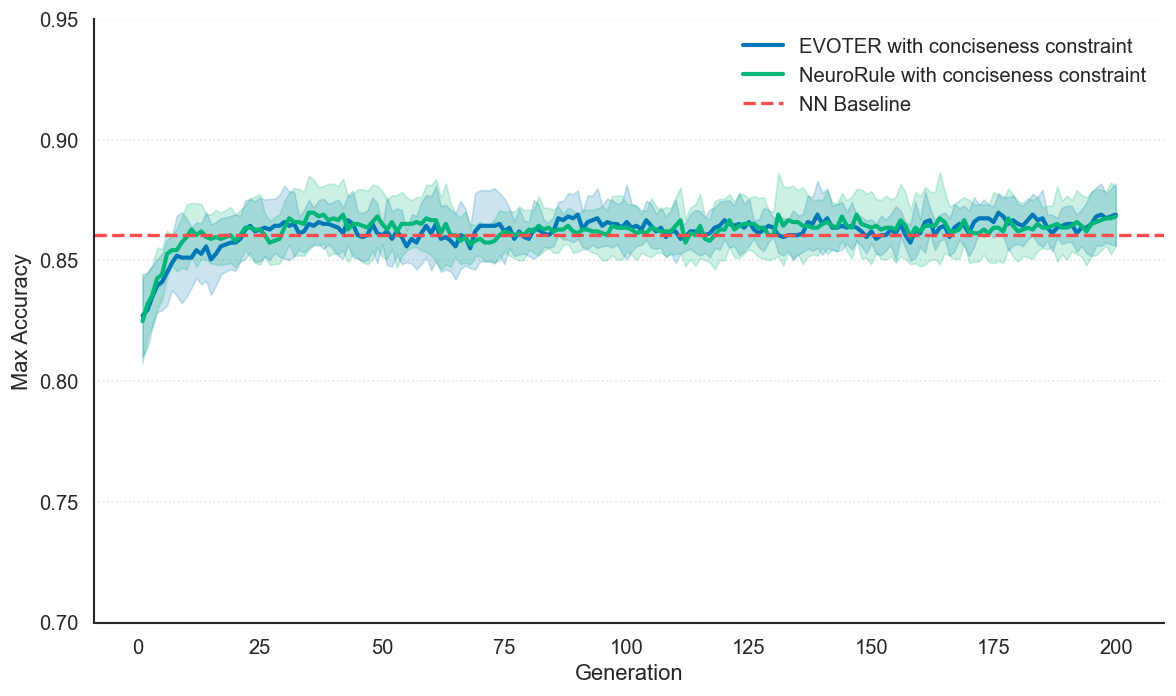}
\hfill
\includegraphics[width=0.49\linewidth]{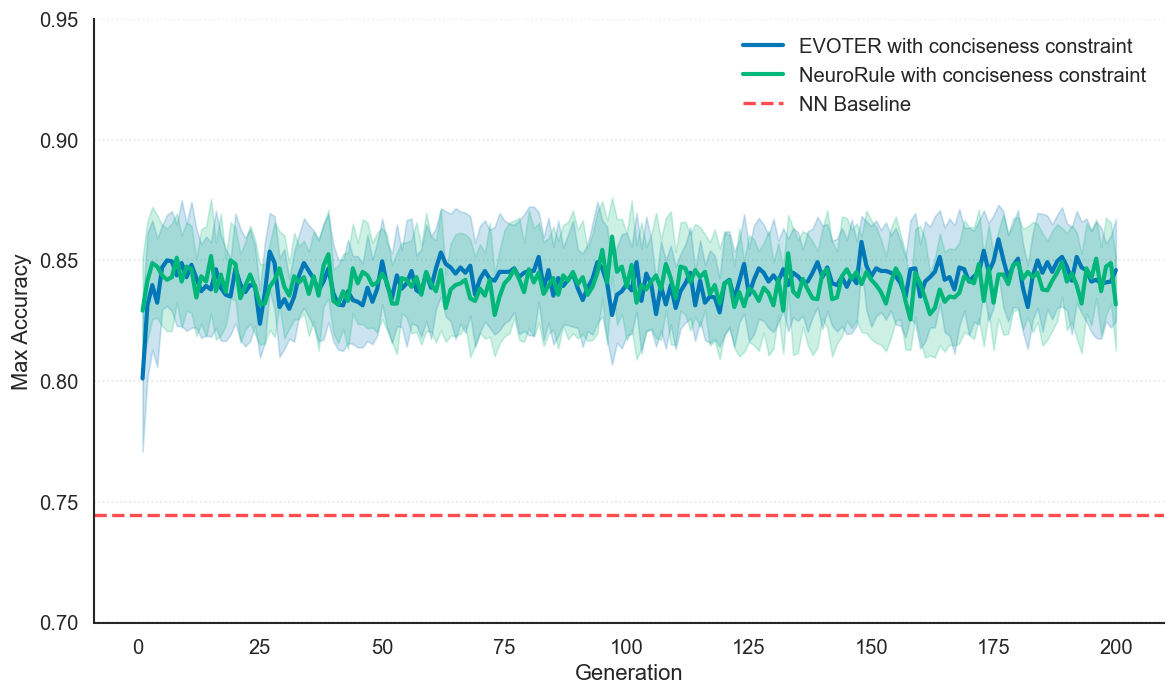}\\[-1ex]
\centerline{\small 90\% ID/OOD split: In-Distribution (Left) vs. Out-of-Distribution (Right)}
\end{minipage}\\[3ex]
\begin{minipage}{1.0\textwidth}
\includegraphics[width=0.49\linewidth]{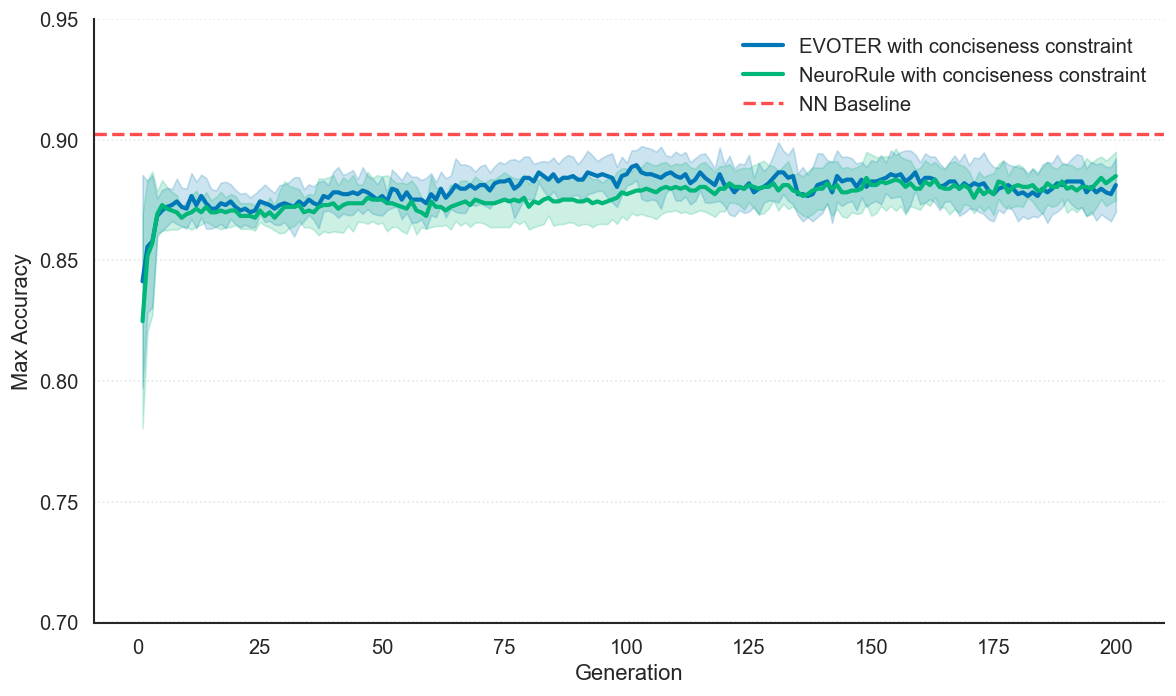}
\hfill
\includegraphics[width=0.49\linewidth]{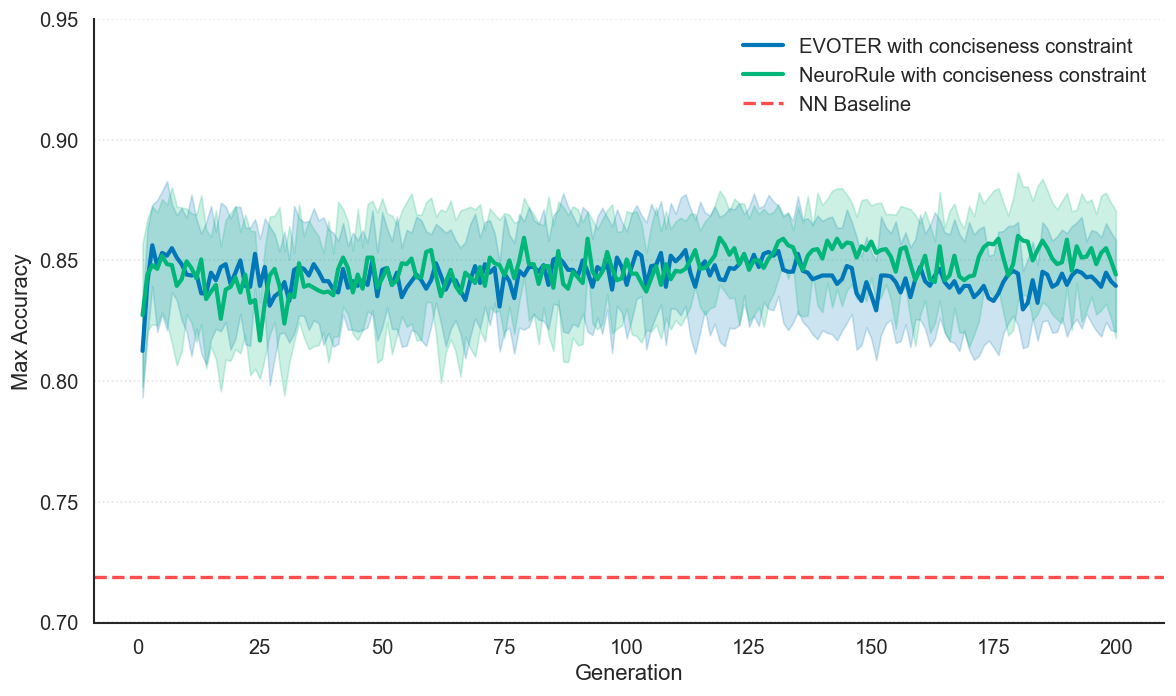}\\[-1ex]
\centerline{\small 95\% ID/OOD split: In-Distribution (Left) vs. Out-of-Distribution (Right)}
\end{minipage}
\vspace*{-1ex}
\caption{Learning curves for EVOTER and NeuroRule compared to NN accuracy at different ID/OOD splits on the Heart Disease dataset under multi-objective optimization of accuracy and conciseness. The results are similar to those with the Diabetes dataset, demonstrating significant improvement over single-objective evolution, and less variance.}
\label{fig:heart_multi_curves}
\end{figure}

\begin{figure}[!b]
\centering
\begin{minipage}{1.0\textwidth}
\includegraphics[width=0.49\linewidth]{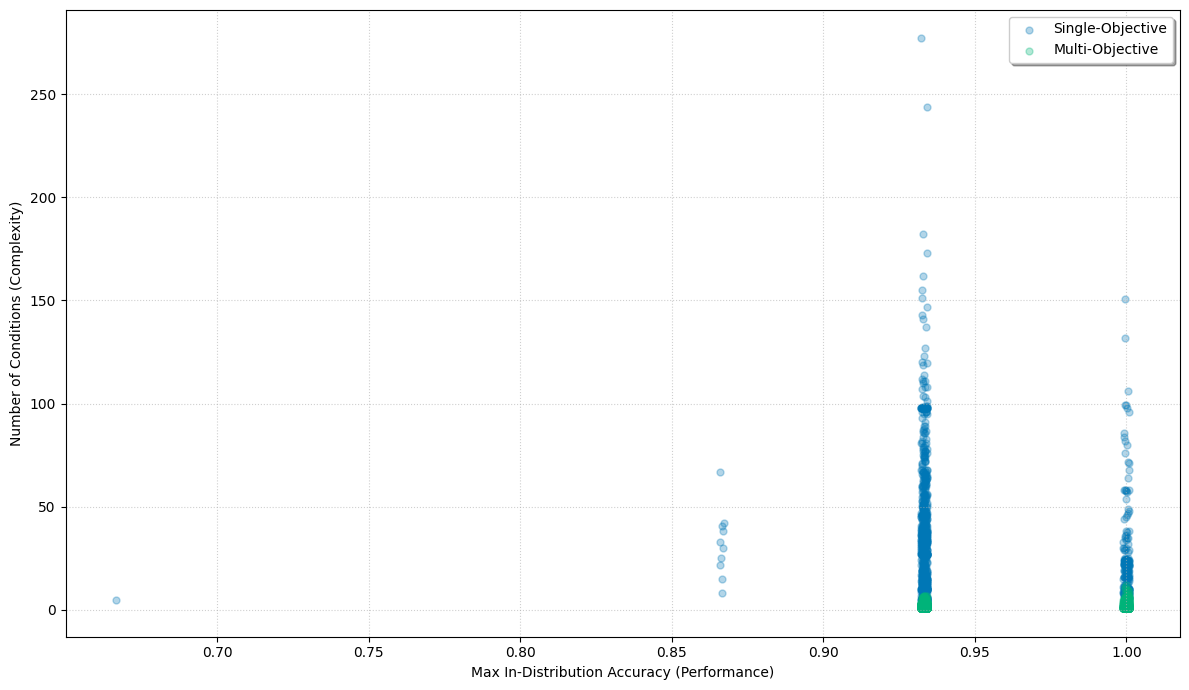}
\hfill
\includegraphics[width=0.49\linewidth]{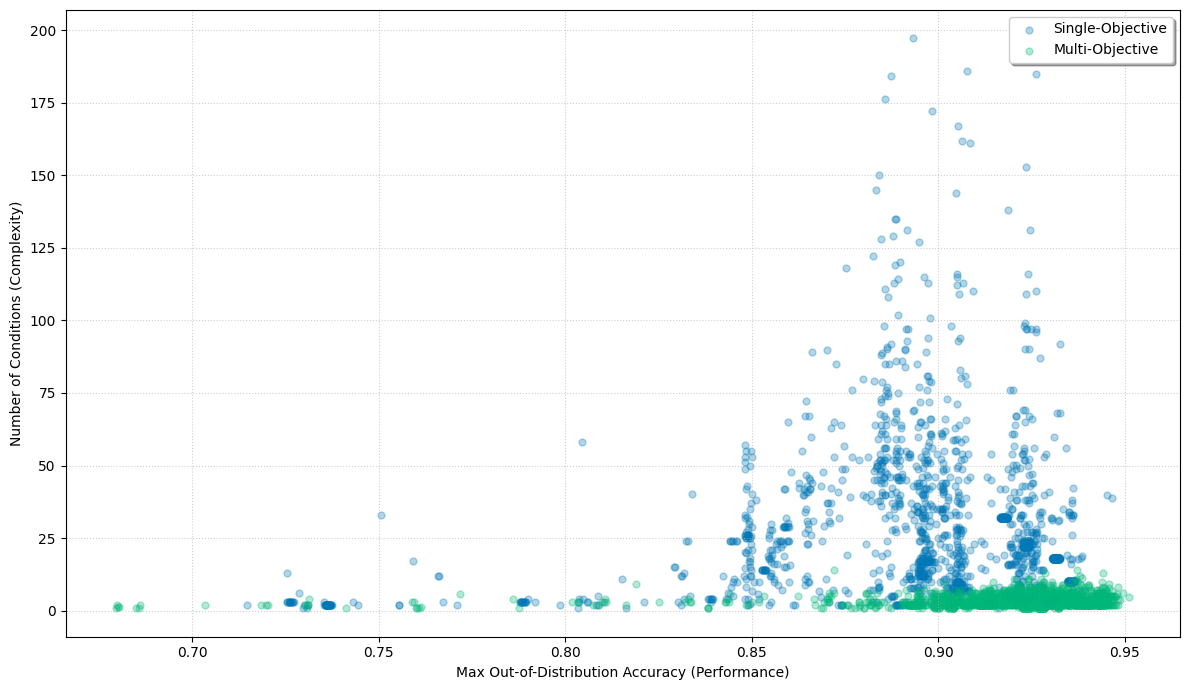}\\[-1ex]
\centerline{\small 80\% ID/OOD split: In-Distribution (Left) vs. Out-of-Distribution (Right)}
\end{minipage}\\[2ex]
\begin{minipage}{1.0\textwidth}
\includegraphics[width=0.49\linewidth]{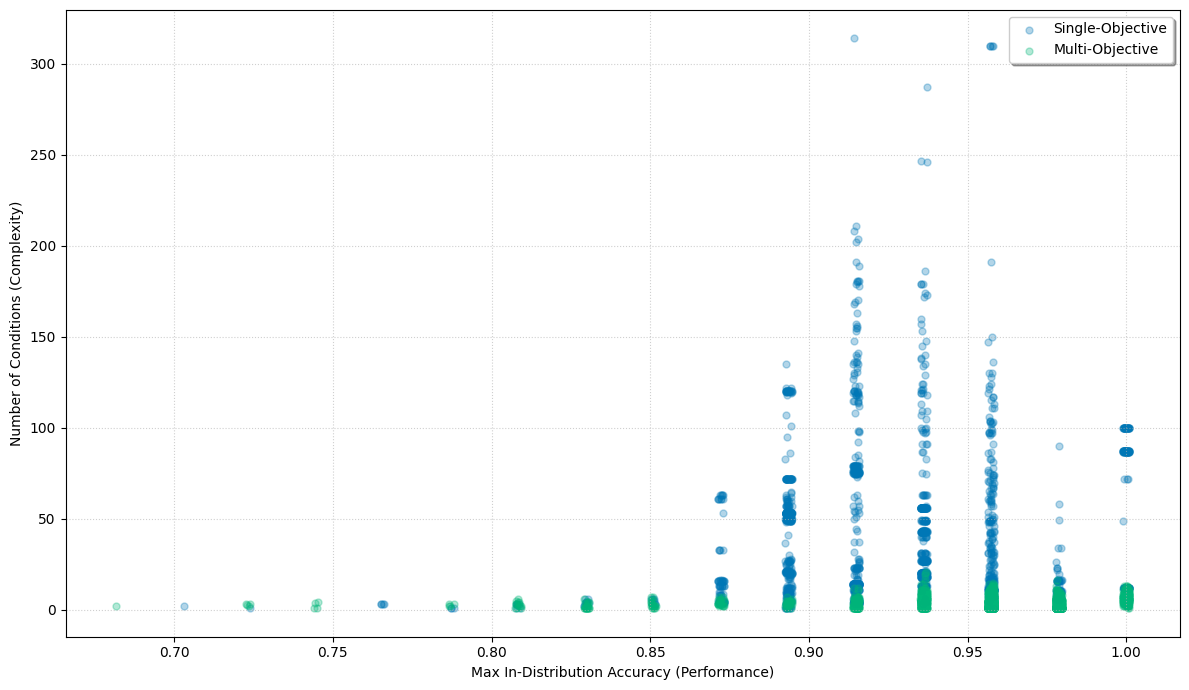}
\hfill
\includegraphics[width=0.49\linewidth]{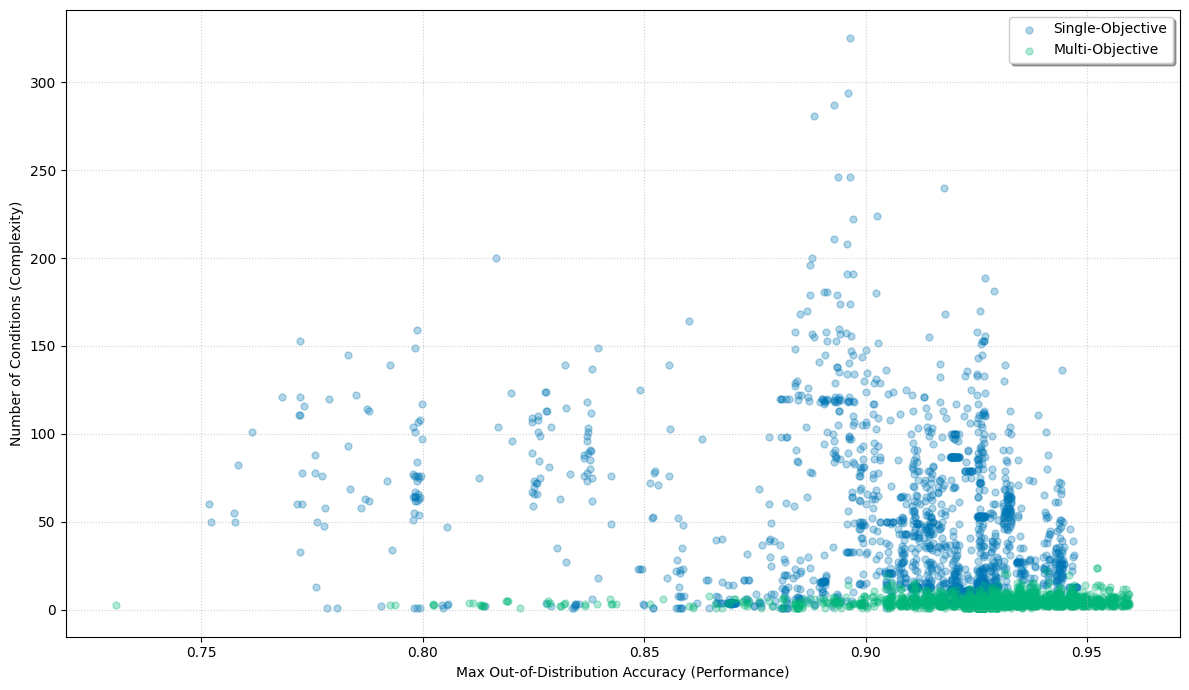}\\[-1ex]
\centerline{\small 90\% ID/OOD split: In-Distribution (Left) vs. Out-of-Distribution (Right)}
\end{minipage}\\[2ex]
\begin{minipage}{1.0\textwidth}
\includegraphics[width=0.49\linewidth]{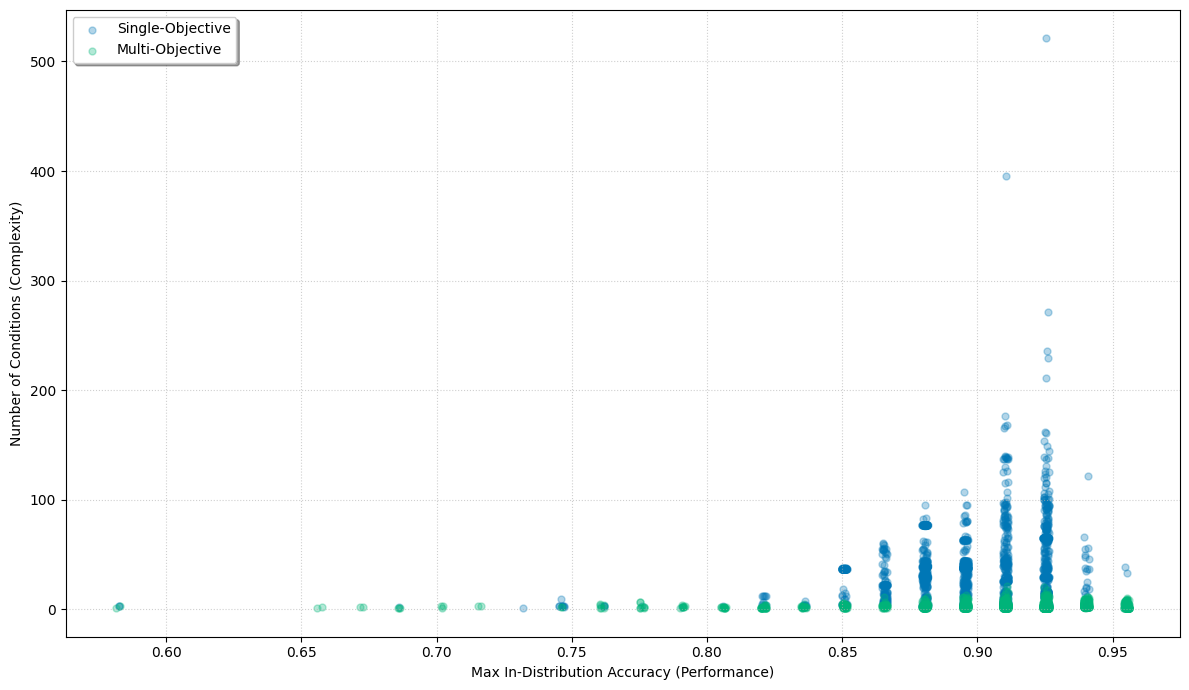}
\hfill
\includegraphics[width=0.49\linewidth]{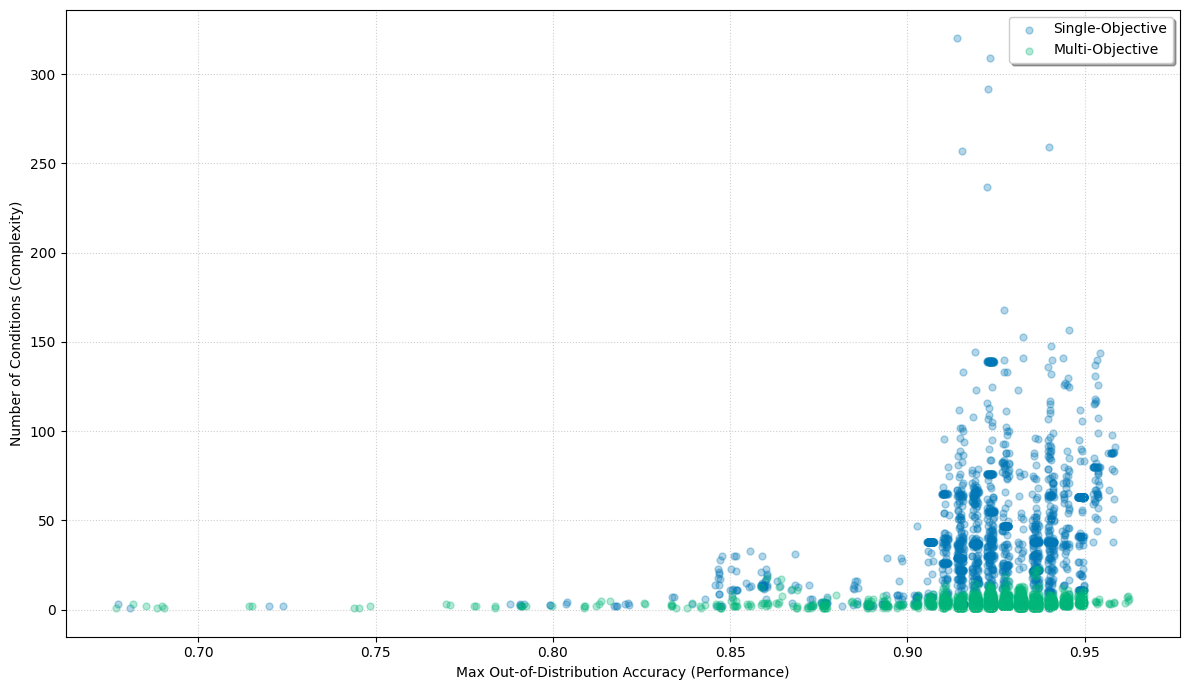}
\centerline{\small 95\% ID/OOD split: In-Distribution (Left) vs. Out-of-Distribution (Right)}
\end{minipage}
\vspace*{-1ex}
\caption{Accuracy vs.\ conciseness in single and multi-objective NeuroRule evolutionary runs at different ID/OOD splits on the Breast Cancer dataset. The ID results initial form vertical lines due to limited data but gradually become more scattered like the OOD results as more data is introduced in the training interval.  Otherwise, the results are similar to those with the Diabetes dataset}
\label{fig:bc_pareto_comparisons}
\end{figure}

\begin{figure}[!t]
\centering
\begin{minipage}{1.0\textwidth}
\includegraphics[width=0.49\linewidth]{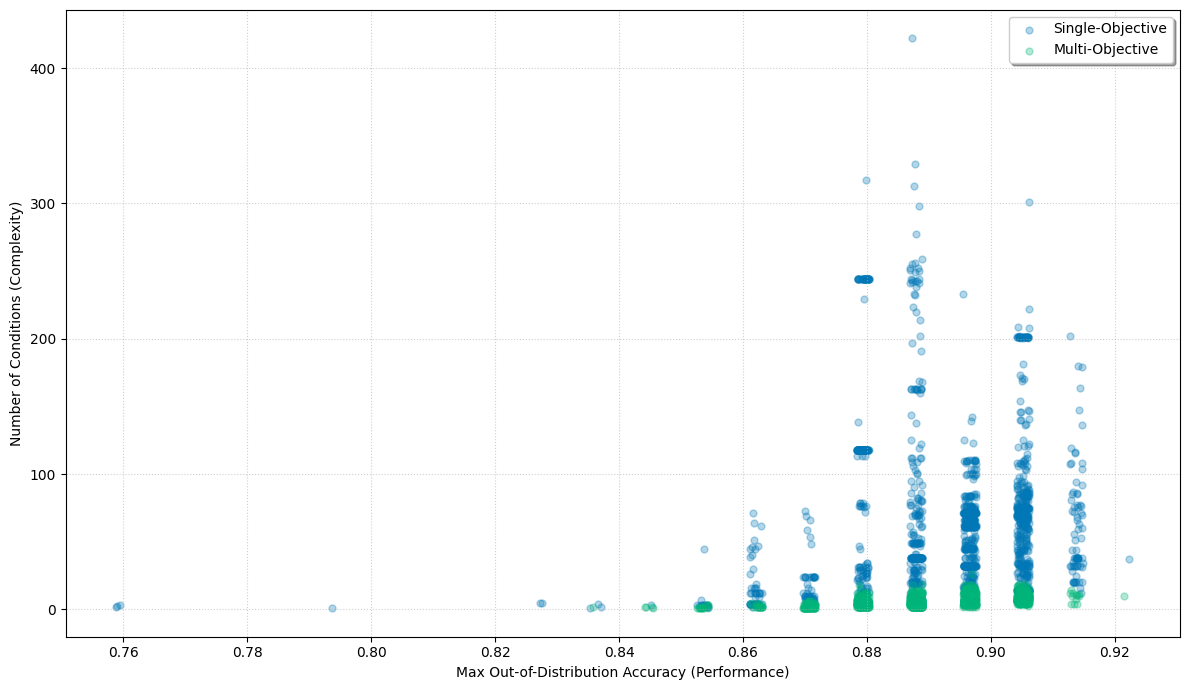}
\hfill
\includegraphics[width=0.49\linewidth]{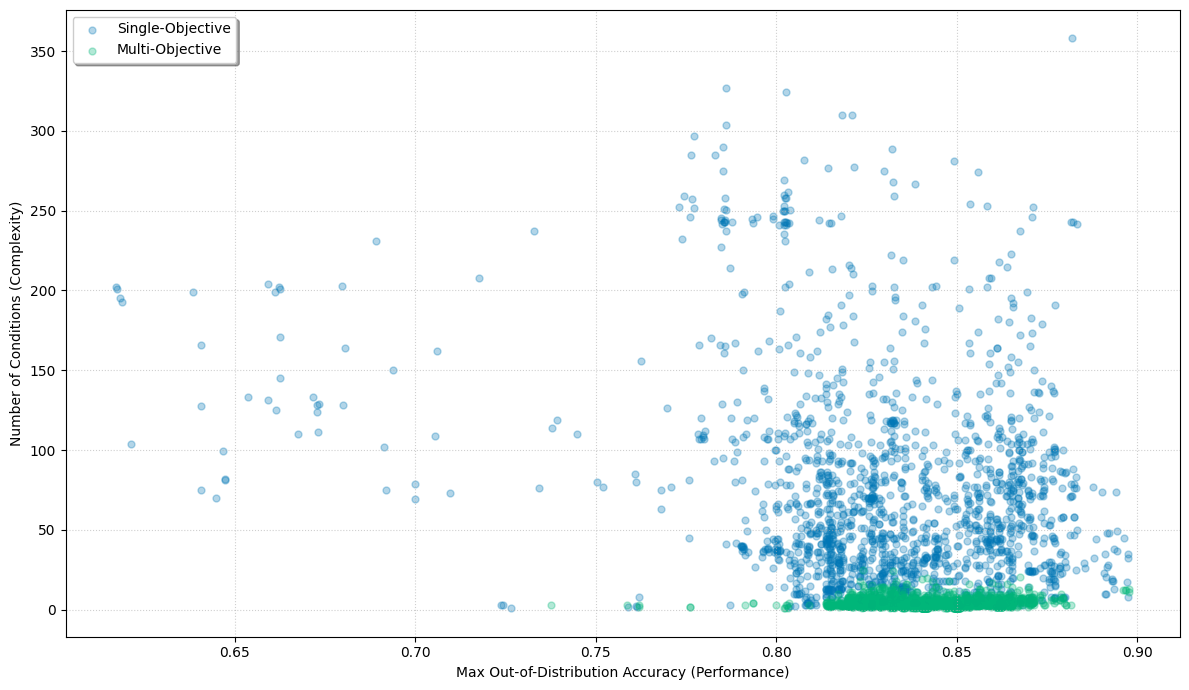}\\[-1ex]
\centerline{\small 80\% ID/OOD split: In-Distribution (Left) vs. Out-of-Distribution (Right)}
\end{minipage}\\[2ex]
\begin{minipage}{1.0\textwidth}
\includegraphics[width=0.49\linewidth]{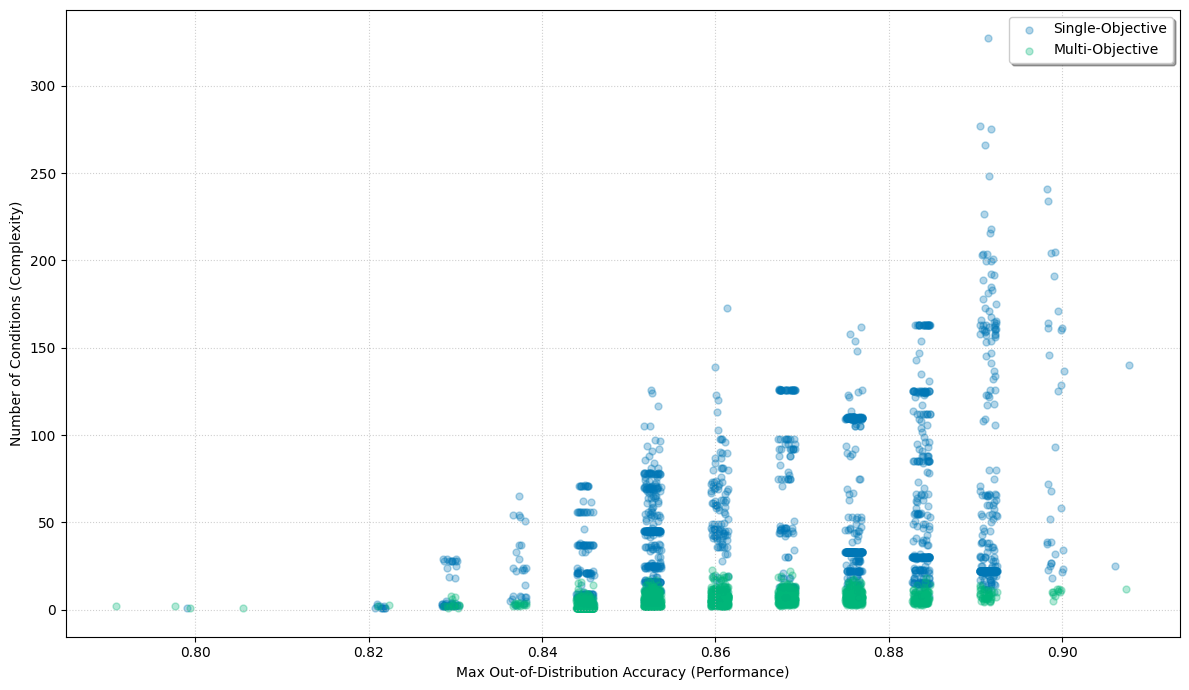}
\hfill
\includegraphics[width=0.49\linewidth]{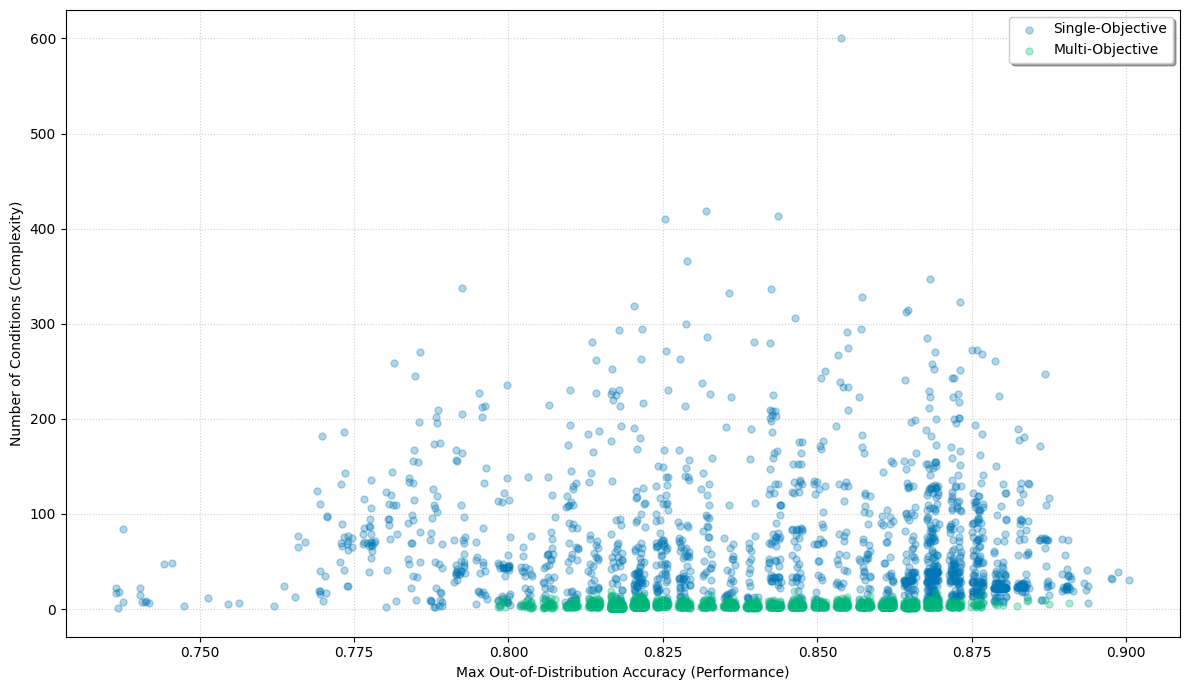}\\[-1ex]
\centerline{\small 90\% ID/OOD split: In-Distribution (Left) vs. Out-of-Distribution (Right)}
\end{minipage}\\[2ex]
\begin{minipage}{1.0\textwidth}
\includegraphics[width=0.49\linewidth]{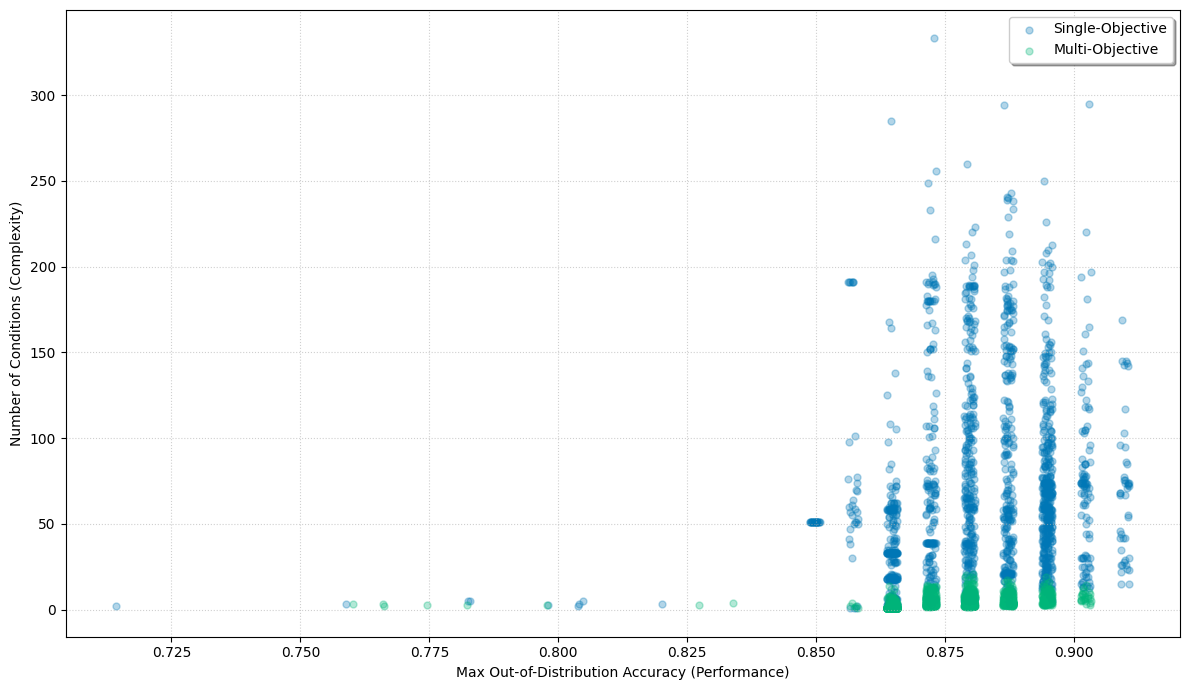}
\hfill
\includegraphics[width=0.49\linewidth]{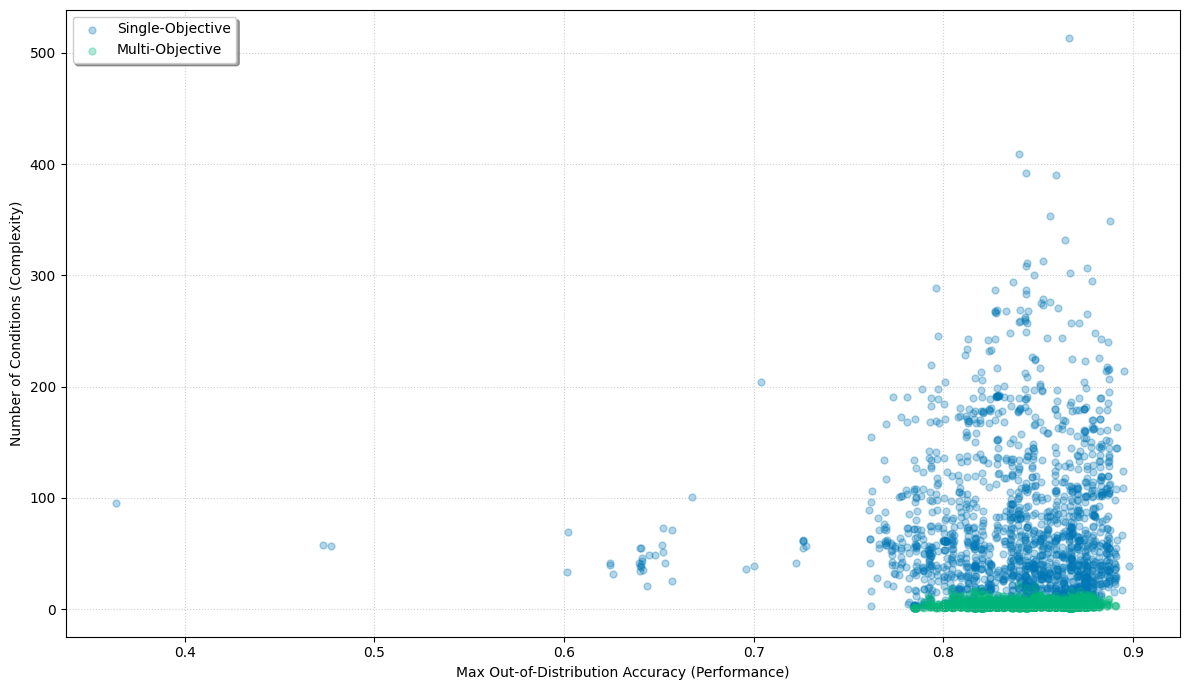}\\[-1ex]
\centerline{\small 95\% ID/OOD split: In-Distribution (Left) vs. Out-of-Distribution (Right)}
\end{minipage}
\vspace*{-1ex}
\caption{Accuracy vs.\ conciseness in single and multi-objective NeuroRule evolutionary runs at different ID/OOD splits on the Heart Disease dataset. As in Figure~\ref{fig:bc_pareto_comparisons}, the ID results initial form vertical lines due to limited data but gradually become more scattered like the OOD results as more data is introduced in the training interval. The results are otherwise similar to the Diabetes dataset.}
\label{fig:heart_pareto_comparisons}
\end{figure}

The learning curves for Breast Cancer and Heart Disease are in Figures~\ref{fig:bc_multi_curves} and~\ref{fig:heart_multi_curves}, and the accuracy vs.\ conciseness plots in Figures~\ref{fig:bc_pareto_comparisons} and~\ref{fig:heart_pareto_comparisons}. The ID results initial form vertical lines due to limited data but gradually become more scattered like the OOD results as more data is introduced in the training interval. The accuracy vs.\ conciseness plots for the EVOTER experiments are similar to those for NeuroRule and omitted for conciseness.

\clearpage
\section{Experiment~3 Learning Curves for Breast Cancer and Heart Disease Datasets}
\label{appendix:synthetic}

\begin{figure}[!b]
\centering
\begin{minipage}{1.0\textwidth}
\includegraphics[width=0.49\linewidth]{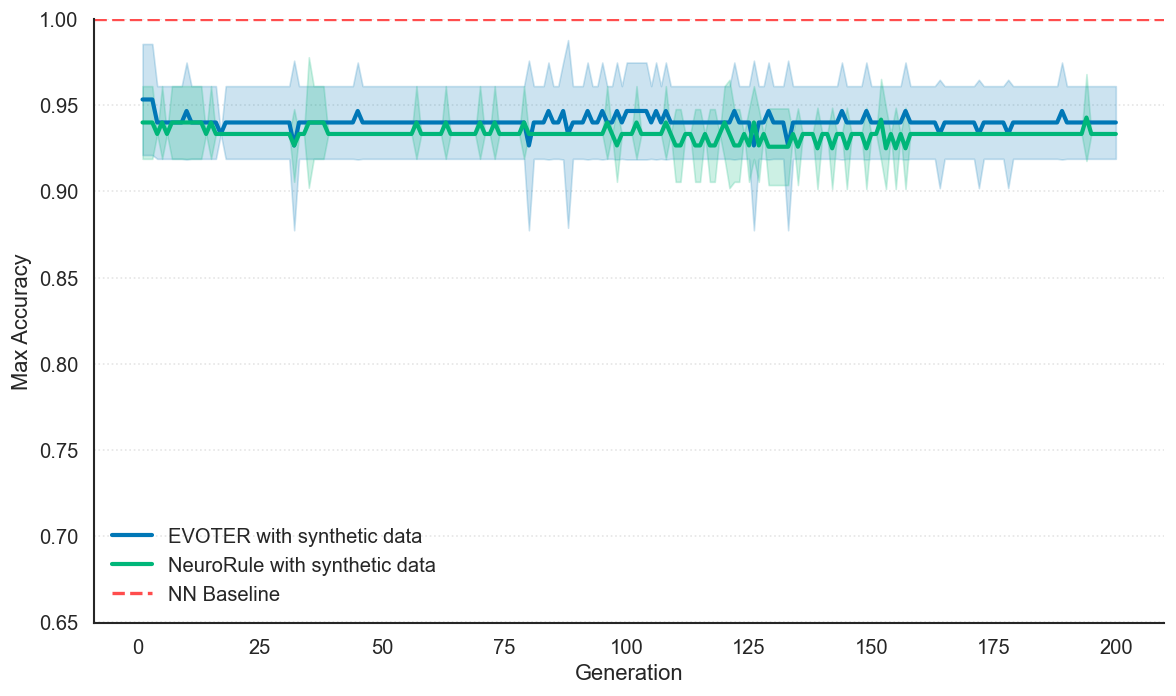}
\hfill
\includegraphics[width=0.49\linewidth]{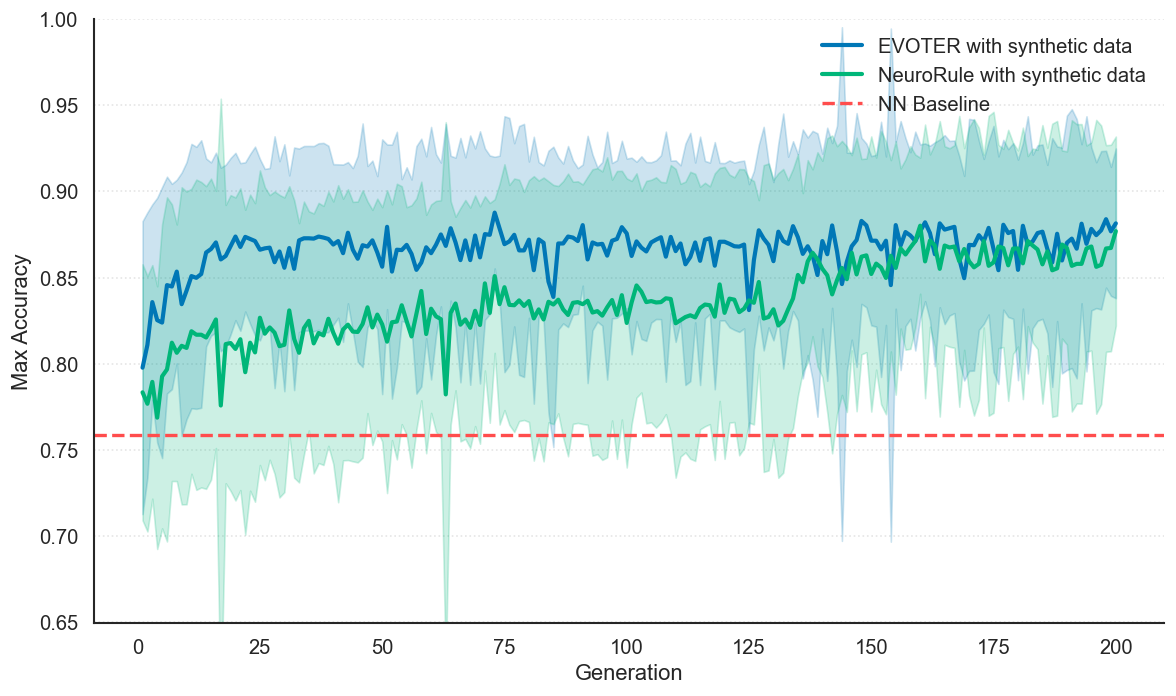}\\[-1ex]
\centerline{\small 80\% ID/OOD split: In-Distribution (Left) vs. Out-of-Distribution (Right)}
\end{minipage}\\[3ex]
\begin{minipage}{1.0\textwidth}
\includegraphics[width=0.49\linewidth]{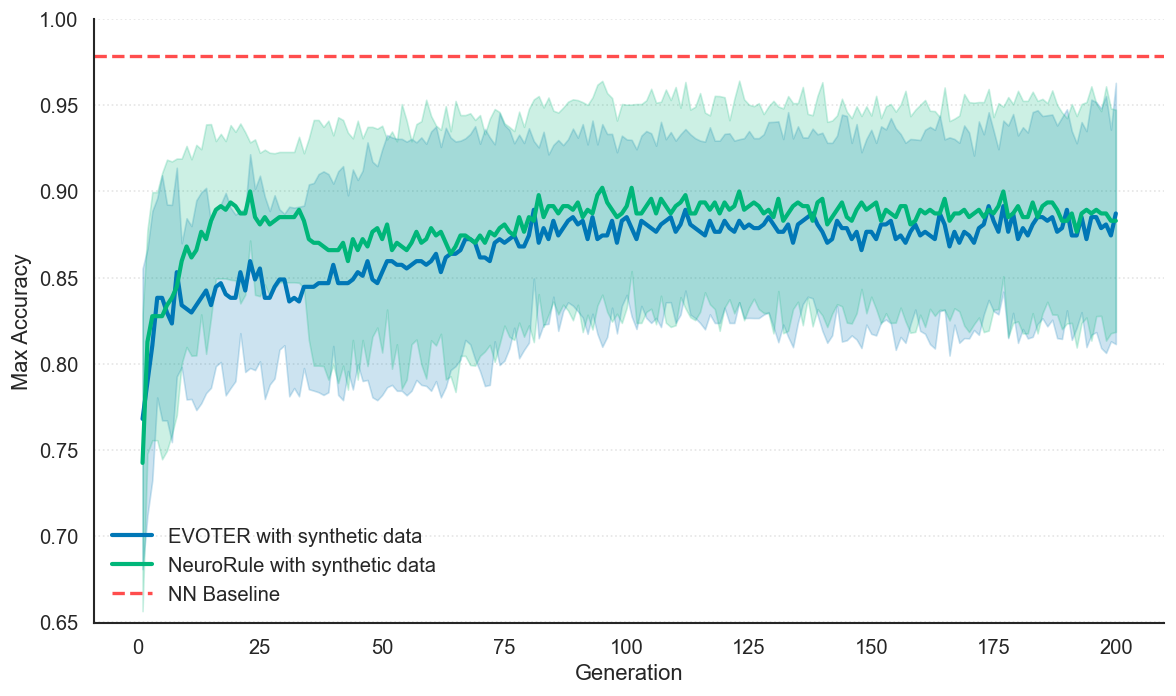}
\hfill
\includegraphics[width=0.49\linewidth]{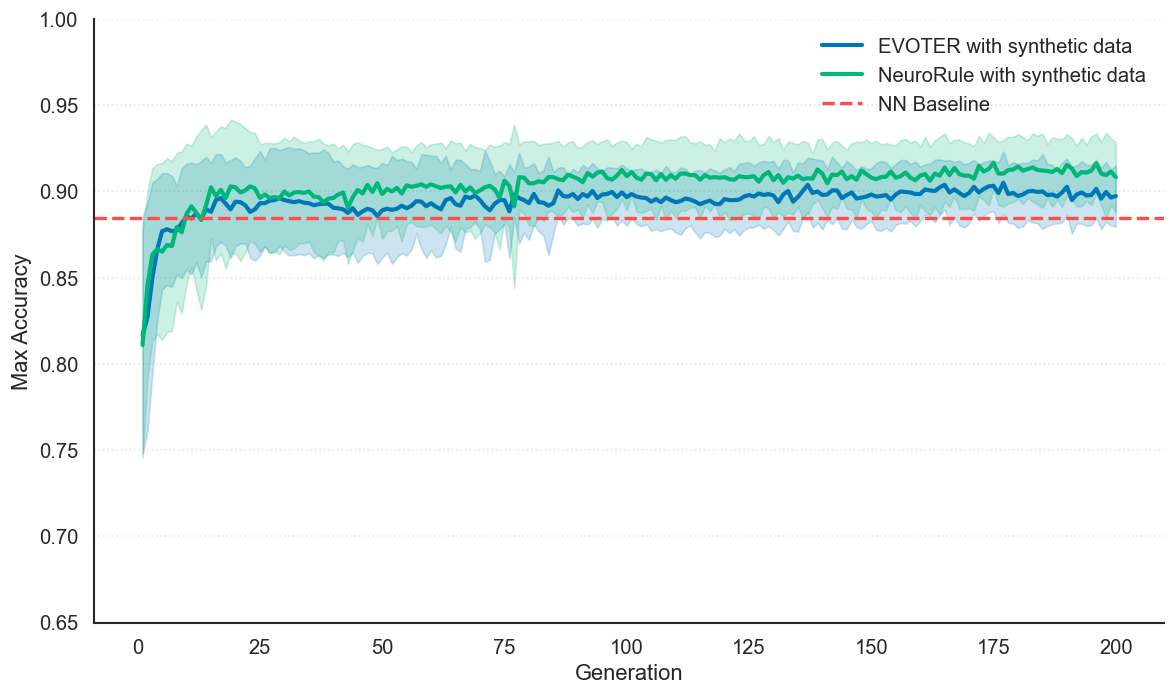}\\[-1ex]
\centerline{\small 90\% ID/OOD split: In-Distribution (Left) vs. Out-of-Distribution (Right)}
\end{minipage}\\[3ex]
\begin{minipage}{1.0\textwidth}
\includegraphics[width=0.49\linewidth]{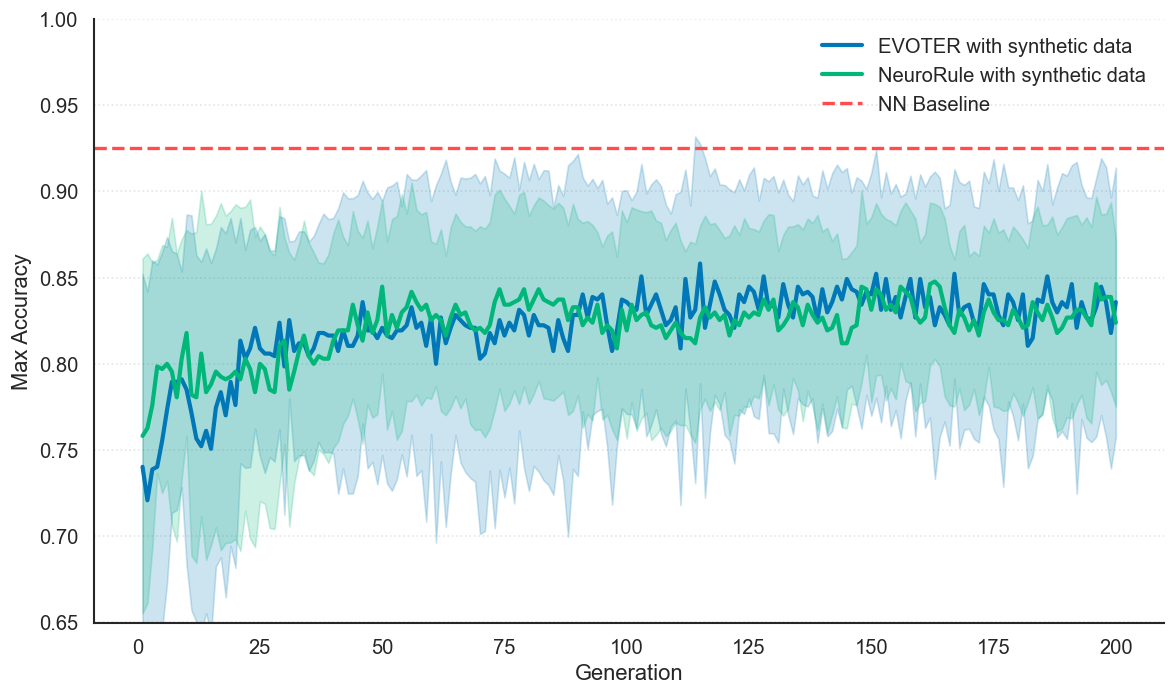}
\hfill
\includegraphics[width=0.49\linewidth]{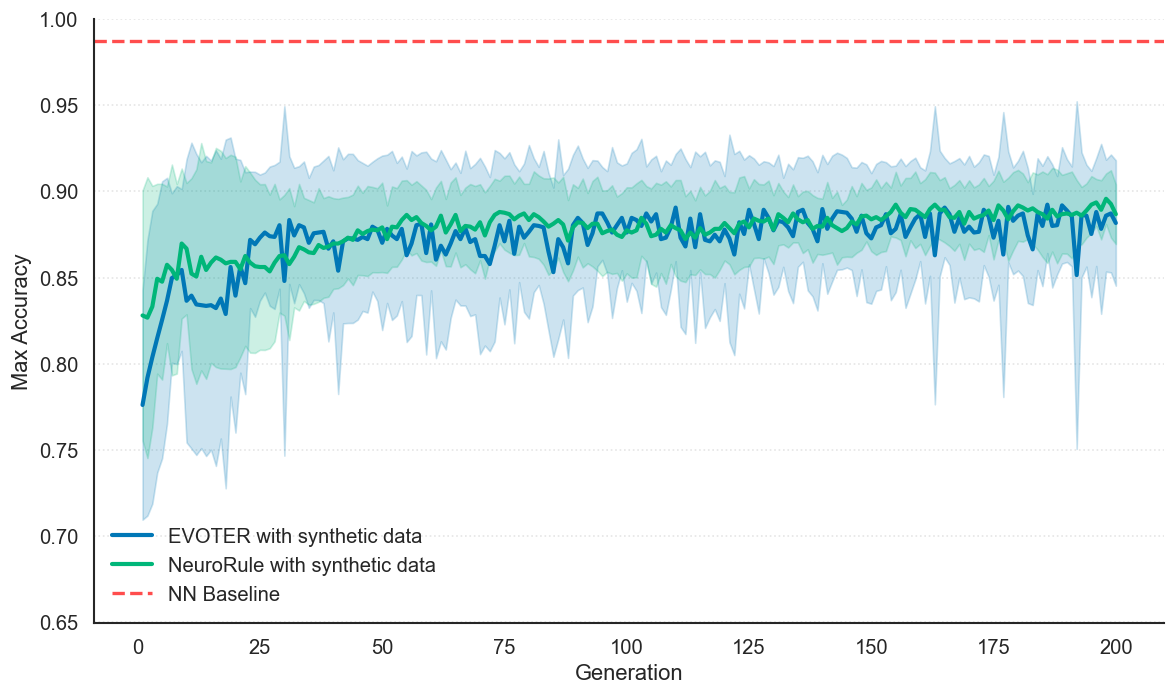}\\[-1ex]
\centerline{\small 95\% ID/OOD split: In-Distribution (Left) vs. Out-of-Distribution (Right)}
\end{minipage}
\vspace*{-1ex}
\caption{Learning curves for EVOTER and NeuroRule compared to NN accuracy at different ID/OOD splits on the Breast Cancer dataset when NeuroRule was evolved with synthetic data only. The results are similar to those with the Diabetes dataset, demonstrating lower but still viable accuracy even when the original data is not available.}
\label{fig:bc_synthetic}
\end{figure}

\begin{figure}[!t]
\centering
\begin{minipage}{1.0\textwidth}
\includegraphics[width=0.49\linewidth]{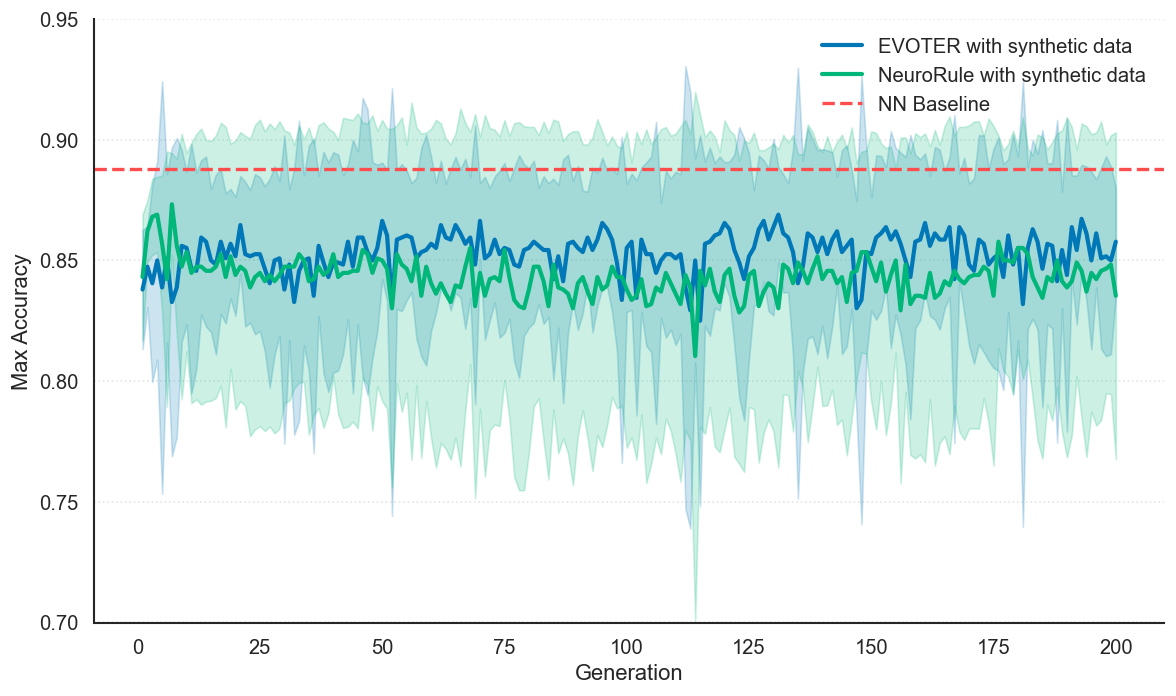}
\hfill
\includegraphics[width=0.49\linewidth]{samples/data/bc/synthetic/ood/0.80.png}\\[-1ex]
\centerline{\small 80\% ID/OOD split: In-Distribution (Left) vs. Out-of-Distribution (Right)}
\end{minipage}\\[3ex]
\begin{minipage}{1.0\textwidth}
\includegraphics[width=0.49\linewidth]{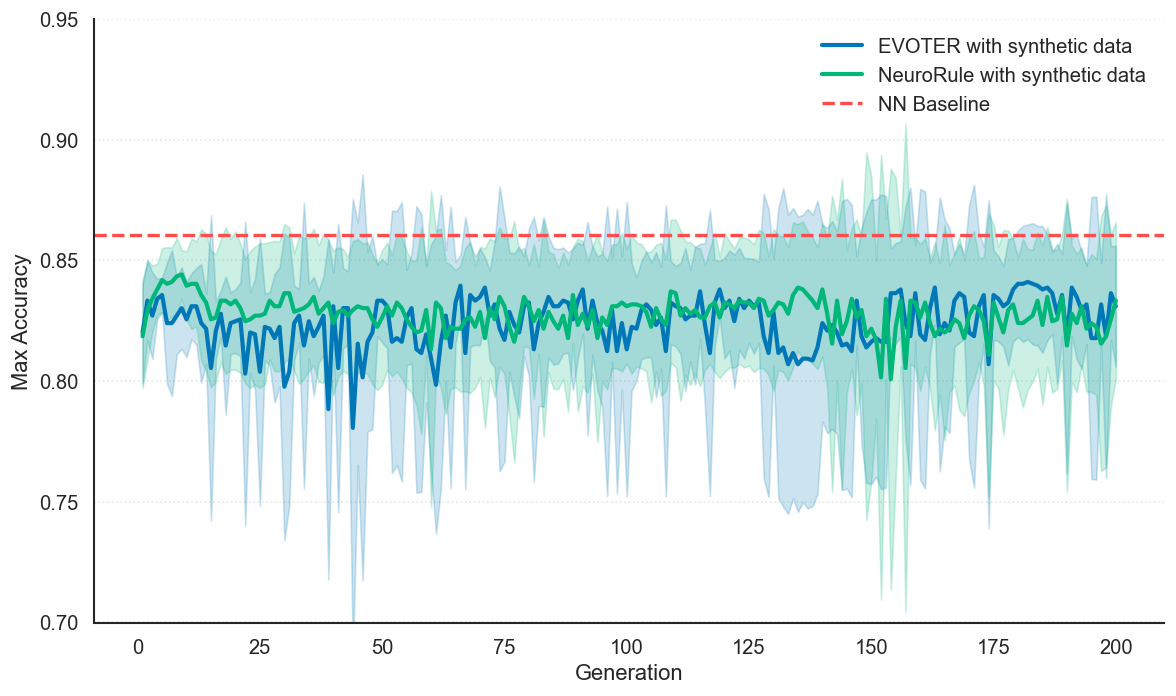}
\hfill
\includegraphics[width=0.49\linewidth]{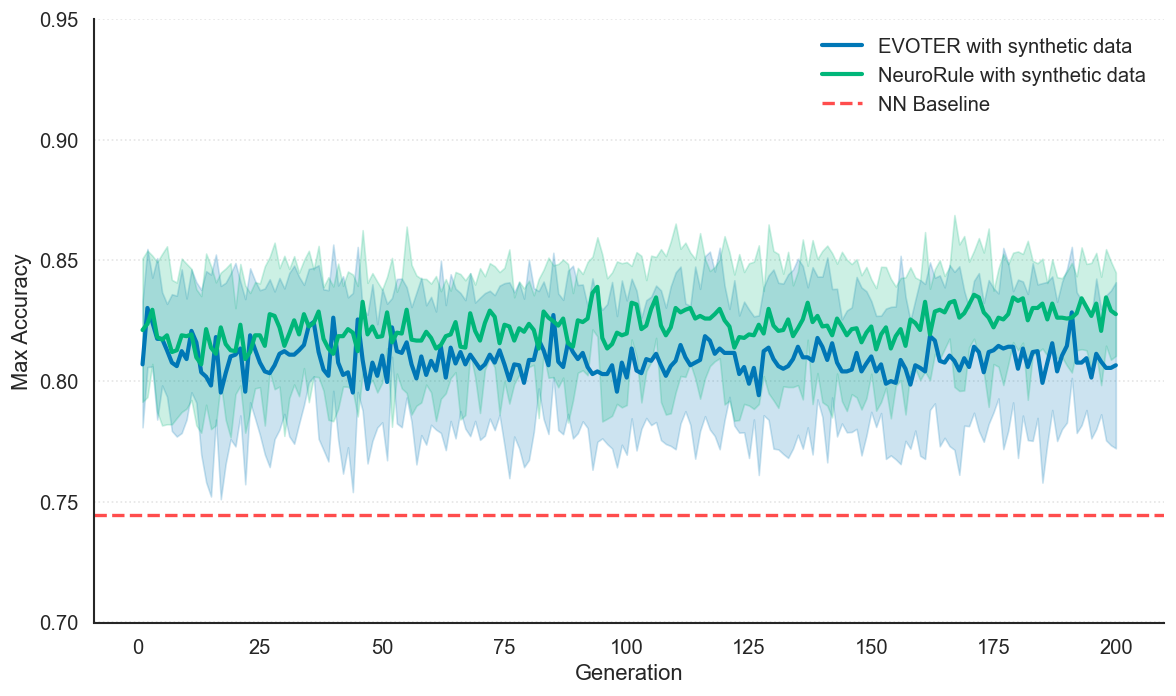}\\[-1ex]
\centerline{\small 90\% ID/OOD split: In-Distribution (Left) vs. Out-of-Distribution (Right)}
\end{minipage}\\[3ex]
\begin{minipage}{1.0\textwidth}
\includegraphics[width=0.49\linewidth]{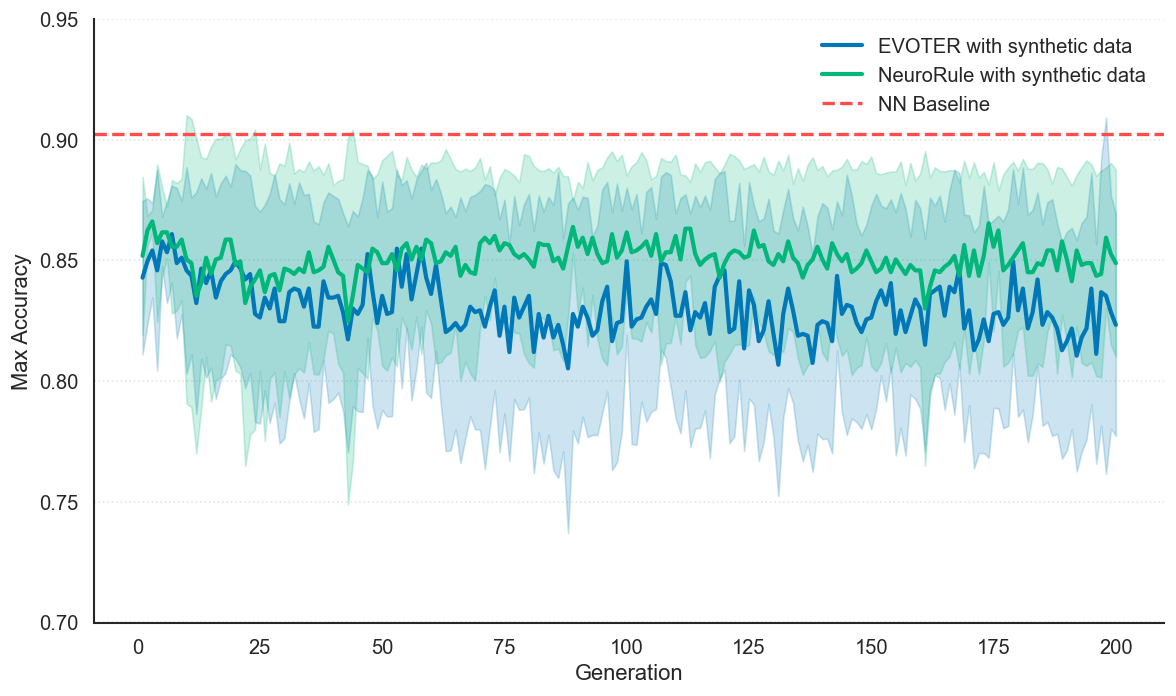}
\hfill
\includegraphics[width=0.49\linewidth]{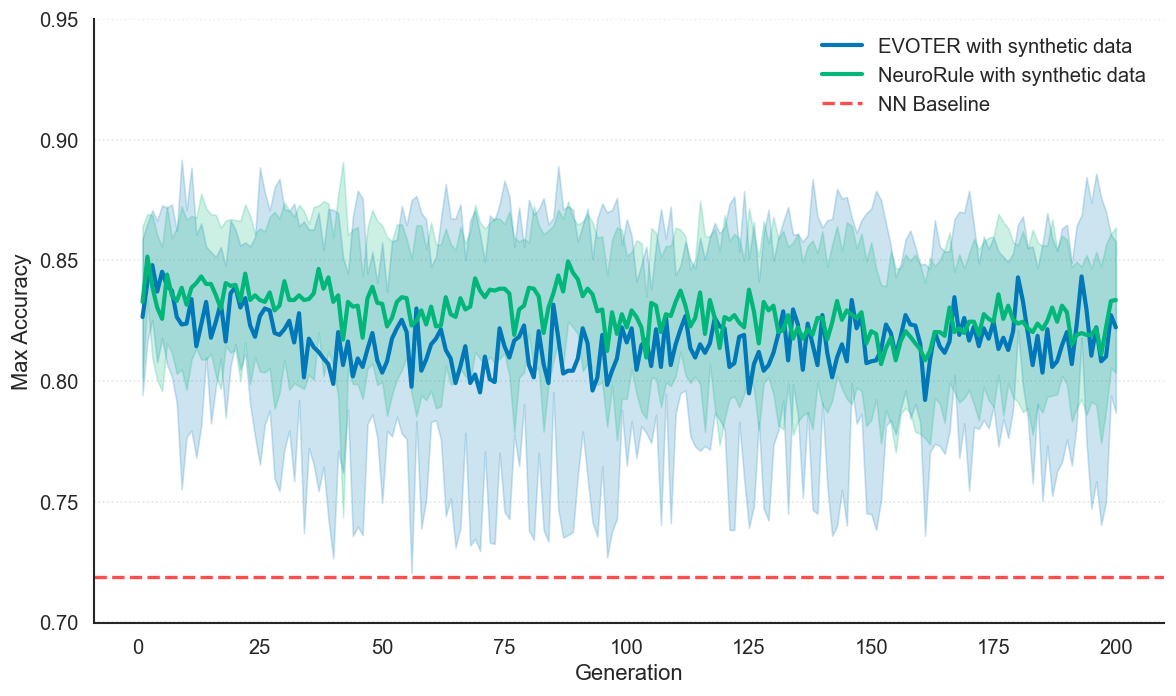}\\[-1ex]
\centerline{\small 95\% ID/OOD split: In-Distribution (Left) vs. Out-of-Distribution (Right)}
\end{minipage}
\vspace*{-1ex}
\caption{Learning curves for EVOTER and NeuroRule compared to NN accuracy at different ID/OOD splits on the Heart Disease dataset when NeuroRule was evolved with synthetic data only. The results are similar to those with the Diabetes dataset, demonstrating lower but still viable accuracy even when the original data is not available.}
\label{fig:heart_synthetic}
\end{figure}

The learning curves for the Breast Cancer and Heart Disease Datasets are in Figures~\ref{fig:bc_synthetic} and~\ref{fig:heart_synthetic}.

\end{document}